\documentclass[10pt,twocolumn,letterpaper]{article}

\usepackage{arxiv}
\usepackage{times}
\usepackage{epsfig}
\usepackage{graphicx}
\usepackage{amsmath}
\usepackage{amssymb}

\usepackage{cite}
\usepackage{hyperref}
\usepackage{amsfonts}
\usepackage{amsmath}
\usepackage{graphicx}
\usepackage{enumitem}
\usepackage[dvipsnames]{xcolor}
\usepackage{booktabs}
\usepackage[export]{adjustbox}
\usepackage{bm}
\usepackage{makecell}
\usepackage{colortbl}
\usepackage{rotating}
\usepackage{listings}
\usepackage{times}
\usepackage{epsfig}
\usepackage{graphicx}
\usepackage{amsmath}
\usepackage{amssymb}
\usepackage{flushend}
\usepackage{comment}
\usepackage{ragged2e}
\usepackage{pifont}

\definecolor{mygreen}{rgb}{0,0.6,0}
\definecolor{mygray}{rgb}{0.5,0.5,0.5}
\definecolor{mymauve}{rgb}{0.58,0,0.82}

\newcommand{\shape}{\mathcal{S}}
\newcommand{\surface}{S}
\newcommand{\loss}{\mathcal{L}}
\newcommand{\recons}{\text{recons}}
\newcommand{\opposite}{\text{opp}}
\newcommand{\same}{\text{same}}
\newcommand{\lossrecons}{\loss_\recons}
\newcommand{\losssame}{\loss_\same}
\newcommand{\lossopp}{\loss_\opposite}

\newcommand{\vect}[1]{\mathbf{#1}}

\newcommand{\decoder}{\phi}

\newcommand{\network}{\Phi}
\newcommand{\latent}{\mathbf{z_\shape}}
\newcommand{\pointcloud}{P}
\newcommand{\rthree}{\mathbb{R}^3}

\newcommand{\bernoulli}{\mathcal{B}}
\newcommand{\bprob}{b}
\newcommand{\normal}{\mathcal{N}}
\newcommand{\BCE}{{\rm BCE}}

\newcommand{\cmark}{\textcolor{ForestGreen}{\ding{51}}}%
\newcommand{\xmark}{\textcolor{BrickRed}{\ding{55}}}%

\newcommand*\rot{\rotatebox{90}}

\newcommand{\method}{{NeeDrop}}
\newcommand{\articleTitle}{NeeDrop: Self-supervised Shape Representation from Sparse Point Clouds \\ using Needle Dropping}

\newcommand{\articleAbstract}{
There has been recently a growing interest for implicit shape representations. Contrary to explicit representations, they have no resolution limitations and they easily deal with a wide variety of surface topologies. To learn these implicit
representations, current approaches rely on a certain level of shape supervision (e.g., inside/outside information or distance-to-shape knowledge), or at least require a dense point cloud (to approximate well enough the distance-to-shape). In contrast, we introduce {\method}, an self-supervised method for learning shape representations from possibly extremely sparse point clouds.
Like in Buffon's needle problem, we ``drop'' (sample) needles on the point cloud and consider that, statistically, close to the surface, the needle end points lie on opposite sides of the surface.
No shape knowledge is required and the point cloud can be highly sparse, e.g., as lidar point clouds acquired by vehicles. Previous self-supervised shape representation approaches fail to produce good-quality results on this kind of data.
We obtain quantitative results on par with existing supervised approaches on shape reconstruction datasets and show promising qualitative results on hard autonomous driving datasets such as KITTI.
}

\begin{document}
\title{\articleTitle}

\author{
Alexandre Boulch$^1$
\and 
Pierre-Alain Langlois$^2$
\and
Gilles Puy$^1$
\and
Renaud Marlet$^{1,2}$
\and
\large
\hspace{-3mm}\textsuperscript{1}Valeo.ai, Paris, France  \hspace{1mm} \textsuperscript{2}LIGM, Ecole des Ponts, Univ Gustave Eiffel, CNRS, Marne-la-Vall\'ee, France
}

\maketitle
\thispagestyle{empty}

\begin{abstract}
   \articleAbstract
\end{abstract}

\section{Introduction}

Learning-based approaches for 3D reconstruction from sparse 3D data have recently attracted of lot of interest. As opposed to classical approaches~\cite{berger2017survey}, such as Poisson reconstruction~\cite{kazhdan2006poisson}, these methods enable prior knowledge to be used to enrich the representation of information-deficient inputs, e.g., low density point clouds or partial scene views.

Most of the recent learning-based methods for shape reconstruction from point clouds fall into two categories.
A first category produces an \emph{explicit} or \emph{parametric} representation of the shape: point cloud, voxel or mesh.
For instance, some may deform a geometric primitive or a template mesh~\cite{groueix2018papier,groueix2018b}, e.g., a planar patch or a sphere. The topology of both the template and the reconstruction are thus identical, which is a significant limitation.
The second category of methods operates on an \emph{implicit} formulation of the shapes. These methods build a continuous function over the 3D space, either based on occupancies~\cite{mescheder2019occupancy}, or on signed~\cite{park2019deepsdf} or unsigned~\cite{chibane2020ndf} distance functions.
They are not bound by the topology of a template but require an extra-processing step for mesh extraction.
Beyond shape reconstruction, several approaches also tackle the problem of shape generation by encoding shapes in a low-dimensional latent space with a constraint on the distribution of the latent variables.
All existing learning-based approaches use a certain level of supervision during training, either using the information of distance to the shape or using the knowledge of which points fall inside or outside the shape.

In this work, we propose what we believe is the first \emph{self-supervised} approach for shape reconstruction from \emph{sparse} point cloud. It is self-supervised in the sense that it does not need to learn from actual shapes, or even from densely sampled point clouds; sparse, noisy, and even partial point clouds are enough. And it reconstructs a shape in the sense that an actual mesh can directly be produced from the occupancy field we predict, e.g., using a Marching Cubes algorithm \cite{lorensen1987marching}, as opposed to approaches that only produce renderings or generate points on the underlying surface~\cite{cai2020learning}.

Like some other methods, we build a shape representation in a latent space and predict an implicit occupancy field. An actual surface can then be easily extracted as the zero-level set of the function. However, unlike most existing methods, we only use point clouds as input for learning, instead of meshes. We thus do not have to worry about shape watertightness, which is a major concern for occupancy-based methods training on meshes because most shape datasets collect good-looking but actually ill-formed meshes, thus requiring a significant preprocessing.
Furthermore, we show that our method can deal with highly sparse input point clouds without surface supervision, such as point clouds captured by Lidars on moving vehicles, whereas all existing methods that use point clouds as input (except \cite{cai2020learning}) assume that points are dense enough so that the distance to the shape (supervision) can be closely approximated by the distance to the point cloud, which leads to failures to learn from sparse inputs.

Our method is inspired by Buffon’s needle problem~\cite{buffon1777} where one wants to compute the probability that a needle dropped on a wooden floor with parallel strips lies across two strips. Similarly, we drop (sample) needles on the point cloud such that needles built on input points have a high probability of crossing the surface, while needles constructed a bit away from input points have a low probability.
%
Our main contributions are:
\begin{itemize}[noitemsep,topsep=0pt]
    \item a new loss for self-supervised shape reconstruction formulated via needle dropping on the point cloud;
    \item the use of this loss to learn only from point clouds how to predict shape representations;
    \item a overall method that can intrinsically deal with highly sparse and partial point cloud, at train and test time;
    \item the validation of our method on both synthetic data and real point clouds for which no supervision is possible.
\end{itemize}

The paper is organized as follows: section~\ref{sec:related} presents related work; section~\ref{sec:method} describes the loss function and the network used for shape reconstruction; section~\ref{sec:experiments} presents experiments showing the performance of our method and its comparison with competing techniques.

\section{Related work}
\label{sec:related}

Representing shapes, with applications such as reconstruction and generation, has been widely studied.
This section only presents \emph{learning}-based methods; classical (non-learning-based) methods are surveyed in~\cite{berger2017survey}.
We classify related work based on the type of shape representation used and on the level of supervision used at training time.

\textbf{Point clouds}
are a common way to represent a shape. They can be obtained, e.g., directly from depth sensors, or via photogrammetry.
Pioneered by PointNet-based architectures\cite{qi2017pointnet,qi2017pointnet++}, learning-based methods have reached the state of the art in multiple tasks such as classification and semantic segmentation \cite{li2018pointcnn,thomas2019kpconv,boulch2020fkaconv}.
Point clouds have also been successful for shape generation from images~\cite{fan2017point,insafutdinov2018unsupervised,pumarola2020c} or from a prior distribution~\cite{yang2019pointflow}.
Yet, a point cloud remains a sparse representation of the underlying surface and, when generating points, their number is often a fixed parameter.

\textbf{Voxels}
allow to directly adapt techniques developed on 2D pixels to 3D data.
They are used in many tasks ranging from classification and semantic segmentation \cite{maturana2015voxnet,qi2016volumetric,wu20153d} to shape generation \cite{tatarchenko2017octree,wang2018global}, representation \cite{kar2017learning,ji2017surfacenet,rezende2016unsupervised} or completion~\cite{dai2017shape,stutz2018learning}.
Represented as an occupancy grid, the reconstructed surface however suffers from quantization effects, which can be mitigated using truncated signed distance functions \cite{curless1996volumetric,dai2017shape,ladicky2017point,riegler2017octnetfusion,stutz2018learning}\rlap.

Besides, using voxels may rapidly lead to a high memory consumption as their number grows cubicly with the size of the scene. A greater accuracy, with finer voxel grids, may be obtained at the cost of a slow training process \cite{wu20153d,choy20163d,tulsiani2017multi} or by integrating surface extraction for occupancy~\cite{liao2018deep}.
Sparse convolutions \cite{graham2014spatially,graham20183d,choy20194d} or multi-resolution approaches~\cite{hane2017hierarchical,wang2018global} such as octrees \cite{riegler2017octnet,tatarchenko2017octree,riegler2017octnetfusion}  can be used to reduce the memory footprint and scale to a complete scene~\cite{yi2020complete}.

\textbf{Meshes}
describe a shape as a set of vertices and faces.
They are the representation of choice in computer graphics and computer-aided design as they are very memory efficient and easily allow geometric operations and rendering with texture. 
Geometric deep learning~\cite{bronstein2017geometric} exploits their graph structure for classification and segmentation \cite{defferrard2016convolutional,feng2019meshnet}.

A sub-category of methods operating on meshes deform one or several shape templates~\cite{groueix2018papier,wang2018pixel2mesh,ranjan2018generating,pons2015dyna}.
While giving good results when the templates and the shapes are relatively similar, the resulting meshes necessarily have the same topology as the templates, which limits the complexity of the shapes that can be modeled. Besides, they may suffer from self-intersections caused by wide deformations.
Template deformation can also be mixed with voxels, to refine a coarse reconstruction~\cite{Gkioxari_2019_ICCV}.
Other approaches directly predict sets of vertices and faces~\cite{dai2019scan2mesh}. The result may not be continuous and can require additional mesh fixing steps.

\textbf{Implicit representations}
model a closed shape via a continuous function of the 3D space.
An actual surface is then extracted as a levelset of the function (or its gradient).

Most existing approaches associate to each point $\mathbf{x}\in \rthree$ a signed distance~\cite{park2019deepsdf,chen2019learning,michalkiewicz2019implicit,jiang2020local,atzmon2020sal, atzmon2021sald, gropp2020implicit} or an occupancy value~\cite{mescheder2019occupancy,genova2020local,chibane2020implicit,deng2020nasa}.
Such implicit representations lead to the reconstruction of closed surfaces (with an inside and an outside) without template-topology limitations.
In this work, the network is trained to output an occupancy value indicating on which side of the shape a query point is.

Unsigned distance fields \cite{chibane2020ndf, cai2020learning} have the advantage to be able to model open surfaces too. But they do not directly yield a mesh; they generate a dense point cloud, that is then given to a meshing algorithm to generate an actual surface.

Some hybrid methods predict a set of convex polytopes with one implicit function per polytope, to produce a piecewise representation of a shape~\cite{tulsiani2017learning,kluger2021cuboids}.
Others reconstruct a globale mesh by predicting voxel occupancy as well as a local mesh configuration~\cite{liao2018deep}.

\begin{figure}
    \centering
    \begin{tabular}{cc}
        \includegraphics[width=0.4\linewidth]{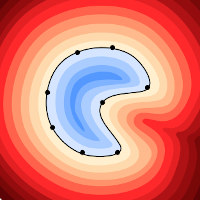}&
        \includegraphics[width=0.4\linewidth]{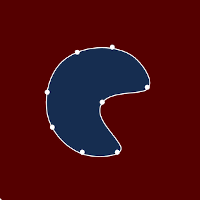}\\
        \small (a) Signed distance to $\shape$ &
        \small (b) Occupancy of $\shape$ \\
        \includegraphics[width=0.4\linewidth]{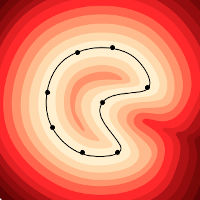}&
        \includegraphics[width=0.4\linewidth]{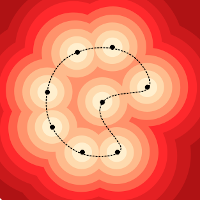}\\
        \small (c) Unsigned distance to $\shape$ &
        \small (d) Distance to nearest point
    \end{tabular}
    \vspace{-0.5mm}
    \caption{Different kinds of supervision.}
    \vspace{-2mm}
    \label{fig:supervision}
\end{figure}

\textbf{Supervision level} is key differentiating factor too. The full supervision of the real, signed distance to the shape (Fig.\,\ref{fig:supervision}(a)) is typically tackled by a regression loss \cite{park2019deepsdf,liao2018deep}. Full supervision of the occupancy information (Fig.\,\ref{fig:supervision}(b)) leads to networks predicting if a point is inside or outside the shape~\cite{mescheder2019occupancy,chibane2020implicit}. A weaker supervision only provides the unsigned distance to the shape \cite{chibane2020ndf} (Fig.\,\ref{fig:supervision}(c)). Finally, in the fully self-supervised setting, which is used in this work, only the distance to the input points is available, thus providing only an approximate distance to the shape (Fig.\,\ref{fig:supervision}(d)).

Several methods fall in this last category.
SAL~\cite{atzmon2020sal} learns a sign-agnostic distance and a specific network initialization favors a sign change when crossing the surface, then allowing surface extraction. But sign change is not guaranteed. In contrast, needles enforce it at loss level. Besides, as underlined in~\cite{chibane2020ndf}, SAL tackles single objects reconstruction and often fails in multi-object reconstruction settings.

SALD~\cite{atzmon2021sald} improves SAL by adding a term favoring supervised gradients on the surface to enforce sign change. However, it introduces an extra hyperparameter to balance reconstruction and regularization, which can be hard to tune because the balance depends on point density, whereas our two loss terms are homogeneous and simply summed.

IGR~\cite{gropp2020implicit}, on the contrary, only supervises (a null) distance on the surface but favors a norm-1 gradient everywhere. Here again, a hyperparameter balances reconstruction and regularization, with the same caveat as SALD.

\textbf{Point cloud sparsity.}
These methods use very dense point clouds: 500k points for SALD \cite{atzmon2021sald}, 16k for SAL \cite{atzmon2020sal} and 8k for IGR \cite{gropp2020implicit}. Distance to the nearest point (Fig\,\ref{fig:supervision}(d)) then becomes a very good approximation of the unsigned distance (Fig.\,\ref{fig:supervision}(c)). On the contrary, our method can cope with very sparse point clouds (300 points in our experiments), much more like in Fig.\,\ref{fig:supervision}(d).
Yet, the sparse point cloud setting is investigated in SAL~\cite{atzmon2020sal} for single shape reconstruction.
However, we observed in our experiments (Tab.\,\ref{tab:exp_complete}) that SAL~\cite{atzmon2020sal} and IGR~\cite{gropp2020implicit} fail to produce meshes when trained on a collection of shapes with a very small number of input points. (No code is available for SALD.)

Only ShapeGF~\cite{cai2020learning} handles very sparse point clouds, but although the paper talks of ``surface extraction'', it does not easily and directly produce a mesh. It actually predicts a shape gradient field that can be used to sample points on the underlying surface using stochastic gradient ascent (and a number of extra parameters), with a density whose uniformity cannot be controlled though. A separate meshing step of those noisy points is then required. (Ray-casting, with extra parameters too, can also produce nice renderings directly from the gradient field, as it also yields normals, which soften noise in rendered pixels.) In contrast, we generate an occupancy field, from which a mesh can be easily extracted \cite{mescheder2019occupancy}. The authors, in their supplementary material~\cite{cai2020learning}, mention failure cases coming from local minima and saddle points (where gradients are close to zero), as well as double surfaces arising when meshing, and they leave the improvement of surface extraction to future work.

\textbf{Reconstruction with partial point clouds.}
Most methods focus on complete shapes, for which a full supervision is possible, or at least can be approximated.
Some methods can deal with partial data as input, but still rely on full shape supervision when training, as DeepSDF~\cite{park2019deepsdf}. It is also the case of single-view reconstruction methods, that take an RGB image as input and reconstruct the whole shape~\cite{groueix2018b, mescheder2019occupancy}.
In~\cite{song2017semantic}, a network is trained from volumetric completion from RGBD data with supervision on the volume to be reconstructed.
In contrast, our method does not require any level of supervision beyond sparse points and can use partial point clouds when training and testing.

\textbf{Needles.}
OccNet~\cite{mescheder2019occupancy} traces rays of possibly infinite length to close shapes, while our needles are bounded.
DeepSDF~\cite{park2019deepsdf} has a kind of needle to sign a distance ($\pm \eta$) w.r.t.\ a viewpoint, whereas our needles have uniform orientations and apply to self-supervised surface reconstruction.
In~\cite{mullen2010signing}, the occupancy is computed with a global per-shape optimization for which potentials are estimated based on ray-shooting and robust ray-shape intersection counting.
In~\cite{giraudot2013noise}, the global optimization applies not only to an occupancy assignment but also to a sign-flipping estimation along edges of a graph randomly constructed over nodes placed at the vertices of a regular grid.

\section{Self-supervised reconstruction}
\label{sec:method}

\subsection{Implicit shape representation}

Like previous work aiming at predicting an implicit shape representation, our goal is to approximate the ideal function $f^0_\shape$ such that some $\alpha\,{\in}\, \mathbb{R}$ level-set of $f^0_\shape$ is the surface $\surface$ of the target shape $\shape$:
\begin{equation}
    \surface = \{ \mathbf{x} \in \mathbb{R}^3 | f^0_\shape(x)= \alpha \}
\end{equation}

For signed distance estimation~\cite{park2019deepsdf,atzmon2020sal}, $\alpha\,{=}\,0$: negative values are inside the shape (by convention), and positive values are outside.

We consider here $f^0_\shape$ to be an occupancy function, 
\begin{equation}
f^0_\shape: \rthree \rightarrow \{0,1\}
\end{equation}
such that $f^0_\shape$ takes value 0 outside the shape and 1 inside the shape. 
The surface of the shape is defined as the location of discontinuity between values 0 and 1.

Given an input point cloud $\pointcloud_\shape$ sampled on the shape, the objective is to estimate a function
\begin{equation}
    f_{\shape}: \rthree \rightarrow [0,1]
\end{equation}
such that $f_{\shape} \approx f^0_\shape$. We consider that $f_{\shape}$ belongs to the family of function that can be predicted using a neural network. Thus, $f_{\shape}$ is continuous and we consider that the surface is the $0.5$-level set of the function.

\subsection{Self-supervised learning from pairs of points}

In other approaches, $f^0_\shape$ is
known~\cite{park2019deepsdf,mescheder2019occupancy,chibane2020implicit}, or partially known~\cite{atzmon2020sal,chibane2020ndf} (known distance with great precision, unknown sign) at each point $\vect{x} \in \rthree$.
The neural network can thus be trained directly to approximate $f^0_\shape$ under supervision.
On the contrary, in our self-supervised setting, $f^0_\shape$ is not directly accessible and $\pointcloud_\shape$ is the only available information about $\shape$. To reconstruct $f^0_\shape$ given only $\pointcloud_\shape$, we take inspiration from Buffon's needle problem.

\textbf{Buffon's needle problem.}
In~\cite{buffon1777}, Buffon estimates the probability that the end points of a needle, dropped on a floor made of parallel strips of wood, lie on different strips.

\begin{figure}
    \centering
    \includegraphics[width=0.75\linewidth]{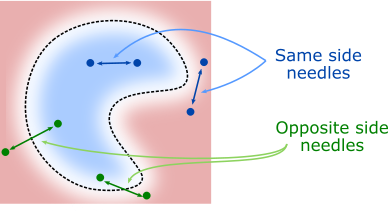}
    \caption{Needles as pairs of points.}
    \label{fig:pairs}
\end{figure}

\textbf{Needle droping for surface reconstruction.}
Inspired by this probability problem, we base our reconstruction strategy using needles (pairs of points), instead of points as in previous works.
Let $\vect{x}$ and $\vect{y}$ be the end points of a needle "dropped" in 3D space. We will construct our self-supervised training loss based on the facts that
if $\vect{x}$ and $\vect{y}$ lie on the same side of $\shape$, then $f_\shape(\vect{x})$ and $f_\shape(\vect{y})$ shall be equal, while if $\vect{x}$ and $\vect{y}$ lie on opposite sides of $\shape$, then $f_\shape(\vect{x})$ and $f_\shape(\vect{y})$ shall be different.
It is illustrated on Figure~\ref{fig:pairs}.

\subsubsection{Needle-based formulation}

Our goal is to estimate the ideal, but unknown, occupancy function $f^0_\shape$. As this function takes values in $\{0,1\}$, we can define a Bernoulli distribution $\bernoulli^0_\vect{x}$ at every position $\vect{x} \in \mathbb{R}^3$. Similarly, we define the Bernoulli distribution $\bernoulli_\vect{x}$ with parameter $\bprob_\vect{x} = f_\shape(\vect{x})$ at the same location $\vect{x} \in \rthree$.

To reconstruct the surface, our goal is now to minimise the distance between $\bernoulli_\vect{x}$ and $\bernoulli^0_\vect{x}$ at every location $\vect{x} \in \rthree$. But this problem is ll-posed because $\bernoulli^0_\vect{x}$ is unknown in our case. 
To overcome this issue, we propose to construct new random variables out of $\bernoulli_\vect{x}$ and $\bernoulli^0_\vect{x}$ for multiple pairs of different locations $\mathbf{x}$ and $\mathbf{y}$. Let $X$ and $Y$ be two independent Bernoulli variables at location $\mathbf{x}$ and $\mathbf{y}$ drawn according to  $\mathcal{B}_\mathbf{x}$ and $\mathcal{B}_\mathbf{y}$. The probability that $X=Y$ satisfies
%
\begin{equation}
    \bprob_{\mathbf{x}, \mathbf{y}} = \bprob_\mathbf{x} \, \bprob_\mathbf{y} + (1-\bprob_\mathbf{x})(1-\bprob_\mathbf{y}).
\end{equation}
We can thus define a new Bernoulli distribution $\mathcal{B}_{\mathbf{x}, \mathbf{y}}$ of parameter $\bprob_{\mathbf{x}, \mathbf{y}}$ for needle $(\mathbf{x},  \mathbf{y})$. We similarly define the target Bernoulli distribution $\mathcal{B}^0_{\mathbf{x}, \mathbf{y}}$ of parameter $\bprob^0_{\mathbf{x}, \mathbf{y}}$ using $\bernoulli^0_\vect{x}$. 
Then to estimate $\bprob^0_\mathbf{x}$, i.e., reconstruct the surface, we propose to minimise the binary
cross-entropy $\BCE$, i.e., the log-loss,  between $\bprob_{\vect{x}, \vect{y}}$ and $\bprob^0_{\vect{x}, \vect{y}}$
\begin{equation}
\label{eq:kl_point}
    \lossrecons(\mathbf{x}, \mathbf{y}) =
    \sum_{(\mathbf{x}_i, \mathbf{y}_i) \in Q} \BCE( \bprob_{\vect{x}_i, \vect{y}_i}, \bprob^0_{\vect{x}_i, \vect{y}_i}),
\end{equation}
for a finite set of well-chosen needles 
\begin{equation}
Q = \{(\mathbf{x}_i, \mathbf{y}_i) \, \vert \, i \in \{1, \ldots, \vert Q \vert \}\}.
\end{equation}
In practice, $\bernoulli^0_{\vect{x},\vect{y}}$ is unknown but 
it is enough to know if the needles $(\mathbf{x}_i, \mathbf{y}_i)$ traverse the surface or not to be able to compute $\lossrecons$ in Eq.\,\eqref{eq:kl_point}.
Inferring if a needle crosses the surface or not, when the needle is well chosen, is much easier than guessing if a point $\mathbf{x}$ is inside or outside the object.

\subsubsection{Needle picking}

We now describe our strategy to pick needles with end points
on opposite sides or on the same side of the surface.

\textbf{Surface-crossing needles (opposite-side end points).}
As the only information we have about the shape is $P_\shape$, and considering that the surface is continuous and locally planar in the neighborhood of $\vect{p} \in \pointcloud_\shape$, which is true at arbitrarily small scales, we drop likely surface-crossing needles as line segments centered at points of $\pointcloud_\shape$, thus with likely opposite-side end points.
The corresponding set $Q_\opposite$ is
\begin{equation}
    Q_\opposite = \{(\vect{p}+\vect{h}, \vect{p}-\vect{h}) \mid \vect{p} \in \pointcloud_\shape, \vect{h} \sim \normal(0,\sigma_h)\},
\end{equation}
where $\vect{h}$ is randomly sampled from the multivariate normal distribution $\normal(0,\sigma_h) \in \rthree$ with standard deviation $\sigma_h$, possibly depending on~$\vect{p}$.
The high probability of crossing the surface actually depends on curvature (see supp.\,mat.). Empirically, exceptions are rare enough not to confuse learning, as with datasets containing some erroneous labels.

\textbf{Non-surface-crossing needles (same-side end points).}
We also have to sample needles not crossing the surface. We may note that if a 3D point $\vect{p}$ is not too close to $\pointcloud_\shape$ and if $\pointcloud_\shape$ is dense enough (which is the case in practice for ordinary objects sampled with as few as 300 points), 
then it is likely that the line segment between $\vect{p}$ and its closest point $\vect{p}'$ in $\pointcloud_\shape$ does not cross the surface $\shape$, unless possibly near to $\vect{p}'$, including due to noise in input point cloud sampling. To simply take that into account, we consider closest points to $\pointcloud_\opposite$, where $\pointcloud_\opposite$ is the set of end points in $Q_\opposite$, i.e., points of $\pointcloud_\shape$ with a slight offset in two opposite directions. 
To create short-enough needles that are unlikely to cross the surface and add variability in needle orientation w.r.t.\ the surface, we actually consider the following set of needles:
\begin{equation}
Q_\same = \{(\vect{p},\vect{p'}) \mid \vect{p} \,{\in}\, \pointcloud_\same, \vect{p}' \,{=}\, \mathrm{nn}(\vect{p},\pointcloud_\opposite \,{\cup}\, \pointcloud_\same)\}.
\end{equation}
where $\pointcloud_\same$ are points sampled in 3D space and $\mathrm{nn}(\vect{p},\pointcloud)$ is the nearest neighbor of $\vect{p}$ in $\pointcloud$ (excluding $\vect{p}$ itself).

\textbf{Reconstruction loss.}
It would be natural to define the loss on $Q = Q_\opposite \cup Q_\same$ in Eq.~\eqref{eq:kl_point}. However, to account for possibly different sizes of $Q_\opposite$ and $Q_\same$, we apply Eq.~\eqref{eq:kl_point} independently on both sets. The loss is thus composed of two terms: a ``data'' term $\lossopp$ defined on $Q_\opposite$ that enforces the surface to be located near $\pointcloud_\shape$ and a ``regularization'' term $\losssame$ ensuring side label consistency inside/outside the shape. The reconstruction loss satisfies
\begin{equation}
\label{eq:loss}
    \lossrecons = \lossopp + \losssame
\end{equation}
where
\begin{eqnarray}
    \lossopp & = & \frac{1}{|Q_\opposite|}\sum_{(\vect{x}, \vect{y})\in Q_\opposite} \BCE(\bprob_{\vect{x}, \vect{y}}, \bprob^0_{\vect{x}, \vect{y}}),\\
    \losssame & = & \frac{1}{|Q_\same|}\sum_{(\vect{x}, \vect{y})\in Q_\same}
    \BCE(\bprob_{\vect{x}, \vect{y}}, \bprob^0_{\vect{x}, \vect{y}}).
\end{eqnarray}
Our loss, as \cite{gropp2020implicit, atzmon2020sal, atzmon2021sald, cai2020learning}, does not ensure an occupied shape interior and void remaining space; the reverse is a solution too, yet yielding the same surface. Due to non determinism, retraining on the same data may lead to either solution. Still, emptyness can be imposed at bounding box boundary.

\subsection{Learning shape representations}

From a collection of shapes provided only as point clouds, we learn a neural network $\network$.  Given a new point cloud $\pointcloud_\shape$ as input, $\network$ predicts an occupancy function $f_\shape$:
\begin{equation}\label{eq:net}
    f_\shape(\mathbf{x}) = S \circ \network(\pointcloud_\shape, \mathbf{x})
\end{equation}
where $S$ is the sigmoid function to ensure $f_\shape(\mathbf{x})\,{\in}\,[0,1]$.

\textbf{Training Loss.}
Substituting Eq.~\eqref{eq:net} in the expression of the log-loss in Eq.~\eqref{eq:kl_point} yields:
\begin{alignat}{3}
& \BCE( \bprob_{\vect{x}, \vect{y}}, \bprob^0_{\vect{x}, \vect{y}})
&&\!=\!
&&\log(e^{\network(\pointcloud_\shape, \vect{x})}+1)+ \log(e^{\network(\pointcloud_\shape, \vect{y})}+1) \nonumber\\
& &&
&&\!-\!\bprob_{\mathbf{x}, \mathbf{y}}^0 \log(e^{\network(\pointcloud_\shape, \vect{x})+\network(\pointcloud_\shape, \vect{y})}\!+\!1) \\
& &&
&&\!-\!(1\!-\!\bprob_{\mathbf{x}, \mathbf{y}}^0)\log(e^{\network(\pointcloud_\shape, \vect{x})}\!+\!e^{\network(\pointcloud_\shape, \vect{y})})\nonumber
\end{alignat}
which can be further simplified by exploiting the fact that $\bprob_{\vect{x},\vect{y}}^0 \,{=}\, 1$ when $ (\vect{x},\vect{y}) \,{\in}\, Q_\same$, and $\bprob_{\vect{x},\vect{y}}^0 \,{=}\, 0$ when $(\vect{x},\vect{y}) \,{\in}\, Q_\opposite$.
In addition, the gradient expression (see supp. mat.) is simple, benefiting from simplifications similarly to the usual BCE loss for singletons.
\section{Experiments}
\label{sec:experiments}

We conduct experiments on different datasets to validate the loss we propose and its use for surface reconstruction.
We first describe the network used in our experiments and the experimental setup.
Then, we evaluate our ability to reconstruct shapes from point clouds sampled on the entire surface (section~\ref{sec:exp:benchmarks}).
Finally, we show that our method can be applied to real point clouds from automotive datasets, either by direct transfer of previously learned models or by training from scratch, justifying our ability to work on highly sparse point clouds (section~\ref{sec:exp:real}).

\textbf{Network.}
We use the encoder/decoder network from \cite{mescheder2019occupancy}.
The encoder is a PointNet~\cite{qi2017pointnet} with 5 residual blocks.
The decoder is a fully connected network conditioned via batch normalization on the latent code~\cite{de2017modulating,dumoulin2016adversarially}.

\textbf{Parameters.}
Our method works both with sparse and dense point clouds; yet it is in sparse regime that it makes a difference.
For all experiments, unless otherwise stated, we use a point cloud size $|\pointcloud_\shape|=|Q_\opposite|=300$ and we set $|Q_\same|=2048$.
The parameter $\sigma_h$ used to draw the vectors $\vect{h}$ in the construction of $Q_\opposite$ is critical.
A large $\sigma_h$ distributes well the end points of both sides of the surface, increasing the chances that no needle in $Q_\same$ actually crosses the surface. However, too large a $\sigma_h$ may lead to needles crossing the surface twice, which we assume does not occurs in our construction of the training loss. On the contrary, too small a $\sigma_h$ reduces the influence of $\lossopp$ to a small neighborhood around the points in $\pointcloud_\shape$, preventing a good surface coverage and also increasing the chance of having needles in $Q_\same$ that accidentally cross the surface. As a rule of thumb, we set $\sigma_h(\vect{p}) = d_\vect{p}/3$, where $d_\vect{p}$ is the distance between a point $\vect{p}$ to its nearest neighbor in $\pointcloud_\shape$.

\textbf{Training procedure.}
We train our model in an end-to-end fashion, with Adam \cite{kingma2017adam} and a learning rate of $5*10^{-4}$.
During training, the points are either randomly sampled on the shape if an input mesh is available or picked in the original point cloud, e.g., for lidar scenes from KITTI.
At test time, we predict the occupancy values on a grid and use a Marching Cubes algorithm~\cite{lorensen1987marching} for shape extraction.

\textbf{Metrics and dataset pre-processing.}
For each experiment, and in order to compare with previous works, we use the same dataset, same pre-processing and same metrics as used in the compared papers,
as mentioned in each table caption (IoU, $\ell_1$ and $\ell_2$ Chamfer distance, worst Chamfer distance of the best $x$\% of points).

\begin{table}
\centering
\setlength{\tabcolsep}{3pt}
\begin{tabular}{lc|cc|cc}
\toprule
&&\multicolumn{2}{c|}{Supervision} & \multicolumn{2}{c}{Chamfer $\ell_2$ $\downarrow$} \\
&& Dist. & Sign & Mean & Median\\
\midrule
\emph{Input: \rlap{300 points}}                            & & -   &   -   &6.649 &6.441 \\
\midrule
PSGN~\cite{fan2017point}                & Points  & \cmark & \xmark &1.986 &1.649 \\
DMC~\cite{liao2018deep}                 & Explicit& \cmark & \cmark &2.417 &0.973 \\
OccNet~\cite{mescheder2019occupancy}    & Implicit& \xmark &\cmark &1.009 &0.590 \\
IF-Net~\cite{chibane2020implicit}       & Implicit& \xmark &\cmark &1.147 &0.482 \\
NDF~\cite{chibane2020ndf}               & Implicit& \cmark &\xmark &0.626 &0.371 \\
\midrule
{\method}          & Implicit&\xmark &\xmark & 1.703 & 1.109\\
{\method}$^*$          & Implicit&\xmark &\xmark & 1.575 & 0.952\\ 
{\method}-FP  & Implicit&\xmark &\xmark & 2.461 & 1.897 \\
\bottomrule
\end{tabular}

{\small (a) ShapeNet cars, closed meshes \cite{NEURIPS2019_39059724}, results $\times 10^{-4}$, cf.~\cite{chibane2020ndf}.}

\vspace{0.15cm}
\setlength{\tabcolsep}{2pt}
\begin{tabular}{lc|cc|cc}
\toprule
&& \multicolumn{2}{c|}{Supervision} & \multicolumn{2}{c}{Eval.\ metrics}\\
\emph{Input: \rlap{300 pts+noise}}& & Dist.& Sign & IoU $\uparrow$ & Cha. $\ell_1$ $\downarrow$\\
\midrule
3D-R2N2~\cite{choy20163d} &Voxels&\xmark& \cmark& 0.565 & 0.169 \\
PSGN~\cite{fan2017point} &Points&\cmark & \xmark& - & 0.144 \\
ShapeGF$^\mathsection$~\cite{cai2020learning} & Points & \xmark & \xmark & - & 0.083 \\
DMC~\cite{liao2018deep} & Explicit& \cmark & \cmark& 0.674 & 0.117 \\
OccNet~\cite{mescheder2019occupancy} & Implicit&\xmark & \cmark& 0.778 & 0.079\\
\midrule
{\method} & Implicit & \xmark & \xmark & 0.666 & 0.112 \\
{\method}$^*$ & Implicit & \xmark & \xmark &0.675 & 0.111\\
{\method}-FP & Implicit & \xmark & \xmark  & 0.669 & 0.106 \\
\bottomrule
\end{tabular}

{\small (b) ShapeNet subset of\cite{choy20163d}, all classes, results $\times 10^{-1}$, cf.~\cite{mescheder2019occupancy}.}

\vspace{0.15cm}
\setlength{\tabcolsep}{3pt}
\begin{tabular}{@{}l@{}cc|c|ccc@{}}
\toprule
                &\multicolumn{2}{c|}{No.\ points}& Superv. & \multicolumn{3}{c}{Chamfer $\ell_2$ $\downarrow$}\\
Method          & Train & Test & Dist. & 5\%   & 50\%& 95\%  \\
\midrule
DeepSDF$^\dagger$~\cite{park2019deepsdf} &16k & 16k&\cmark& 3.45  & 45.03 & 294.15 \\
DeepSDF$^\ddagger$~\cite{park2019deepsdf} &16k & 16k&\cmark& 1.88  & 31.05 & 489.35 \\
AtlasNet~\cite{groueix2018papier} & 16k & 16k & \cmark &  0.10  & 0.17  & 0.37  \\
SAL~\cite{atzmon2020sal} & 16k & 16k & \cmark& 0.07  & 0.12  & 0.35  \\
IGR~\cite{gropp2020implicit} & 8k & 16k & \cmark &4.79 & 12.04 & 104.08 \\
\midrule
SAL$^\mathsection$~\cite{atzmon2020sal}   &   300 &   300 & \cmark & 0.086 & 0.134 & 0.421 \\
SAL$^\mathsection$~\cite{atzmon2020sal}   &   300 &   - & \xmark & \multicolumn{3}{l}{\emph{Failed to converge}} \\
IGR$^\mathsection$~\cite{gropp2020implicit}   &   300 &   - & \xmark & \multicolumn{3}{l}{\emph{Failed to converge}} \\
\midrule
{\method}  & 300 & 300 & \xmark & 0.269 & 0.433 & 1.149 \\
{\method}$^*$  & 300 & 300 & \xmark & 0.107 & 0.175 & 0.811 \\
{\method}-FP& 300 & 300 & \xmark & 0.202 & 0.526 & 1.322 \\
\bottomrule
\multicolumn{7}{l}{\makecell[l]{\scriptsize Here, \cmark denotes true distance to shape or distance computation from dense input}}

\end{tabular}

{\small (c) DFaust, results $\times 10^{-3}$, cf.~\cite{atzmon2020sal}.}

\vspace{-2mm}
\begin{justify}
\scriptsize
\begin{itemize}[nosep,labelsep=2pt]
\item[$^{\dagger \ddagger}$:] Implemented in \cite{atzmon2020sal}, where the sign of the distance is computed from the oriented normals provided with the scans ($^\dagger$) or locally with Jets~\cite{cazals2005estimating} followed by a consistent orientation based on minimal spanning trees ($^\ddagger$).
\item[$^\mathsection$:] Trained by ourselves using original code made available.
\item[$^*\vphantom{^\mathsection}$:] Model finetuned with $\sigma_\vect{h}/2$ from the initial model.
\item[-FP:] Model trained on the same (not re-sampled) points at each epoch.
\end{itemize}
\end{justify}
\vspace*{-2.5mm}
\caption{Reconstruction from complete point clouds.}
\vspace*{-3mm}
\label{tab:exp_complete}
\end{table}

\subsection{Synthetic point clouds}
\label{sec:exp:benchmarks}

\begin{figure*}

    \centering
    
    \setlength{\tabcolsep}{5pt}
    \renewcommand{\arraystretch}{0.2}
    \begin{tabular}{c|c|c|ccc}
    & & \emph{Supervised} & \multicolumn{3}{c}{\emph{Self-supervised}} \\
    \includegraphics[trim={0 10 0 0},clip,width=0.135\linewidth]{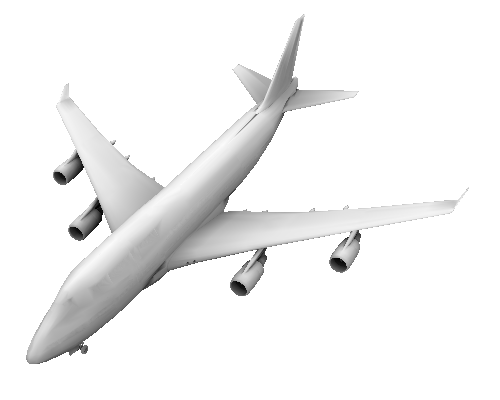}&
    \includegraphics[trim={0 10 0 0},clip,width=0.135\linewidth]{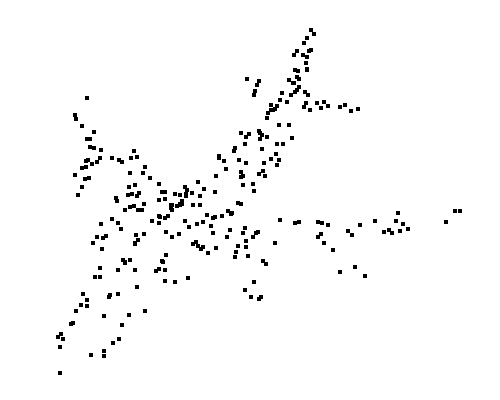}&
    \includegraphics[trim={0 10 0 0},clip,width=0.135\linewidth]{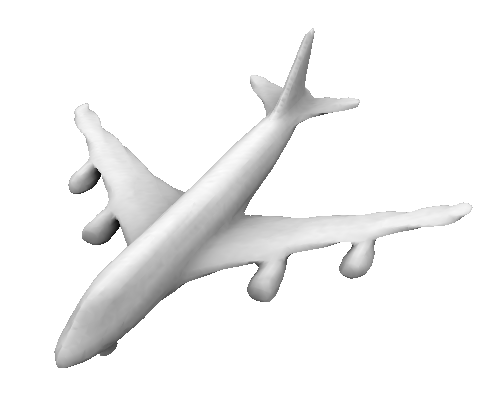}&
    \includegraphics[trim={0 10 0 0},clip,width=0.135\linewidth]{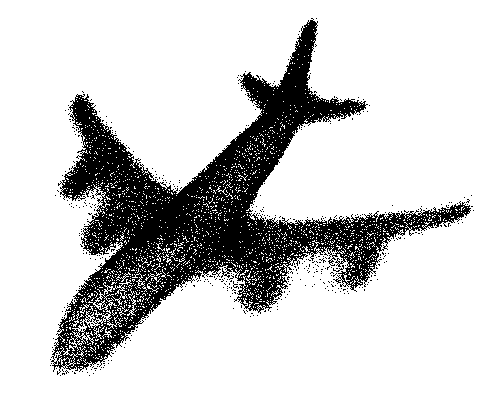}&
    \includegraphics[trim={0 10 0 0},clip,width=0.135\linewidth]{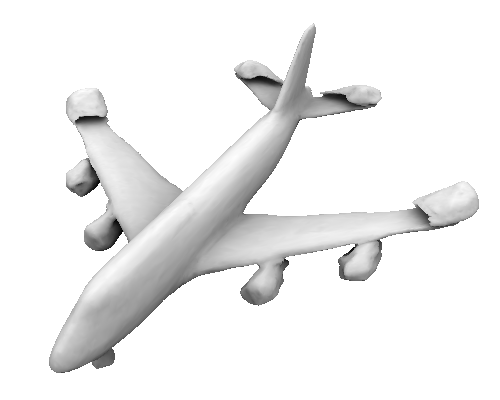}&
    \includegraphics[trim={0 10 0 0},clip,width=0.135\linewidth]{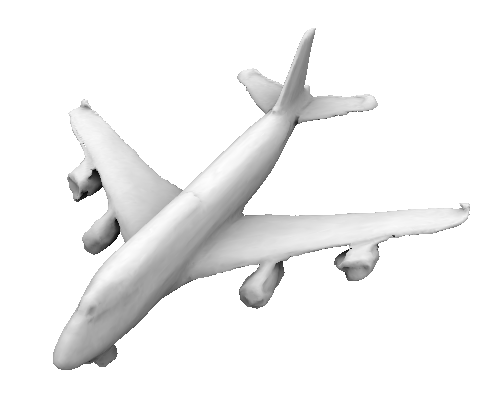}\\
    \includegraphics[trim={0 0 0 10},clip,width=0.135\linewidth]{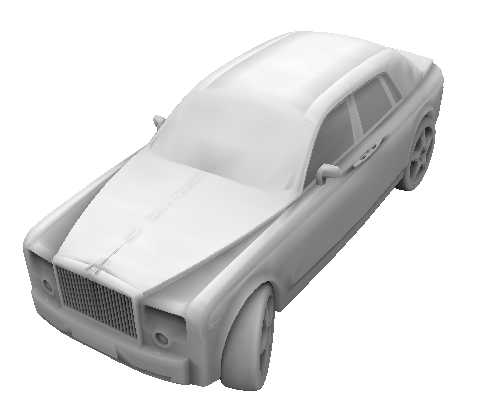}&
    \includegraphics[trim={0 0 0 10},clip,width=0.135\linewidth]{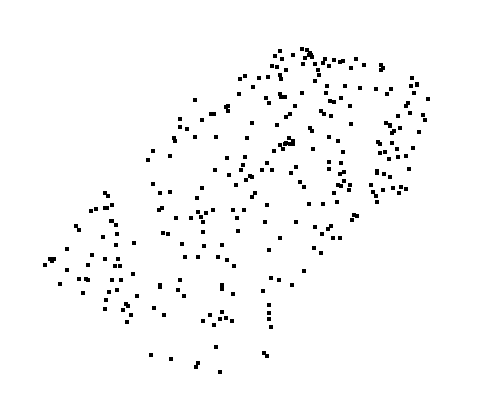}&
    \includegraphics[trim={0 0 0 10},clip,width=0.135\linewidth]{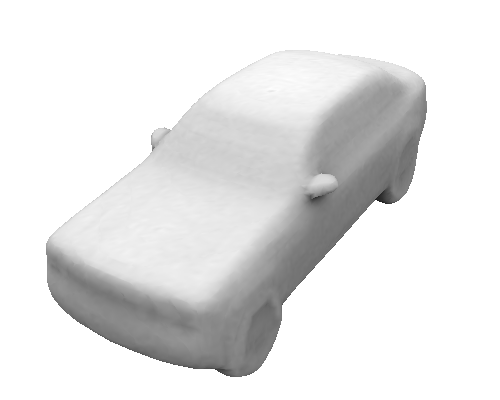}&
    \includegraphics[trim={0 0 0 10},clip,width=0.135\linewidth]{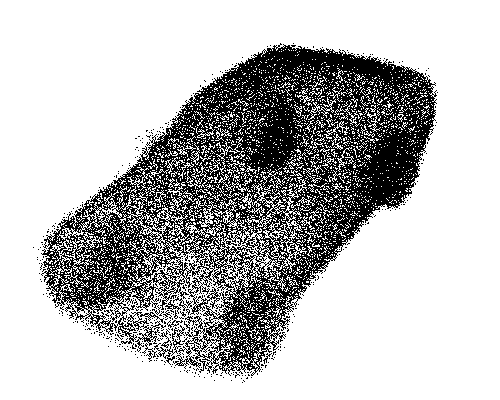}&
    \includegraphics[trim={0 0 0 10},clip,width=0.135\linewidth]{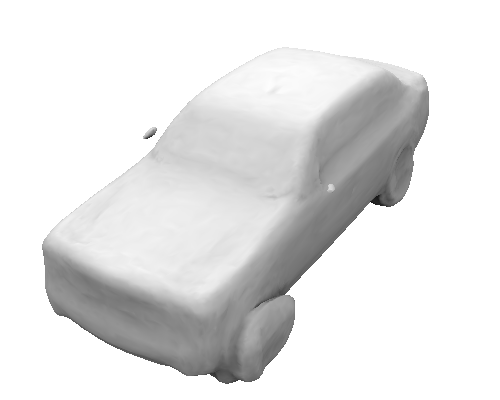}&
    \includegraphics[trim={0 0 0 10},clip,width=0.135\linewidth]{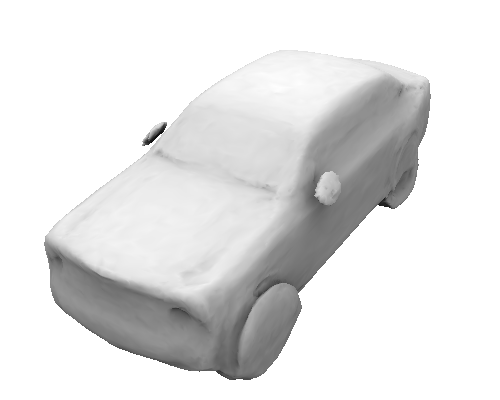}\\
    \includegraphics[width=0.135\linewidth]{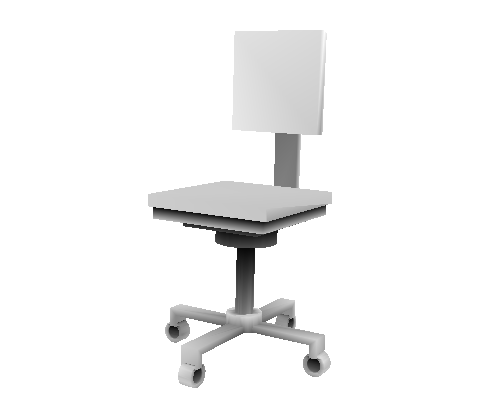}&
    \includegraphics[width=0.135\linewidth]{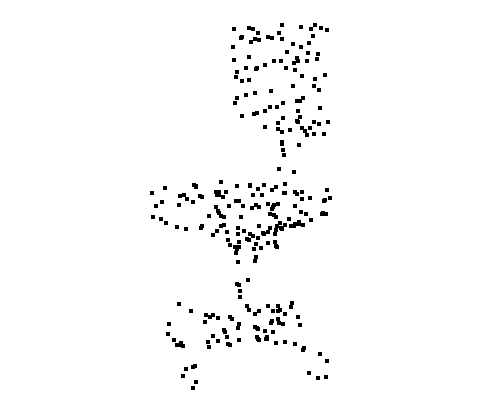}&
    \includegraphics[width=0.135\linewidth]{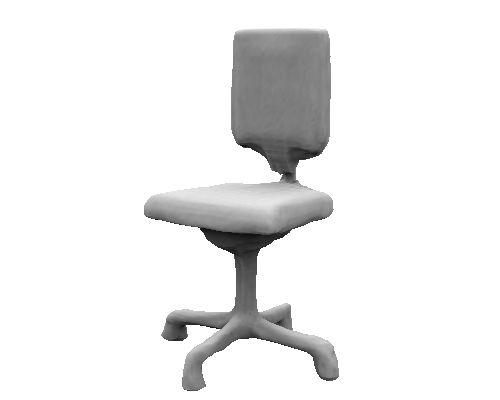}&
    \includegraphics[width=0.135\linewidth]{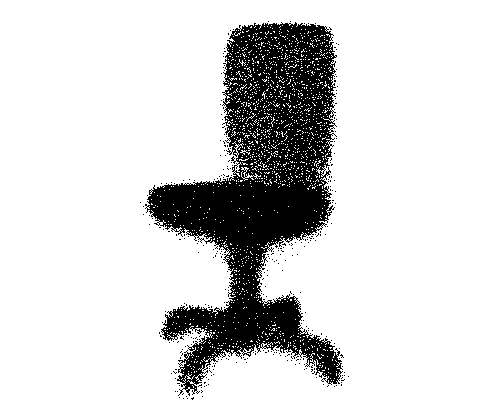}&
    \includegraphics[width=0.135\linewidth]{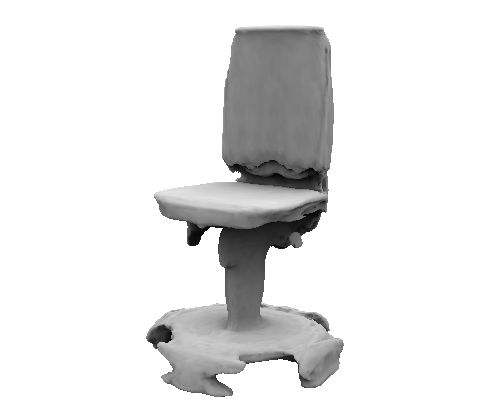}&
    \includegraphics[width=0.135\linewidth]{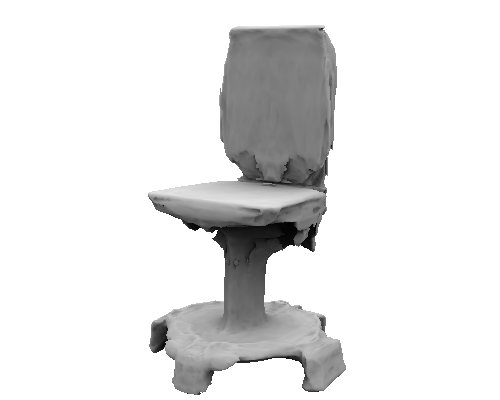}\\
    Ground truth & Input & ONet & ShapeGF & NeeDrop & NeeDrop*
    \end{tabular}
    
    ~\\
    (a) ShapeNet
    
    ~\\
    \begin{tabular}{ccc|c|ccc}
        \includegraphics[width=0.12\linewidth]{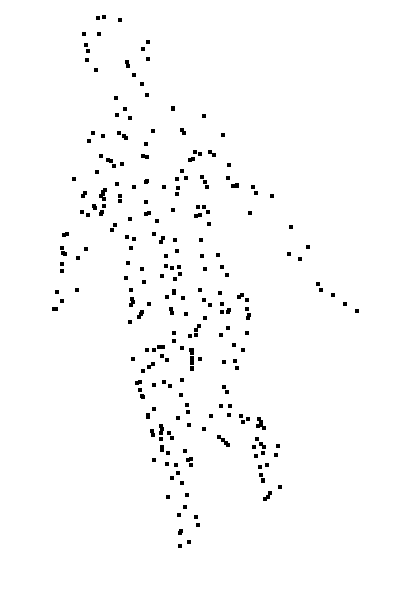}&
        \includegraphics[width=0.12\linewidth]{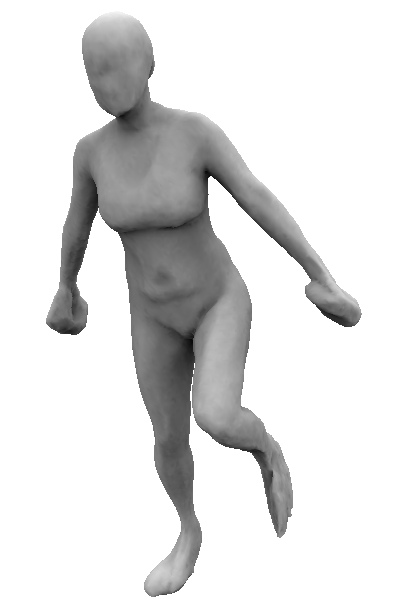}&
        \includegraphics[width=0.12\linewidth]{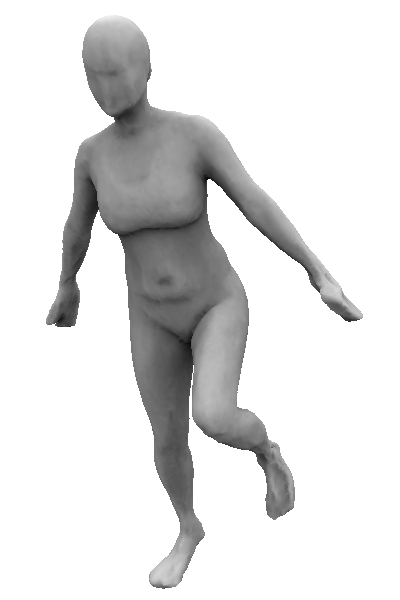}&
        \includegraphics[width=0.12\linewidth]{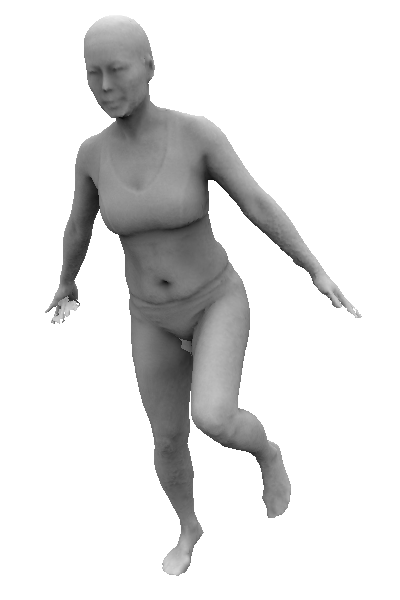}&
        \includegraphics[width=0.12\linewidth]{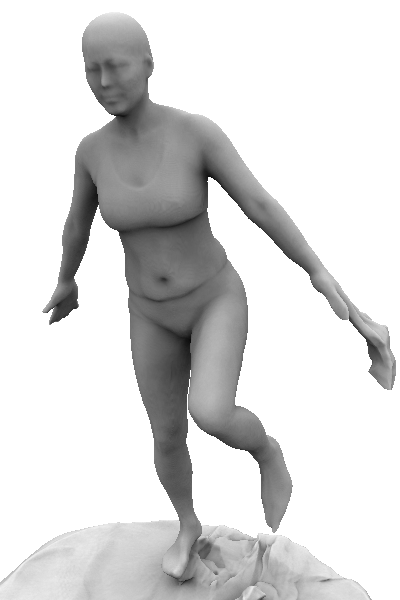}&
        \includegraphics[width=0.12\linewidth]{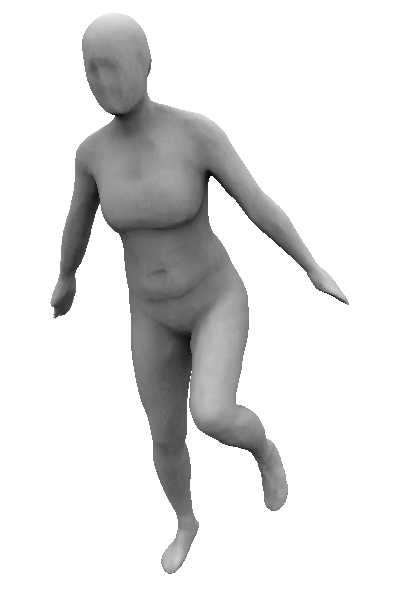}&
        \includegraphics[width=0.12\linewidth]{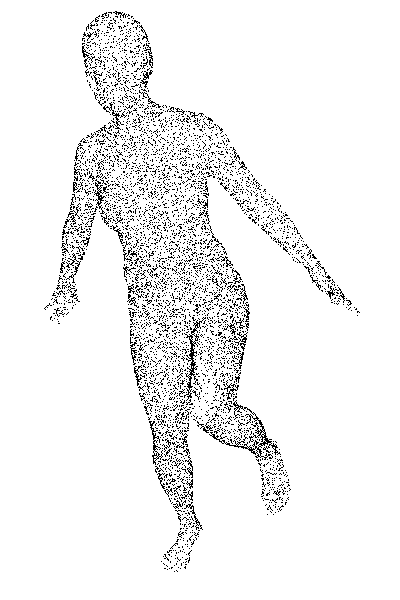}\\
        \includegraphics[trim={0 20 0 0},clip,width=0.12\linewidth]{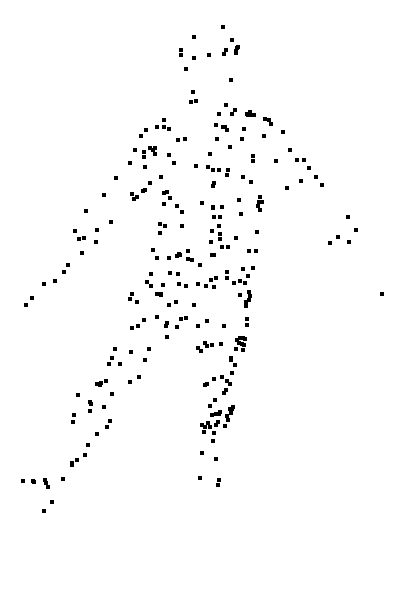}&
        \includegraphics[trim={0 20 0 0},clip,width=0.12\linewidth]{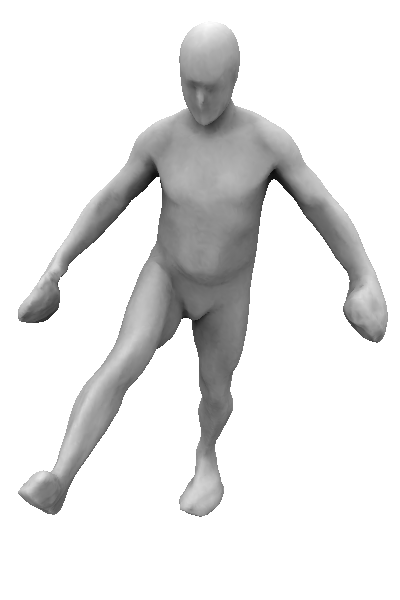}&
        \includegraphics[trim={0 20 0 0},clip,width=0.12\linewidth]{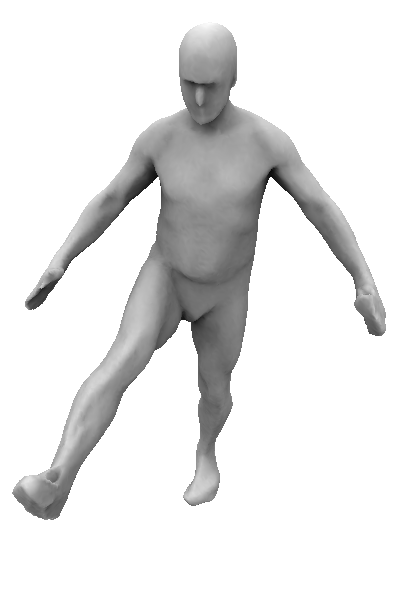}&
        \includegraphics[trim={0 20 0 0},clip,width=0.12\linewidth]{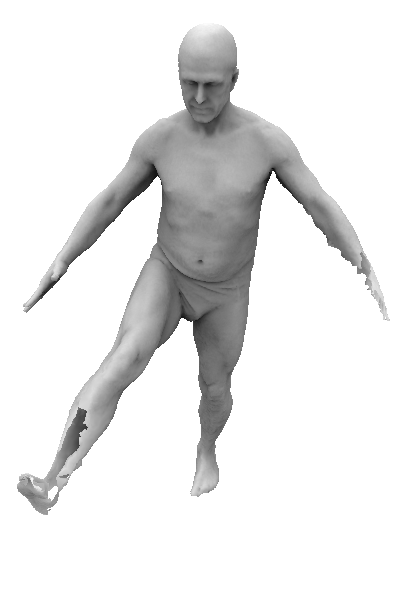}&
        \includegraphics[trim={0 20 0 0},clip,width=0.12\linewidth]{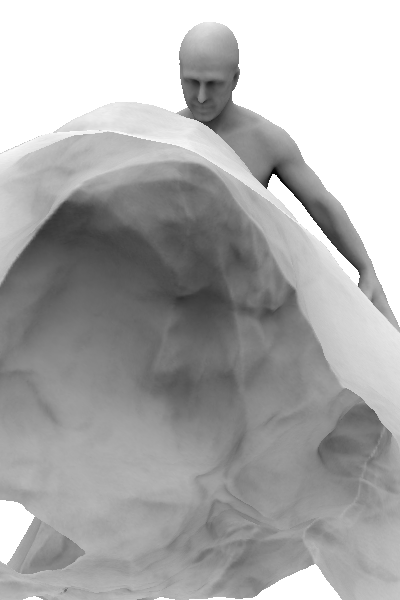}&
        \includegraphics[trim={0 20 0 0},clip,width=0.12\linewidth]{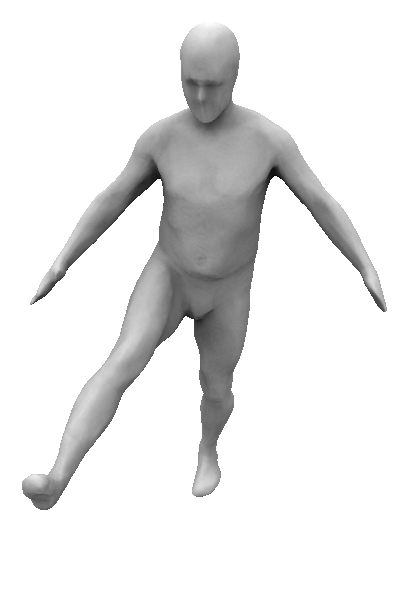}&
        \includegraphics[trim={0 20 0 0},clip,width=0.12\linewidth]{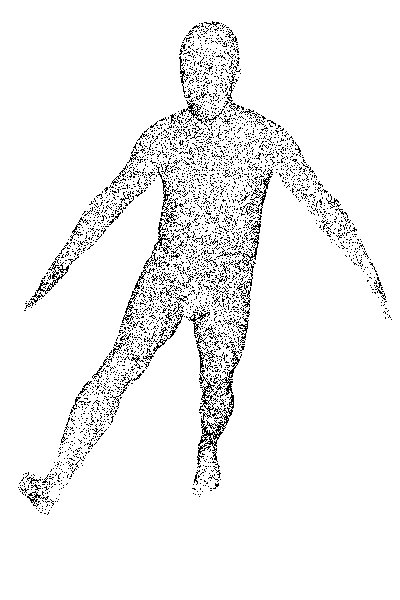}\\
        \small {\method} input &
        \small {\method} &
        \small {\method}$^*$ &
        \small Ground truth &
        \small IGR &
        \small SAL &
        \small SAL, IGR input\\
    \end{tabular}
    
    ~\\
    (b) DFaust.

\setlength{\tabcolsep}{8pt}
\begin{tabular}{ccccc|c}
\includegraphics[trim={0 30 0 10},clip, width=0.13\linewidth]{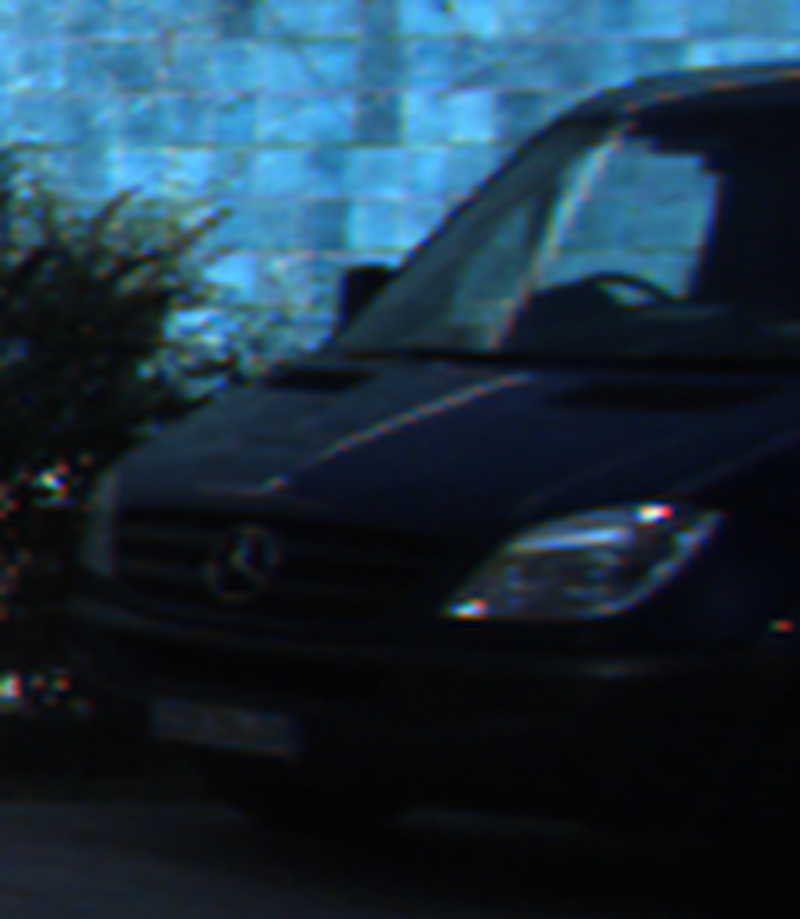}&
\includegraphics[width=0.13\linewidth]{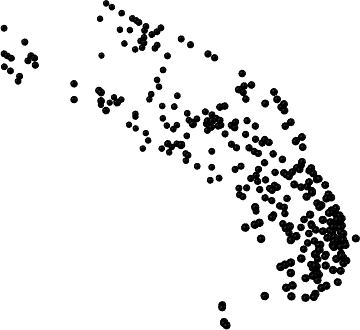}&
\includegraphics[trim={0 15 0 25},clip,width=0.13\linewidth]{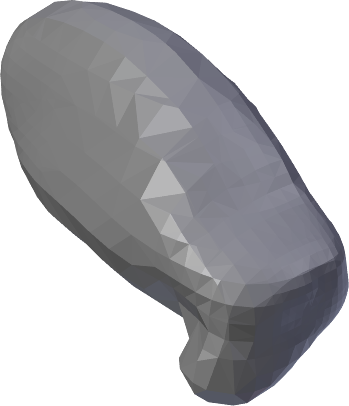}&
\includegraphics[width=0.13\linewidth]{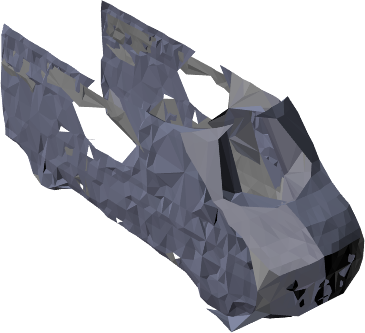}&
\includegraphics[width=0.13\linewidth]{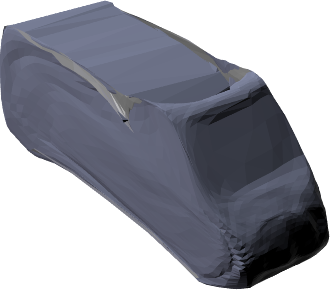}&
\includegraphics[width=0.13\linewidth]{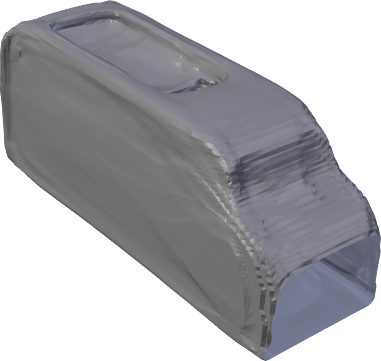}
\\
\includegraphics[width=0.13\linewidth]{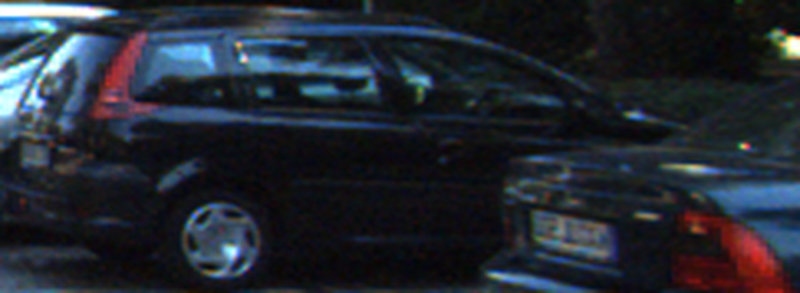}&
\includegraphics[width=0.13\linewidth]{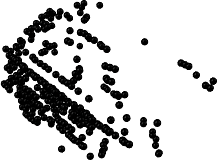}&
\includegraphics[width=0.13\linewidth]{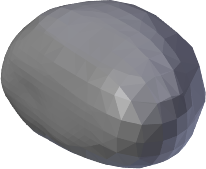}&
\includegraphics[width=0.13\linewidth]{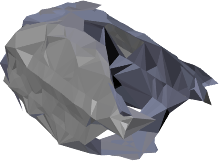}&
\includegraphics[width=0.13\linewidth]{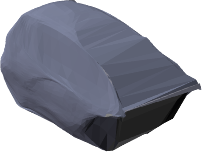}&
\includegraphics[width=0.13\linewidth]{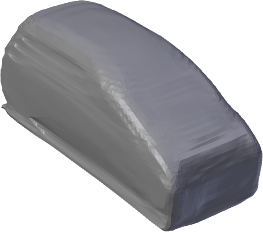}
\\
\includegraphics[width=0.13\linewidth]{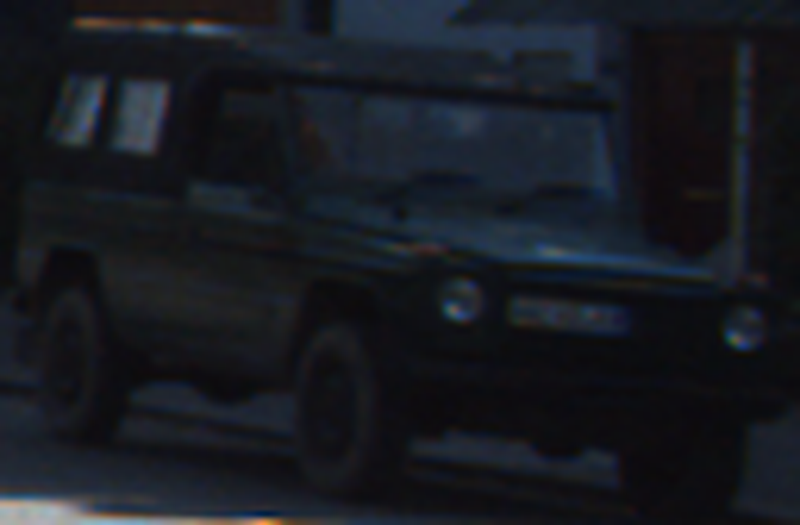}&
\includegraphics[width=0.13\linewidth]{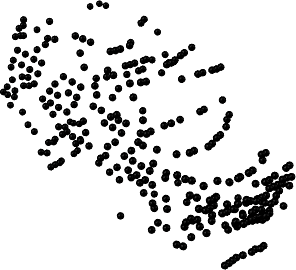}&
\includegraphics[width=0.13\linewidth]{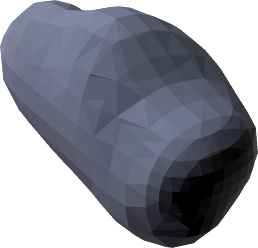}&
\includegraphics[width=0.13\linewidth]{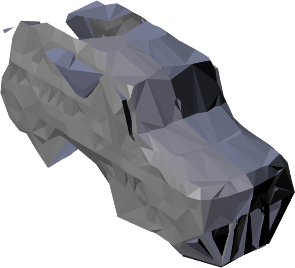}&
\includegraphics[width=0.13\linewidth]{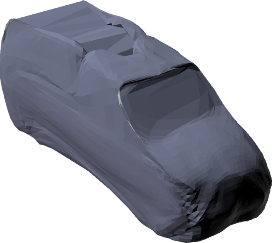}&
\includegraphics[width=0.13\linewidth]{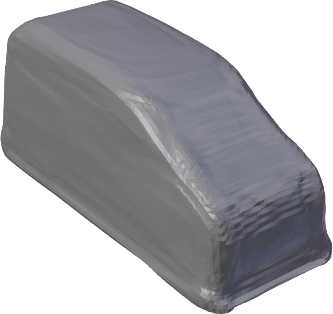}
\\
Ref image &
Input&
Poisson&
Ball &
AtlasNet&
\multicolumn{1}{c}{{\method}}\\
&&& Pivoting && 
\multicolumn{1}{c}{(Kitti training)}\\

\end{tabular}

(c) Comparison of the reconstruction on KITTI point clouds.

\caption{Qualitative comparison of {\method} to state-of-the-art methods, in ShapeNet, DFaust and KITTI.}
\label{fig:sigma}
\end{figure*}

To compare with state-of-the art methods, we evaluate our self-supervised formulation, named {\method}, on point clouds sampled uniformly on closed shapes.
Tab.\,\ref{tab:exp_complete} and Fig.\,\ref{fig:sigma} 
show our results on ShapeNet~\cite{chang2015shapenet} and DFaust~\cite{dfaust:CVPR:2017}.

\textbf{Evaluation on ShapeNet.}
We train our model on two configurations of ShapeNet.
We first present results on the car subset (Tab.\,\ref{tab:exp_complete}(a)) as in~\cite{chibane2020implicit,chibane2020ndf}, where the raw meshes are closed using the pre-processing of~\cite{NEURIPS2019_39059724}.

In Tab.\,\ref{tab:exp_complete}(b), we extend the experiment to all ShapeNet categories of~\cite{mescheder2019occupancy}; qualitative examples are presented in Fig.\,\ref{fig:sigma}(a).
For each compared method, we highlight its level of supervision, on distance and/or sign/occupancy.
Unlike {\method}, all other methods (except ShapeGF) have a certain level of supervision.
Despite the absence of supervision in {\method}, we obtain competitive results on both benchmarks. We are on par with DMC~\cite{liao2018deep}, and only outperformed by OccNet~\cite{mescheder2019occupancy} among those methods.
ShapeGF has a lower Chamfer distance but outputs noisy point clouds (Fig.\,\ref{fig:sigma}(a)) from which constructing a mesh in not straightforward. 
In contrast, the occupancy predicted by {\method} allows for direct mesh extraction.
(Trying to compare with IGR~\cite{gropp2020implicit} on ShapeNet failed because we could not find parameters to reconstruct acceptable surfaces.)

\textbf{Evaluation on DFaust.}
When comparing on DFaust (Tab.\,\ref{tab:exp_complete}(c)), we include results for two variants of DeepSDF \cite{park2019deepsdf} reported in SAL~\cite{atzmon2020sal}. The problem is that DeepSDF learns from a signed distance to a closed shape, whereas DFaust meshes are open. In the first variant DeepSDF$^\dagger$, the sign of the distance is computed based on the oriented normals provided with the scans; in the second variant DeepSDF$^\ddagger$, the normals are estimated locally with Jets~\cite{cazals2005estimating} and oriented consistently using minimal spanning trees \cite{atzmon2020sal}.

All methods reported in~\cite{atzmon2020sal} were trained on very dense point cloud (16k points), while we target the much more difficult case of learning 
from low resolution point clouds. 
We thus also trained SAL~\cite{atzmon2020sal} with the same amount of point as {\method} (300 pts), with supervision on the distance to the shape as well as without supervision, i.e., replacing the distance to the surface by the distance to the point cloud.
Results in Tab.\,\ref{tab:exp_complete}(c) with identical training settings (batch size 32, 2048 query pts) show that, as long as distances are measured w.r.t.\ the real surface, the performance of SAL degrades gracefully when the number of training points goes down from 16k to 300. 
However, when the distance to the surface has to be approximated from the training points, SAL fails to produce meshes when training only on 300 pts (and even on 1k pts).
In fact, SAL may yield \emph{unsigned} distance functions, which are also solutions to the loss; Marching Cubes then produce no surface. The specific network initialization of SAL tries to make it less likely, but unsigned distances still seems to be preferred solutions for sparse point clouds without exact distance supervision.
The likely reason is that SAL solves an ill-posed problem: both the signed and the unsigned distance to the shape are solutions to the optimization problem. Obtaining a signed function is only favored by a particular initialization of the network, but not explicitly enforced during optimisation.
In the self-supervised scenario with sparse point clouds, no regularisation prevents the implicit function to change sign between two points, where the actual surface should be, as the actual distance to the surface is not available.
In contrast, \method\ enforces the surface to be supported by $\pointcloud_\shape$ and does not require any well-designed initialization.

SALD~\cite{atzmon2021sald} does not offer code. 
Yet we expect sensitivity to the regularization weight for sparse data as a higher value is required to connect distant points with an iso-surface.

For IGR~\cite{gropp2020implicit}, we ran an experiment with 300 points as input for training (default is 8k pts).
After a good training start, we observed divergence around epoch 200.
We also used the the pre-trained model (trained with 8k points, tested with 16k) provided by the authors.
IGR produces artifacts around the shape (see Fig.\,\ref{fig:sigma}), which explains its worse quantitative performance in terms of Chamfer distance. 
However, we observe that it is the model which qualitatively recovers the best the details of the shapes.

\textbf{Finetuning with a smaller $\sigma_\vect{h}$.}
This parameter is critical in our method.
Its default value ensures a good convergence for all datasets we experimented with.
Yet, small details may be lost with a too large $\sigma_\vect{h}$.
In the spirit of curriculum learning, we finetune on some epochs 
the model previously learned with $\sigma_\vect{h}/2$, (named {\method}$^*$ in Tab.\,\ref{tab:exp_complete}(c)) and observe a significant improvement.
Fig.\,\ref{fig:sigma} visually illustrates the improvement on a sample from the DFaust dataset.
With the initial model, hands and feet, i.e., thin surfaces, are estimated with surface blobs, which are reduced after finetuning.
We also observe the appearance of the belly button and the eyebrows. Besides, our reconstruction with 300 pts is almost on par with SAL with 16,384 pts. 
See the supp.\,mat.\ for a study of needle crossing validity for varying $\sigma_\vect{h}$.

\textbf{Fixed-point training.} For comparison purposes, we sample new points on each shape at each epoch as in previous work. Yet, we also test our method with the same fixed 300 points sampled on each shape ({\method}-FP in Tab.\,\ref{tab:exp_complete}). 
We observed that fixed-point training is stable. 
Besides, training with fixed points reaches almost the same performance as training with point re-sampling, except when the dataset is small (ShapeNet car subset vs the 13 ShapeNet classes); but even in this case, we still outperform a few supervised methods while not using exact distance or sign.

\subsection{Real point clouds}
\label{sec:exp:real}

\begin{figure}[t]
    \centering
    \setlength{\tabcolsep}{2pt}
    \begin{tabular}{cc@{\hspace{30pt}}cc}
    \includegraphics[width=0.2\linewidth]{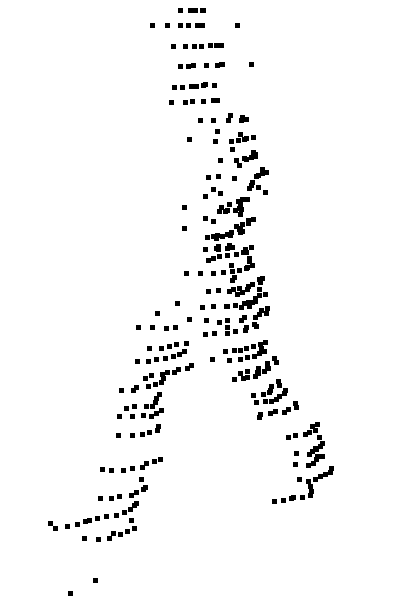}&
    \includegraphics[width=0.2\linewidth]{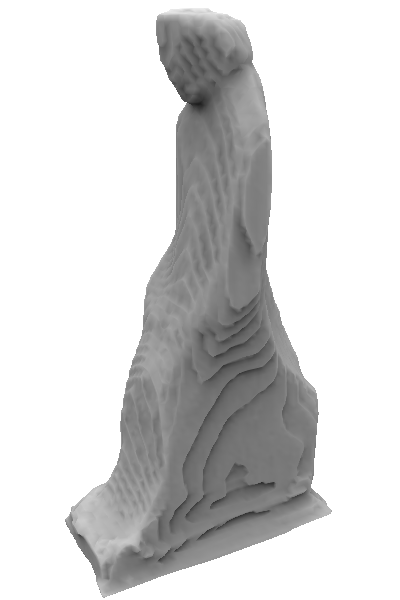}&
    \includegraphics[width=0.2\linewidth]{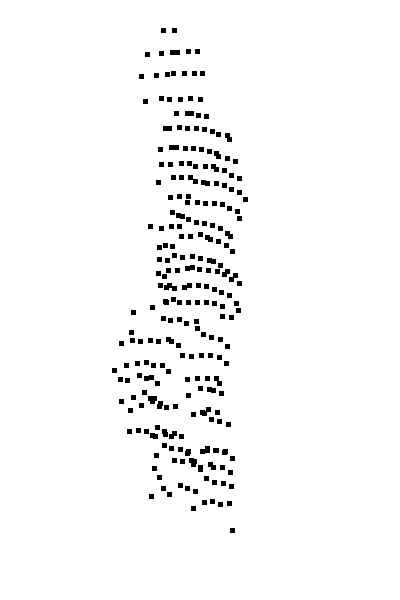}&
    \includegraphics[width=0.2\linewidth]{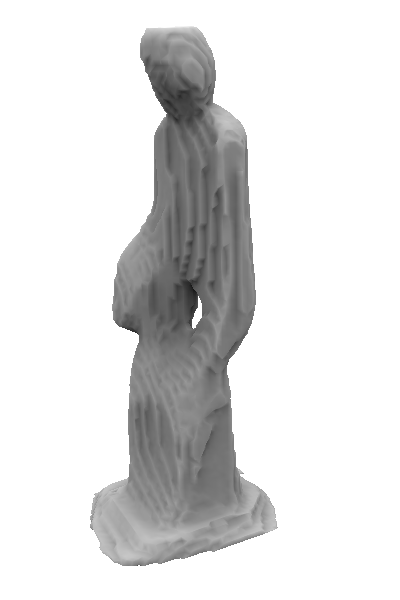}\\
    \includegraphics[width=0.2\linewidth]{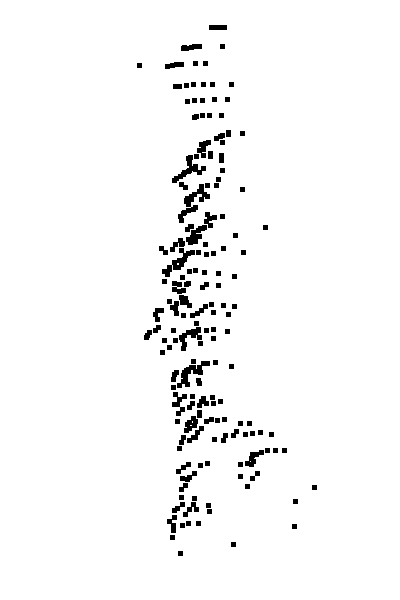}&
    \includegraphics[width=0.2\linewidth]{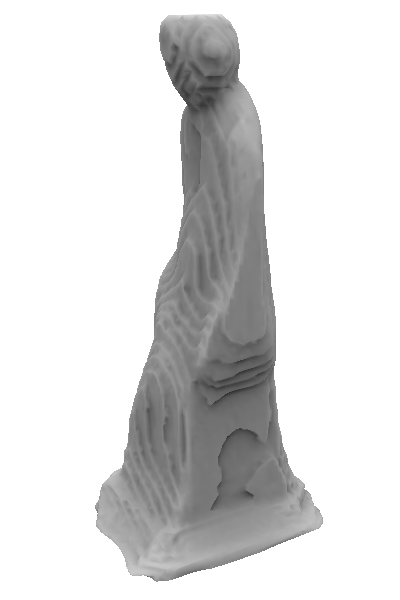}&
    \includegraphics[width=0.2\linewidth]{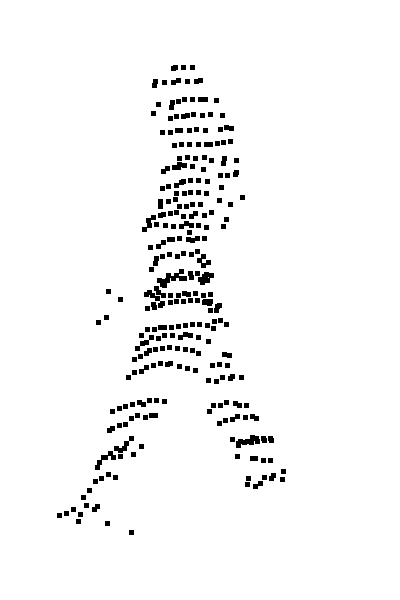}&
    \includegraphics[width=0.2\linewidth]{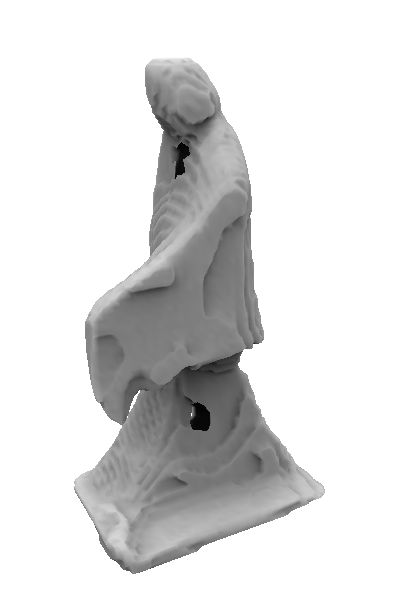}
    \end{tabular}
    \caption{KITTI pedestrians: input point cloud and reconstruction.}
    \label{fig:kitti_pedestrian}
\end{figure}

\textbf{Cars of KITTI.}
We perform a qualitative evaluation of the proposed method on the KITTI dataset. 
The input data are partial point clouds captured with lidars that either represent a car, a truck or a van, as extracted from the boxes of the detection dataset. 
In Figure~\ref{fig:sigma}(c), we present a comparison of our method against two direct (non-learned) methods, Poisson meshing~\cite{kazhdan2006poisson} and Ball Pivoting\cite{BallPivoting}, and a learning-based method, AtlasNet~\cite{groueix2018papier}. 
An image acquired by a RGB camera along with the point cloud is showed on  Figure~\ref{fig:sigma}(c) for illustration purposes, but it is not used as input. 
In this experiment, we take advantage of the car's plan of symmetry for all methods: as KITTI cars are given in 3D boxes, with frame origin on the box center and axis y along the car length, we use as input a symmetrized point cloud w.r.t.\ the plane 0yz.
For Poisson meshing and Ball Pivoting, reasonable hyper parameters were estimated through a grid search. AtlasNet was trained using a sphere primitive. 

As expected, the direct methods fail to provide a proper extrapolation of the shape far from the symmetrized input point cloud. 
While Poisson meshing provides a closed mesh, it barely contains any detail. 
Ball Pivoting however produces a detailed mesh around the input point cloud but it fails to reconstruct the hidden parts. AtlasNet is able to take advantage of the symmetry but fails when the front or back of the car is missing (see first and last row in Figure~\ref{fig:sigma}(c)). 
In constrast to these methods, {\method} is able to take advantage of the existing symmetry, to produce a realistic car shape, and to reconstruct hidden parts of the cars.

\textbf{KITTI pedestrians.}
As a last experiment on KITTI, we tried a very challenging target for shape reconstruction: pedestrians.
Unlike for cars, we cannot symmetrize the input point cloud and must learn directly on the raw point cloud. In addition, pedestrian shapes are much more diverse than cars, which are globally convex.
We see on Fig.\,\ref{fig:kitti_pedestrian} that, despite the task difficulty, our method is able to recover the coarse shape of a pedestrian.
Our failure to capture human limbs on KITTI is not due to $\sigma_\vect{h}$, as our method works well on DFaust. It is mostly due to partial views, that  cannot be symmetrized, unlike cars; it seems there are then many solutions, centered around merged limbs. Other factors are morphology variety, clothing and hand-carried items, while DFaust only features 10 different people in underwear.

\section{Conclusion}

In this study, we investigate self-supervised shape reconstruction and representation.
To this end, we consider ``needles'' dropped in 3D space around the shape, for which we are able to estimate if they cross the surface or not.
We also define a loss on these line segments suitable for neural network training.
The resulting approach, {\method}, is the first totally self-supervised approach for learning an implicit occupancy function from a collection of shape available only as sparse point clouds.
We show that our method is competitive with state-of-the-art supervised or partially supervised reconstruction methods.
We conduct qualitative comparisons on datasets with ground-truth meshes and qualitative experiments on the challenging KITTI dataset.
We successfully reconstruct plausible shapes from partial point cloud of cars and show promising results on the very challenging pedestrian category.


{\small
\bibliographystyle{plain}
\bibliography{biblio}
}

\clearpage

\section*{Supplementary material}

\appendix
\setcounter{equation}{14}
\setcounter{figure}{4}
\setcounter{table}{1}

We provide here complementary information about the paper ``NeeDrop: Self-supervised Shape Representation from  Sparse Point Clouds using Needle Dropping'':
\begin{itemize}[topsep=3pt,itemsep=-1pt]
\item[\ref{supp:sec:network}.] We present the network architecture used in our experiments, as well as
the training details.

\item[\ref{supp:sec:lossderiv}.] We provide a complete derivation of the loss expression defined in the main paper.

\item[\ref{supp:sec:latent}.] We analyze the latent space of shape representations through shape generation, shape interpolation, latent space visualization and classification.

\item[\ref{supp:sec:needles}.] We study the validity of the needle dropping construction hypothesis.

\item[\ref{supp:sec:sampl}.] We analyze the impact of two point-sampling strategies: with or without resampling.

\item[\ref{supp:sec:shapegf}.] We compare the quality of our reconstructions with ShapeGF, that generates point clouds rather than meshes.

\item[\ref{supp:sec:noise}.] We evaluate the robustness of our trained models to input noise and to variations of numbers of input points.

\item[\ref{supp:sec:transfer}.] We present additional results on the KITTI dataset, including reconstruction with networks trained on ShapeNet to test the robustness to domain gaps.

\item[\ref{supp:sec:loss}.] We give pseudo-code for the loss defined in the paper.

\end{itemize}

\emph{Note: reference numbers for citations used here are not the same as references used in the main paper; they correspond to the bibliography section at the end of this supplementary material.}

\begin{figure}
    \centering
    \setlength{\tabcolsep}{1pt}
    \begin{tabular}{c}
    \includegraphics[width=0.8\linewidth]{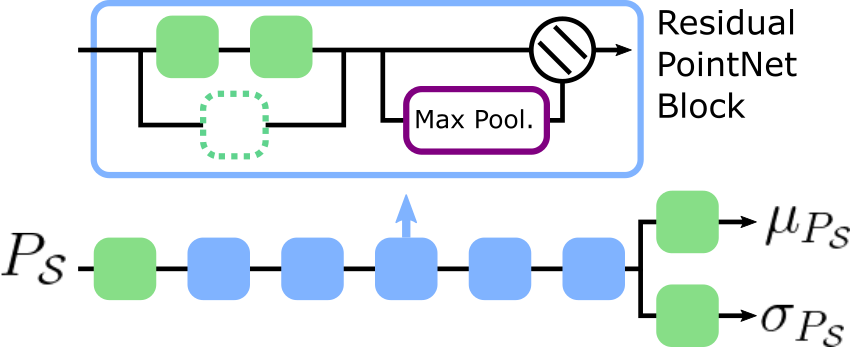}\\
    (a) Encoder 
    \vspace*{3mm} \\
    ~\\
    \includegraphics[width=0.8\linewidth]{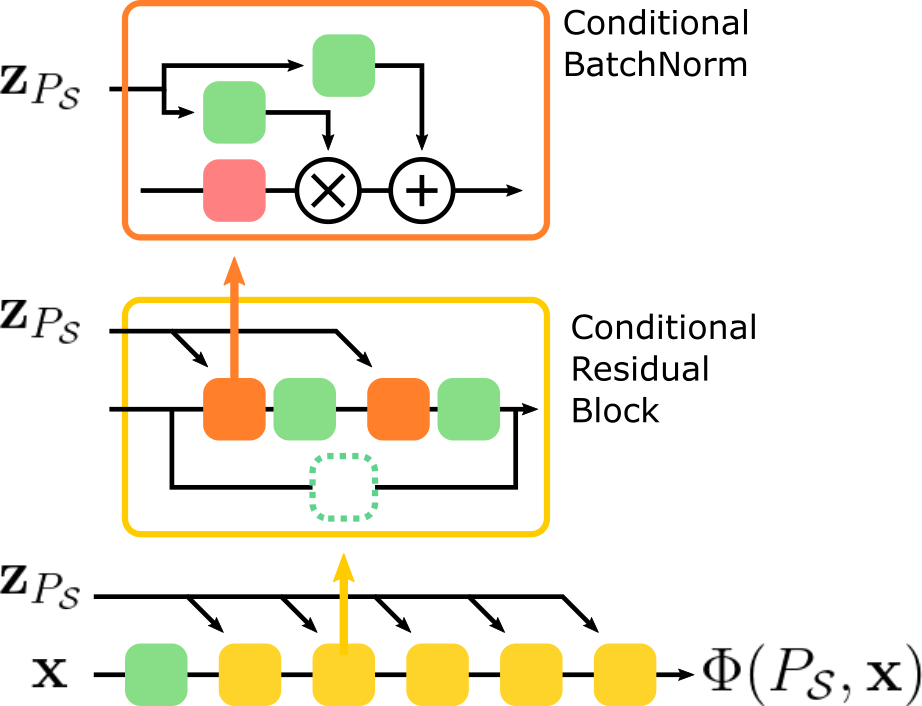}\\
    (b) Decoder
    \vspace*{3mm} \\
    ~\\
    \includegraphics[width=0.8\linewidth]{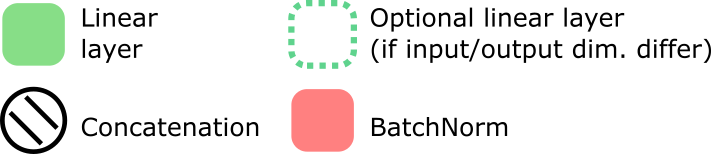}\\
    \end{tabular}
    \vspace*{3mm}
    \caption{Network used in {\method}, similar to~\cite{mescheder2019occupancy}}
    \label{fig:network}
\end{figure}

\section{Network}
\label{supp:sec:network}

In this work, we build on the network from Occupancy Networks~\cite{mescheder2019occupancy}, which has an encoder-decoder architecture as presented in Figure~\ref{fig:network}.

\paragraph{Encoder.}
The encoder, illustrated in Fig.\,\ref{fig:network}(a), is a Residual PointNet~\cite{qi2017pointnet} composed of 5 residual PointNet blocks, each containing two linear layers and a skip connection.

The encoder is applied point-wise.
To gather global information at the output of each residual block, 
we use a global max-pooling layer (over all points of the point cloud) and concatenate the local feature vector (at point level) and the global feature vector (at point-cloud level) before entering the next block.

As mentioned in Section 3.3.1 of the main paper, our model is 
trained as a variational auto-encoder. Hence, our encoder outputs two vectors $\mu_{\pointcloud_\shape}$ and $\sigma_{\pointcloud_\shape}$, that encode respectively the mean and standard deviation of the normal distribution used for generating the latent code $\vect{z}_{\pointcloud_\shape}$ at training time.
At test time, we use $\vect{z}_{\pointcloud_\shape} = \mu_{\pointcloud_\shape}$ to be deterministic with respect to the input point cloud.

\paragraph{Decoder.}
The decoder, illustrated in Figure~\ref{fig:network}(b), is a conditional multi-layer perceptron (MLP)~\cite{de2017modulating,dumoulin2016adversarially}.
It is composed of five conditional residual block.
Each block is conditioned at batch normalization by the latent code $\vect{z}_{\pointcloud_\shape}$.
The query point, $\vect{x} \,{\in}\, \rthree$, is the location at which the occupancy is estimated.
Note that the occupancy is estimated with $S(\decoder(\pointcloud_\shape, \vect{x}))$, where $S$ is the sigmoid function, not represented in Figure~\ref{fig:network}(b).

\paragraph{Parameters.}
In all our experiments, the latent vector size representing a shape is set to 256.
The hidden size of the encoder is also 256, and the latent size of the decoder is set to 512.
The batch-norm layer is the usual 1-dimensional batch normalization.

\paragraph{Training procedure.}
The encoder and decoder are trained end-to-end.
In all experiments, we use Adam~\cite{kingma2017adam} as optimizer with an initial learning rate set to $10^{-3}$.
To train with mini-batches, we generate the same number of opposite-side needles (300, the same as the input point cloud size) and the same number of same-side needles (2048, to cover the space around the shape) for each model.
The batch size is set to 32.

In practice, we observed a low dependency of the performance to the number of needles (e.g., doubling the size of $Q_\opposite$ and $Q_\same$), so we chose these parameters values to permit us to train the model with a middle range GPU NVidia 2080Ti.

\paragraph{Computational cost.}
Our only additional cost w.r.t. occupancy-supervised methods is the computation, at training time, of nearest neighbors, which is little in our setting. Distance-based methods have a similar ``extra'' cost as ours.

\section{Loss derivation}
\label{supp:sec:lossderiv}

In the main paper, the final loss and partial derivative expressions, i.e., Eq.\,(13) and Eq.\,(14), are given without a complete derivation. We derive here these expressions.

\subsection{Training loss}

The binary cross entropy reads as
\begin{alignat}{3}
\label{eq:loss_derive1}
& \BCE( \bprob_{\vect{x}, \vect{y}}, \bprob^0_{\vect{x}, \vect{y}})
&&=
&&-\bprob^0_{\vect{x}, \vect{y}} \log(\bprob_{\vect{x}, \vect{y}}) \nonumber\\
& &&
&& - (1-\bprob^0_{\vect{x}, \vect{y}}) \log(1-\bprob_{\vect{x}, \vect{y}}),
\end{alignat}
where, recalling Eq.\,(4) of the main paper,
\begin{equation}
    \bprob_{\mathbf{x}, \mathbf{y}} = \bprob_\mathbf{x} \, \bprob_\mathbf{y} + (1-\bprob_\mathbf{x})(1-\bprob_\mathbf{y}).
\end{equation}
Subsbituting the expression of $\bprob_{\mathbf{x}, \mathbf{y}}$ in Eq.~\eqref{eq:loss_derive1} yields
\begin{alignat}{3}
\label{eq:loss_derive2}
& \BCE && ( \bprob_{\vect{x}, \vect{y}}, \bprob^0_{\vect{x}, \vect{y}})= \nonumber \\
& &&-\bprob^0_{\vect{x}, \vect{y}} \log(\bprob_{\vect{x}}\bprob_{\vect{y}} + (1-\bprob_{\vect{x}})(1-\bprob_{\vect{y}})) \nonumber\\
& && - (1-\bprob^0_{\vect{x}, \vect{y}}) \log(1-\bprob_{\vect{x}}\bprob_{\vect{y}} - (1-\bprob_{\vect{x}})(1-\bprob_{\vect{y}})).
\end{alignat}
Then, using the fact that 
$\bprob_{\vect{x}} = S \circ \network(\pointcloud_\shape, \mathbf{x})$ and
$\bprob_{\vect{y}} = S \circ \network(\pointcloud_\shape, \mathbf{y})$,
where $S$ is the sigmoid function, we obtain:
\begin{alignat}{3}
\label{eq:loss_derive3}
& \BCE &&( \bprob_{\vect{x}, \vect{y}}, \bprob^0_{\vect{x}, \vect{y}})= \nonumber\\
& &&-\bprob^0_{\vect{x}, \vect{y}} \log(\frac{
    e^{\network(\pointcloud_\shape, \mathbf{x})}
    e^{\network(\pointcloud_\shape, \mathbf{y})}
    +1
    }{
    (e^{\network(\pointcloud_\shape, \mathbf{x})}+1)
    (e^{\network(\pointcloud_\shape, \mathbf{y})}+1)
    })\nonumber\\
& && - (1-\bprob^0_{\vect{x}, \vect{y}}) \log(\frac{
    e^{\network(\pointcloud_\shape, \mathbf{x})}+
    e^{\network(\pointcloud_\shape, \mathbf{y})}
    }{
    (e^{\network(\pointcloud_\shape, \mathbf{x})}+1)
    (e^{\network(\pointcloud_\shape, \mathbf{y})}+1)
    }).
\end{alignat}
Finally, we can simplify the expression by noticing that the denominators in the log terms are identical. It yields:
\begin{alignat}{3}
\label{eq:loss_derive4}
& \BCE( \bprob_{\vect{x}, \vect{y}}, \bprob^0_{\vect{x}, \vect{y}})
&&=
&&  \log(e^{\network(\pointcloud_\shape, \mathbf{x})}+1) +
    \log(e^{\network(\pointcloud_\shape, \mathbf{y})}+1)  \nonumber\\
& &&
&&-\bprob^0_{\vect{x}, \vect{y}} \log(
    e^{\network(\pointcloud_\shape, \mathbf{x})}
    e^{\network(\pointcloud_\shape, \mathbf{y})}
    +1
    )\nonumber\\
& &&
&& - (1-\bprob^0_{\vect{x}, \vect{y}}) \log(
    e^{\network(\pointcloud_\shape, \mathbf{x})}+
    e^{\network(\pointcloud_\shape, \mathbf{y})}
    )\nonumber \\
& &&=
&&  \log(e^{\network(\pointcloud_\shape, \mathbf{x})}+1) +
    \log(e^{\network(\pointcloud_\shape, \mathbf{y})}+1)  \nonumber\\
& &&
&&-\bprob^0_{\vect{x}, \vect{y}} \log(
    e^{\network(\pointcloud_\shape, \mathbf{x}) + \network(\pointcloud_\shape, \mathbf{y})}
    +1
    )\nonumber\\
& &&
&& - (1-\bprob^0_{\vect{x}, \vect{y}}) \log(
    e^{\network(\pointcloud_\shape, \mathbf{x})}+
    e^{\network(\pointcloud_\shape, \mathbf{y})}
    ).
\end{alignat}

\subsection{Gradient}
The gradient that backpropagates in $\network$ is 
$\partial \BCE/\partial \network(\vect{x}) + \partial \BCE/\partial \network(\vect{y})$.
As $\vect{x}$ and $\vect{y}$ are interchangeable in the loss expression, we derive only the $\vect{x}$-term. We have:
\begin{alignat}{3}
& \frac{\partial \BCE(\bprob_{\mathbf{x}, \mathbf{y}}, \bprob^0_{\mathbf{x}, \mathbf{y}})}{\partial \network(\vect{x})}
&&= 
&& \; S(\network(\vect{x})) 
- \bprob_{\mathbf{x}, \mathbf{y}}^0 \; S(\network(\vect{x})+\network(\vect{y})) \nonumber \\
& && 
&&- (1 - \bprob_{\mathbf{x}, \mathbf{y}}^0) \; S(\network(\vect{x})-\network(\vect{y})),
\end{alignat}
where we used the facts that
\begin{alignat}{3}
& \frac{\partial \log(e^a+1)}{\partial a}
= 
\frac{e^a}{e^a+1} 
= 
S(a),
\end{alignat}
\begin{alignat}{3}
& \frac{\partial \log(e^{a+b}+1)}{\partial a} 
= 
\frac{e^{a+b}}{e^{a+b}+1} 
= 
S(a+b),
\end{alignat}
and
\begin{alignat}{3}
& \frac{\partial \log(e^a+e^b+1)}{\partial a}
= 
\frac{e^a}{e^a+e^b} 
= 
\frac{e^{a-b}}{e^{a-b}+1} 
= 
S(a-b).
\end{alignat}
%
The gradient expression can be further simplified by exploiting the fact that $\bprob_{\vect{x},\vect{y}}^0 \,{=}\, 1$ when $ (\vect{x},\vect{y}) \,{\in}\, Q_\same$, and $\bprob_{\vect{x},\vect{y}}^0 \,{=}\, 0$ when $(\vect{x},\vect{y}) \,{\in}\, Q_\opposite$.

\section{Latent space analysis}
\label{supp:sec:latent}

In this section, we propose an analysis of the latent space through an evaluation of the generation capacity of the network, the possibility of interpolating in the latent space between two shapes, as well as a visualization of the latent space itself over a multi-class dataset.

\subsection{Shape generation}

\begin{figure}[t]
    \centering
    \setlength{\tabcolsep}{1.5pt}
    \begin{tabular}{cc}
        \includegraphics[width=0.40\linewidth]{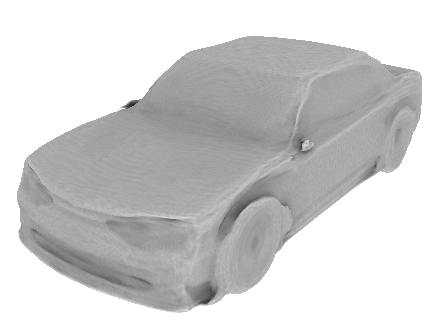}&
        \includegraphics[width=0.40\linewidth]{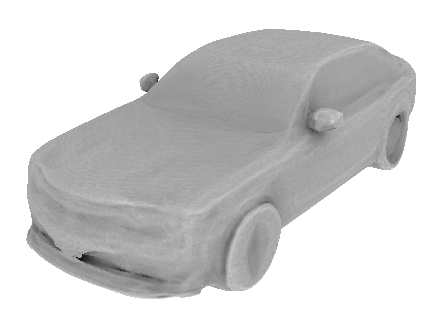}\\
        \includegraphics[width=0.40\linewidth]{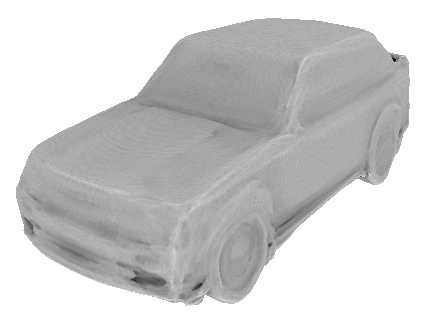}&
        \includegraphics[width=0.40\linewidth]{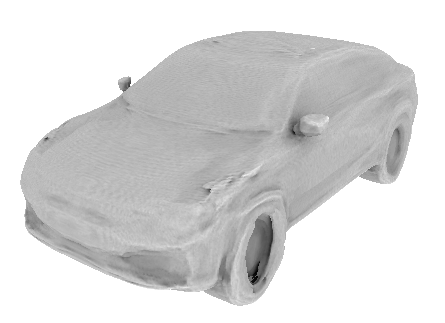}\\
        \\
        \multicolumn{2}{c}{\small (a) Random generation with a model trained only on cars}\\
        ~\\
        \includegraphics[width=0.40\linewidth]{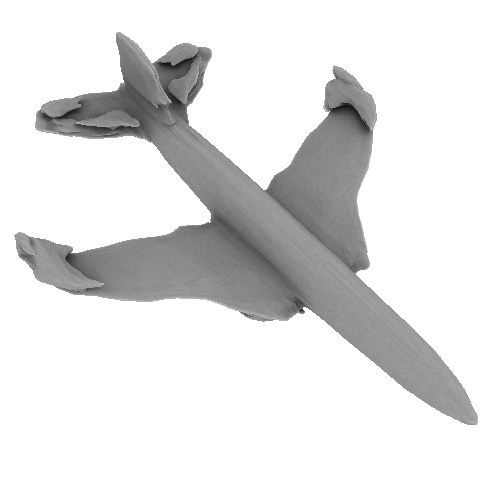}&
        \includegraphics[width=0.40\linewidth]{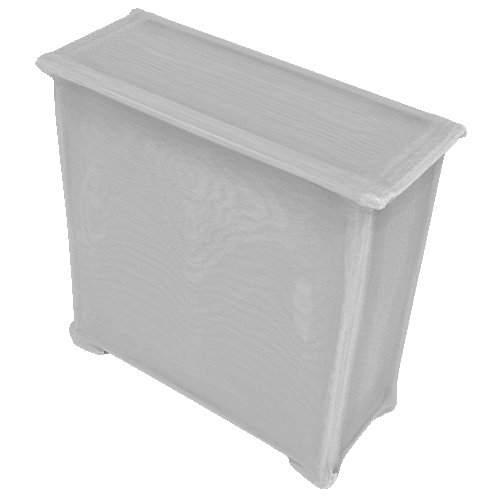}\\
        \includegraphics[width=0.40\linewidth]{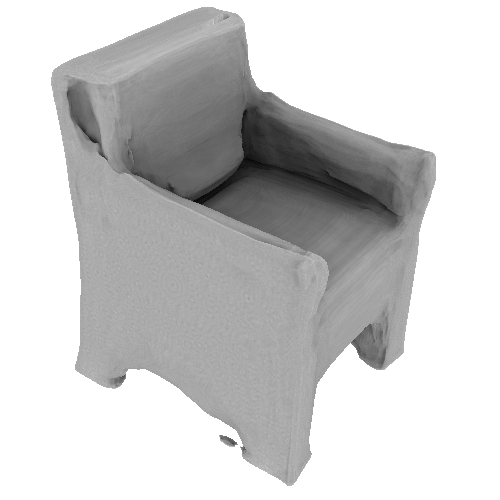}&
        \includegraphics[width=0.40\linewidth]{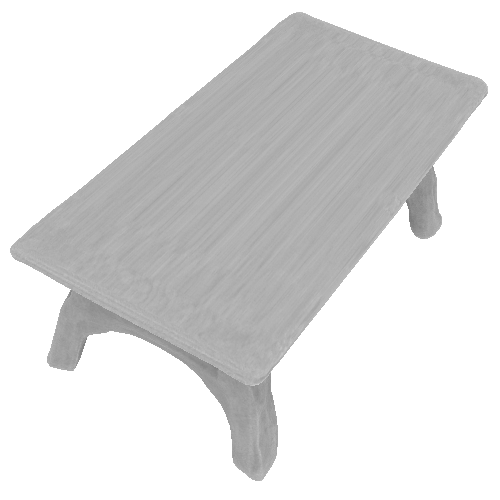}\\
        \multicolumn{2}{c}{\small (b) Random generation with a model trained on all categories}\\
        ~\\
    \end{tabular}
    \caption{Shape generation with {\method} using latent vectors randomly sampled according to the target class distribution.}
    \label{fig:generation}
\end{figure}

We use the same network as in Occupancy Networks~\cite{mescheder2019occupancy}, which uses a variational auto-encoder formulation.
As in previous works, we can use our trained network for shape generation.
In our network, the encoder outputs a latent vector $\latent$ representing the shape; the decoder outputs the probability of occupancy given the latent code and the spatial coordinates of a point at which to evaluate the occupancy.
We add a latent regularization term similar to~\cite{mescheder2019occupancy} to the training loss so that the latent space allows generation and interpolation between shapes.
For a complete overview of variational auto-encoders, one could refer to~\cite{doersch2016tutorial}.
Instead of generating a latent vector from $\pointcloud_\shape$, we pick a random $\latent$
and reconstruct the shape based on this latent vector.
A few generation results are shown in Figure~\ref{fig:generation}.

\paragraph{Generation for a model trained on a single category.}
We show in Figure~\ref{fig:generation}(a) some shapes generated with a model trained only on cars.
The latent code $\latent$ is generated according to a normal law where the mean and standard deviation are computed over the train set.
We observe that our {\method} model is able to generate plausible shapes, including details such as wing mirrors or bumpers.

\paragraph{Conditional generation.}
In Figure~\ref{fig:generation}(b), we show a few conditional generation results for several classes of the ShapeNet dataset.
For conditional generation, we follow the same procedure as in the previous section, except for the normal distribution parameters that are evaluated on the train set restricted to the desired category.

\begin{figure*}[t]
    \centering
    \setlength{\tabcolsep}{1.5pt}
    \begin{tabular}{c|ccccc|c}
        \includegraphics[width=0.14\linewidth]{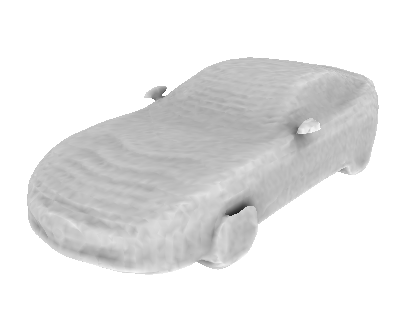}&
        \includegraphics[width=0.14\linewidth]{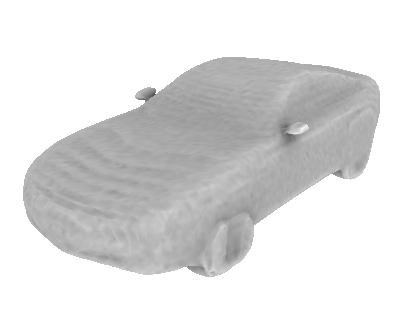}&
        \includegraphics[width=0.14\linewidth]{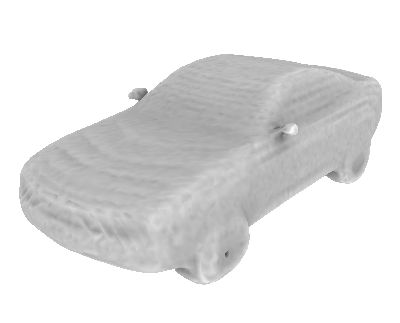}&
        \includegraphics[width=0.14\linewidth]{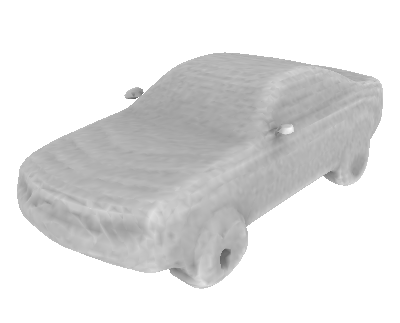}&
        \includegraphics[width=0.14\linewidth]{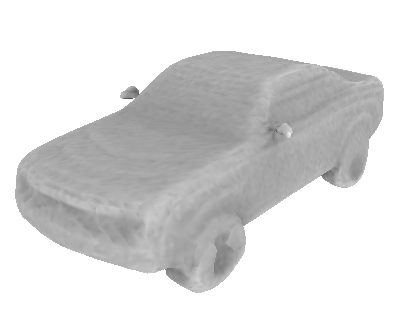}&
        \includegraphics[width=0.14\linewidth]{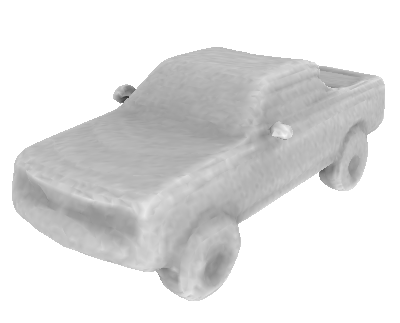}&
        \includegraphics[width=0.14\linewidth]{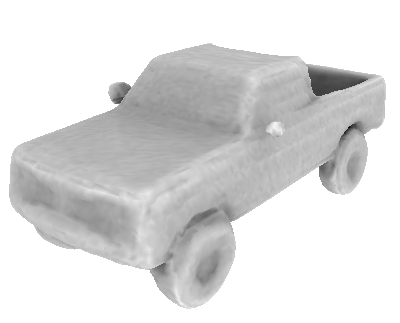}
        \\
        \includegraphics[width=0.14\linewidth]{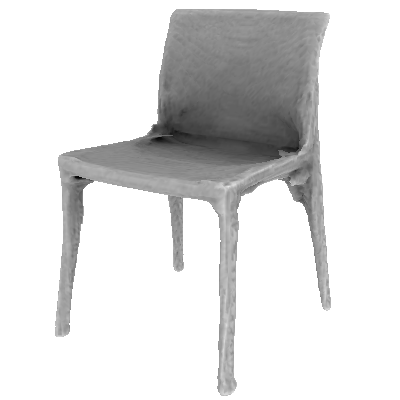}&
        \includegraphics[width=0.14\linewidth]{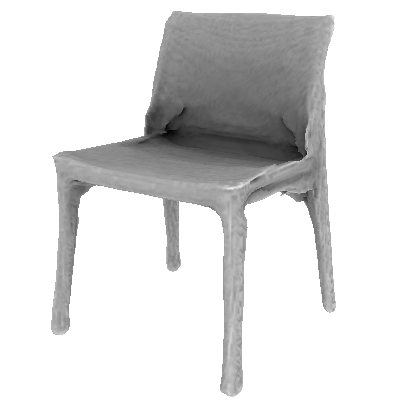}&
        \includegraphics[width=0.14\linewidth]{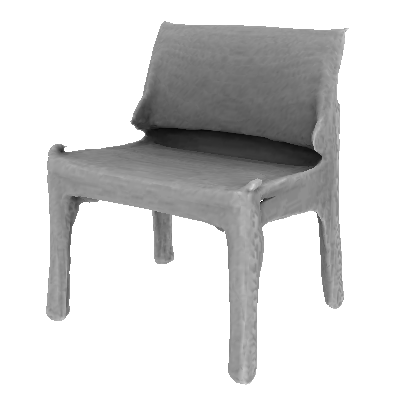}&
        \includegraphics[width=0.14\linewidth]{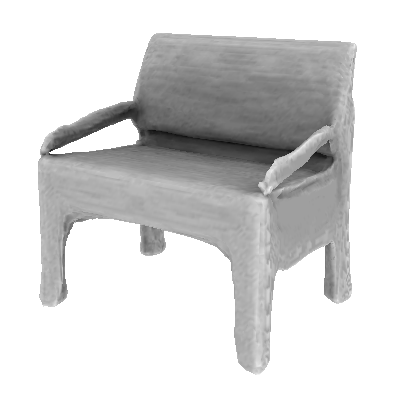}&
        \includegraphics[width=0.14\linewidth]{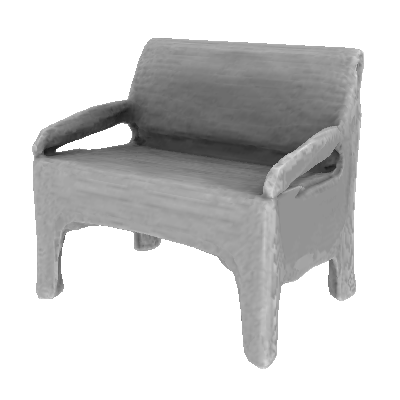}&
        \includegraphics[width=0.14\linewidth]{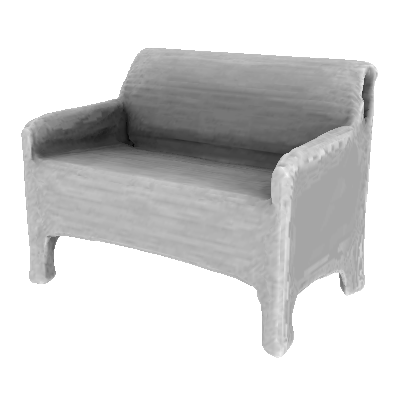}&
        \includegraphics[width=0.14\linewidth]{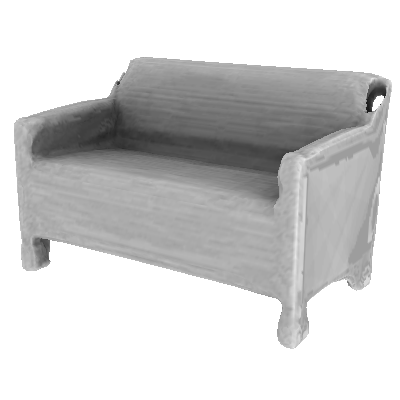}
        \\
        \includegraphics[width=0.14\linewidth]{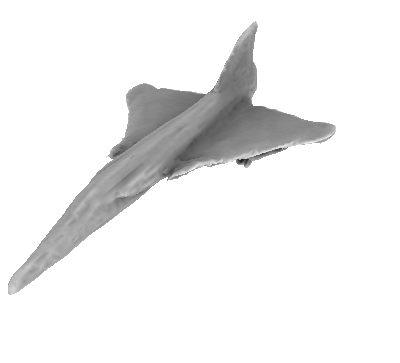}&
        \includegraphics[width=0.14\linewidth]{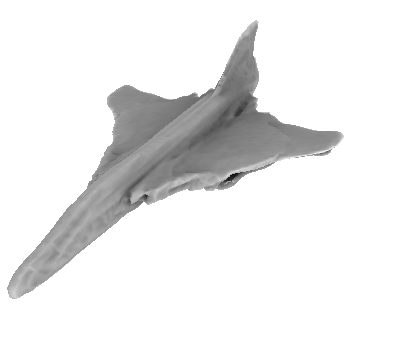}&
        \includegraphics[width=0.14\linewidth]{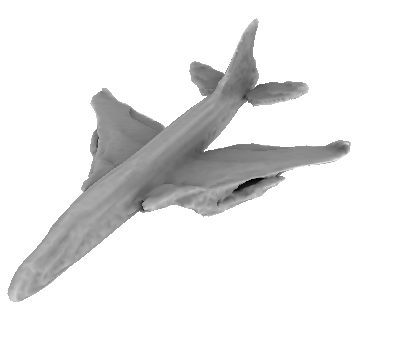}&
        \includegraphics[width=0.14\linewidth]{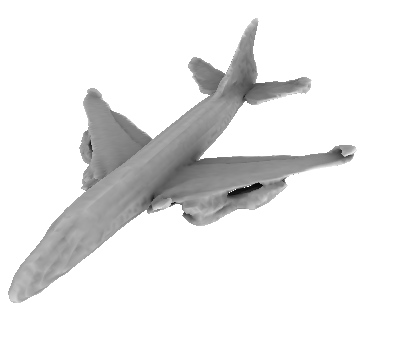}&
        \includegraphics[width=0.14\linewidth]{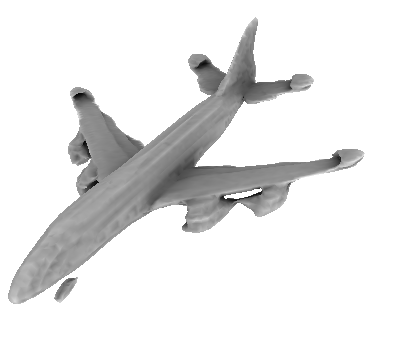}&
        \includegraphics[width=0.14\linewidth]{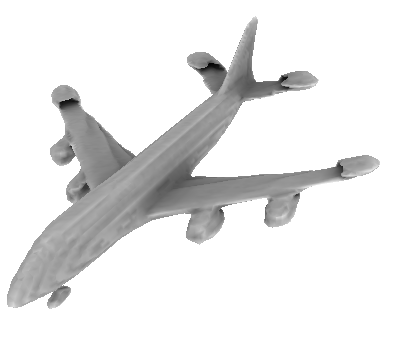}&
        \includegraphics[width=0.14\linewidth]{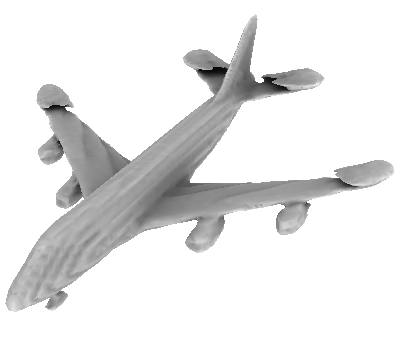}
        \\
        Shape 1 & &&&&& Shape 2\\
    \end{tabular}
    \vspace*{0mm}
    \caption{Linear interpolation in \method's latent space between two shapes (shape 1 and shape 2) of the test set.}
    \label{fig:interpolation}
\end{figure*}

\subsection{Shape interpolation}

To illustrate the consistency of \method's latent space, we show here that it can be used to interpolate smoothly between two shapes via a simple linear interpolation between the corresponding latent codes: no spurious artifacts or intermediate shapes unrelated to the end-point shapes appear on the linear path from one latent code to another.

In Figure~\ref{fig:interpolation}, we present interpolations between three couples of shapes.
For each row of the figure, we linearly interpolate the latent vectors with a fixed step and show the reconstructions.
Please note that each row corresponds to different object categories but a single network is used for all of them, trained on all categories of ShapeNet.

We observe good interpolations between the shapes, with 
a smooth transformation from one shape to the other and without unrelated shapes appearing along the path (e.g., a chair interpolated between cars).
Each interpolated shape seems a plausible realization of the category it belongs to.

\begin{figure}[t]
    \centering
    \setlength{\tabcolsep}{1.5pt}\vspace{-3mm}
    \begin{tabular}{cc}
    \multicolumn{2}{c}{\includegraphics[width=0.49\linewidth]{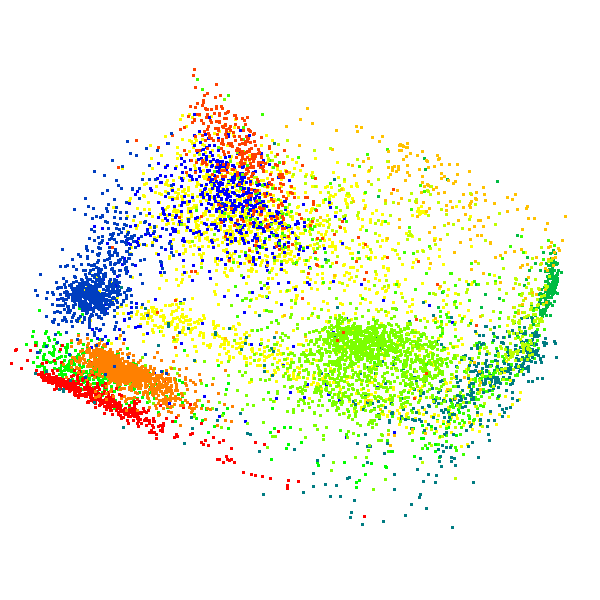}}\\[-3mm]
    \multicolumn{2}{c}{\small (a) 2D PCA}\\[2mm]
    \includegraphics[width=0.49\linewidth]{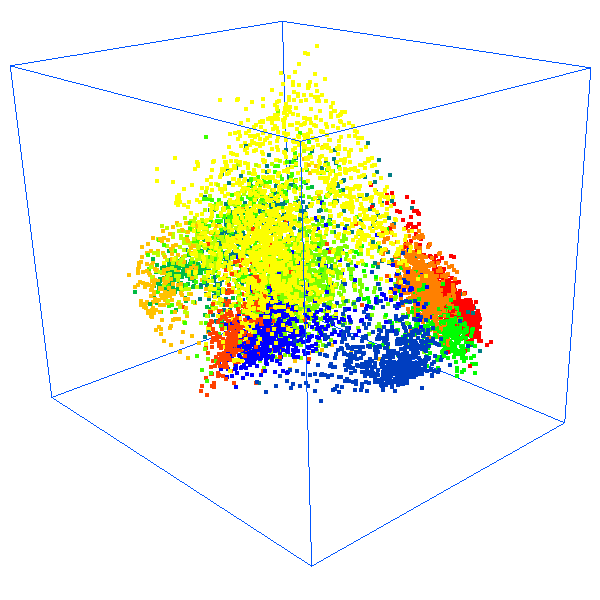}&
    \includegraphics[width=0.49\linewidth]{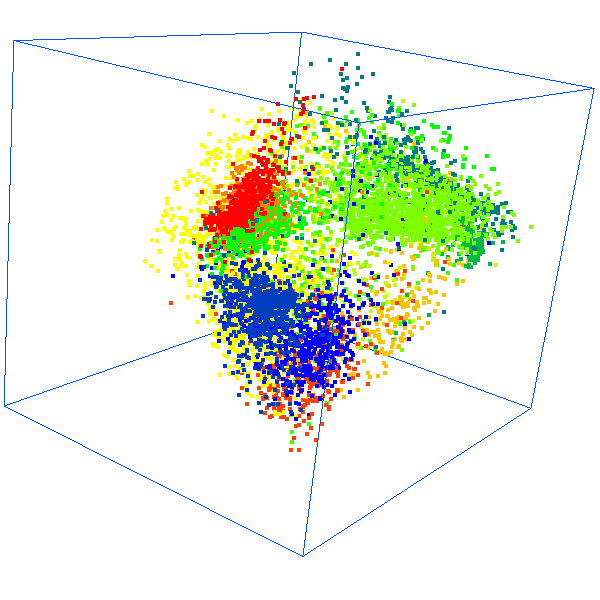}\\[-3mm]
    \multicolumn{2}{c}{\small (b) 3D PCA}
    \end{tabular}
    \vspace*{1mm}
    \caption{Principal component analysis of the latent space of the network trained on ShapeNet parts, with one color per category.}
    \label{fig:pca}
\end{figure}

\subsection{Latent space visualization}

In order to further study the latent space of the multi-category ShapeNet models, we perform a Principal Component Analysis (PCA) of sampled latent vectors with 2 or 3 principal dimensions.
Corresponding visualizations are displayed on Figure~\ref{fig:pca}.
We represent each category with a different color to emphasize the natural clustering of the latent vectors for each shape category. 

We can see clusters of shapes from the same category, sharing common geometric features (e.g., wings for planes).
As the clusters are mostly 
separable (one model category per cluster) and somehow convex, any linear interpolation between two shapes of the same class will likely remain in the cluster.
Figure~\ref{fig:pca} visually accounts for the success of the previously presented interpolations. 

\begin{table}[t]
    \centering
    \vspace*{7mm}
    \begin{tabular}{c|c}
    \toprule
    Method & Overall accuracy \\
    \midrule
        SVM & 94.5\% \\
        MLP & 93.2\% \\
        Random Forest & 93.3\%\\
    \bottomrule
    \end{tabular}
    \vspace*{2mm}
    \caption{Accuracy of classifiers learned on the latent representations of ShapeNet parts.}
    \label{tab:classif}
    \vspace*{-3mm}
\end{table}

\subsection{Classification based on latent vectors}

In order to further investigate the separability of the classes in the latent space, we train three simple classifiers on the latent vectors from the train set and evaluate on the test set.
Please note that for ShapeNet, we use the train/test split used usually for part segmentation. It contains 13 very different model categories, from planes to chairs.
We experimented with three classifiers: a Support Vector Machine (SVM)~\cite{cortes1995support}, 1-hidden-layer multilayer perceptron (MLP) and a Random Forest~\cite{ho1995random}). 

The results are presented in Table~\ref{tab:classif}.
All classifiers perform more or less equally well, with more than 93\% overall accuracy.
Besides, the parameters of the classifiers were not tuned for this experiment; we used the default parameters provided by the Scikit-Learn library~\cite{scikit-learn}.
These results validate the fact that the different shape categories are mostly separable in the latent space, as can be seen in Figure~\ref{fig:pca}.

\section{Needle dropping analysis}
\label{supp:sec:needles}

As described in the main paper, the loss function is composed of two terms: $\loss_{\opposite}$ and $\loss_{\same}$.
The former, computed on the set $Q_{\opposite}$ of ``opposite-side'' needles, aims at setting different labels on the different sides of the surface.
The latter, computed on the set $Q_{\same}$ of ``same-side'' needles, enforces consistent labels inside and outside the shape.

The construction of valid $Q_{\opposite}$ and $Q_{\same}$ is critical in our approach.
Ideally, these two sets of needles should contain only ``good'' needles, i.e., needles for which their end-points are indeed on opposite sides of the surface for $Q_{\opposite}$, and on the same side of the surface for $Q_{\same}$.
In practice, as the method is fully self-supervised, such objective is difficult to reach and each set contains wrong needles.

\subsection{Needle error localization}

\begin{figure}
    \begin{center}
    \vspace*{-3mm}
    \begin{tabular}{cc}
        \includegraphics[width=0.40\linewidth]{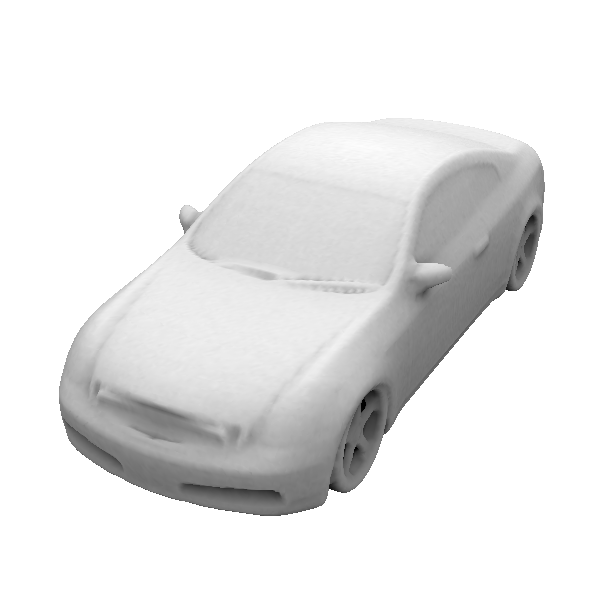}&
        \includegraphics[width=0.40\linewidth]{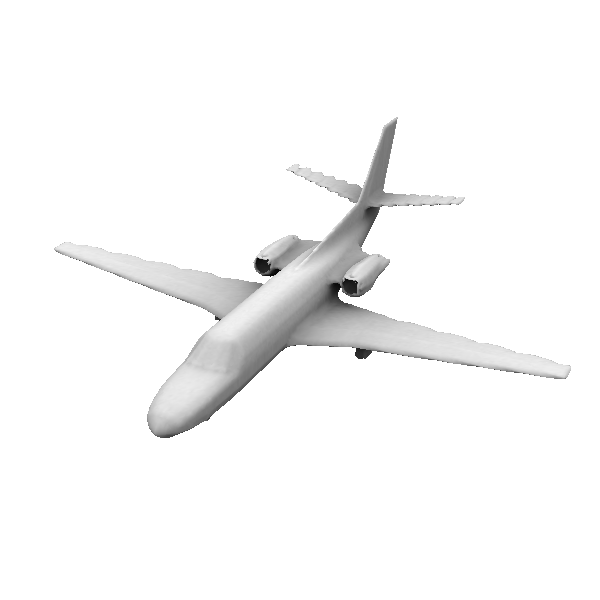}
        \\[-3mm]
        \multicolumn{2}{c}{\small (a) Ground truth mesh.}
        \\[5mm]
    
        \includegraphics[width=0.40\linewidth]{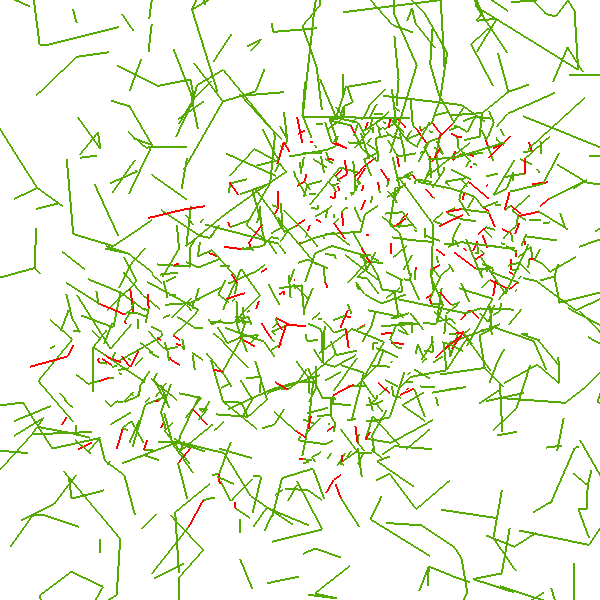}&
        \includegraphics[width=0.40\linewidth]{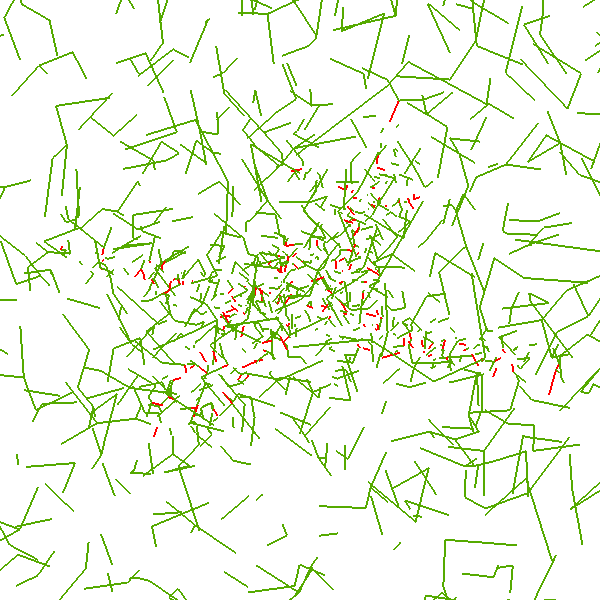}
        \\[1mm]
        \multicolumn{2}{c}{\small (b) Same-side needles, $Q_{\same}$.}
        \\[0mm]

        \includegraphics[width=0.40\linewidth]{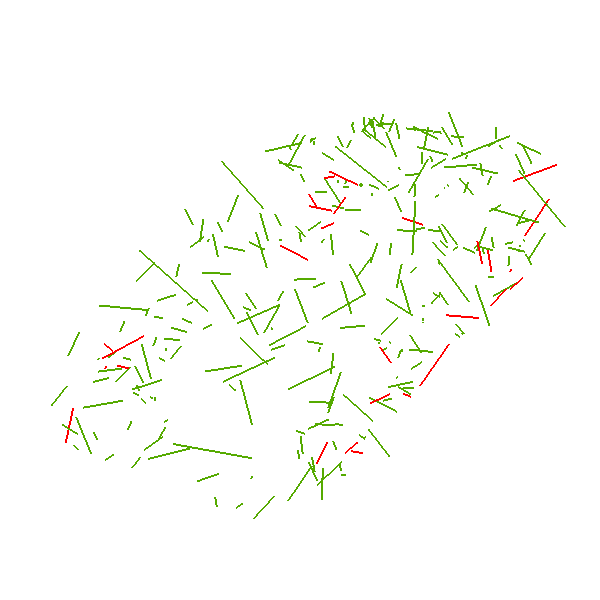}&
        \includegraphics[width=0.40\linewidth]{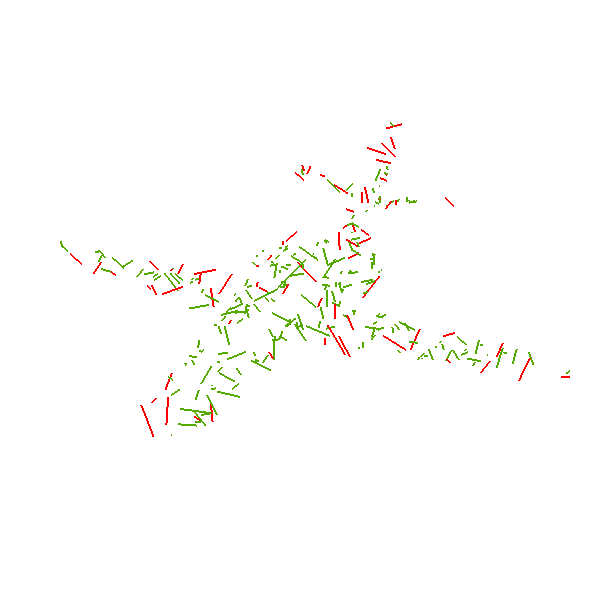}
        \\[-3mm]
        \multicolumn{2}{c}{\small (c) Opposite-side needles, $Q_{\opposite}$.}
        \\
    \end{tabular}
    \end{center}
    
    \caption{Needle dropping visualization for $Q_\same$ (b) and $Q_\opposite$ (c) for two given shapes (a).
    Green needles belong to correct set; red needles are in the incorrect set and should have been placed in the other set according to the ground-truth mesh (if it could have been known at training time).}
    \label{fig:needles}
    
    \begin{center}
    \begin{tabular}{cc}
        \includegraphics[width=0.40\linewidth]{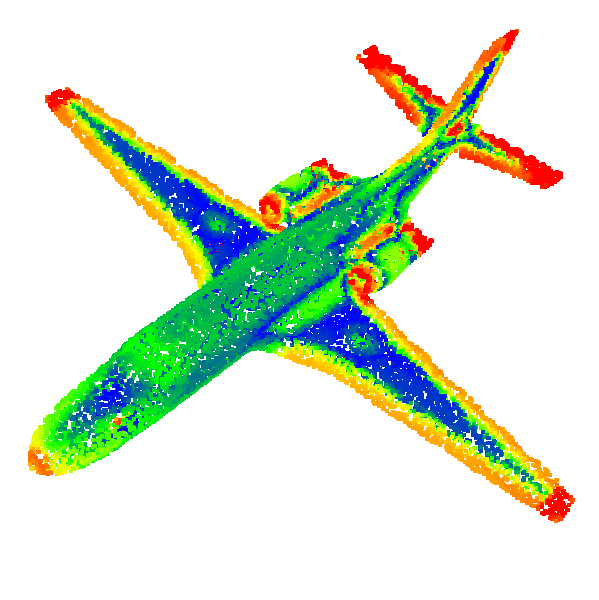}&
        \includegraphics[width=0.40\linewidth]{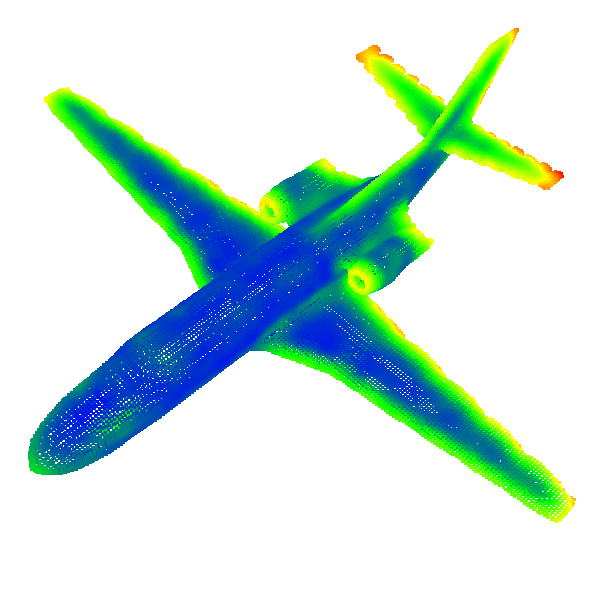}\\
        \small (a) Curvature  & \small (b) $Q_{\opposite}$ error
    \end{tabular}
    \end{center}
    
    \caption{Visualization of the surface curvature (a) and of the frequency of incorrectly classified needles, estimated over several random draw of $Q_\same$ (b). 
    }
    \label{fig:needles_dense}
\end{figure}

Figure~\ref{fig:needles} presents a random draw of $Q_{\opposite}$ and $Q_{\same}$ on two shapes, for the default value of $\sigma_{\vect{h}}$.
Correct needles are green and wrong needles are red.

First, we observe that a majority of the needles are green, which validates the default value used for $\sigma_{\vect{h}}(\vect{p}) = d_{\vect{p}} / 3$, where $d_{\vect{p}}$ is the distance between a point $\vect{p}$ to its closest point in $\pointcloud$.

Second, the number of errors on $Q_{\same}$ appears to be approximately the same on both 3D models.

Third, we observe more errors in $Q_{\opposite}$ on the plane model than on the car model.
One possible explanation is that $Q_{\opposite}$ is sensitive to the curvature of the surface to be estimated.
To validate this assumption, we compute the curvature of the ground-truth surface,  as illustrated on Figure~\ref{fig:needles_dense}(a), and compare it to the concentration of error for $Q_{\opposite}$, as shown in Figure~\ref{fig:needles_dense}(b).
These concentrations are obtained by aggregating a large number of random pickings of $Q_{\opposite}$.
Each wrong needle votes for the closest vertex of the surface mesh and a Gaussian filter is applied for smooth visualization.
We may observe that the errors for $Q_{\opposite}$ indeed concentrate in high-curvature regions.

\subsection{Needle error statistics}

The construction of $Q_{\opposite}$ is based on the hypothesis that the surface is locally planar.
At the scale of observation ($\sigma_{\vect{h}}$), this assumption holds in mostly planar parts of $\shape$ ($0$~curvature), but it is erroneous in high-curvature areas.
A seemingly good solution would be to reduce $\sigma_{\vect{h}}$ such that we observe the surface at a scale where all neighborhoods of points are planar.
In Table~\ref{tab:sigma}, we use different values of $\sigma_{\vect{h}}$ and compute the rate of correct needles for $Q_\opposite$ and $Q_\same$.

\begin{table}
    \centering
    \begin{tabular}{c|c|cc}
    \toprule
        Category & Multiplier $\alpha$ & Opposite- & Same- \\
        & ($\sigma_{\vect{h}}(\vect{p})=\alpha \frac{d_\vect{p}}{3}$) & side & side\\
    \midrule
        Planes  & 2     & 70.2\% & 92.7\% \\
                & 1     & 83.9\% & 92.4\% \\
                & 0.5   & 91.8\% & 92.0\% \\
                & 0.1   & 98.2\% & 91.6\% \\
                & 0.01  & 99.8\% & 91.5\% \\
    \midrule
        Cars    & 2     & 76.0\% & 90.6\% \\
                & 1     & 82.9\% & 90.7\% \\
                & 0.5   & 88.8\% & 90.3\% \\
                & 0.1   & 97.1\% & 89.7\% \\
                & 0.01  & 99.7\% & 89.5\% \\
    \bottomrule
    \end{tabular}
    \vspace*{2mm}
    \caption{Rate of good needles for $Q_\opposite$ and $Q_\same$ as a function of $\sigma_\vect{h}$, when constructed on planes and on ShapeNet cars.}
    \label{tab:sigma}
\end{table}

As expected, the smaller the scale $\sigma_{\vect{h}}$, the better $Q_{\opposite}$.
But on the other hand, reducing the value of $\sigma_{\vect{h}}$ simultaneously decreases the rate of good needles in $Q_{\same}$. This confirms that one should find a trade-off when setting $\sigma_{\vect{h}}$.

\begin{figure}
    \centering
    \setlength{\tabcolsep}{3pt}
    \begin{tabular}{c}
        \includegraphics[width=0.9\linewidth]{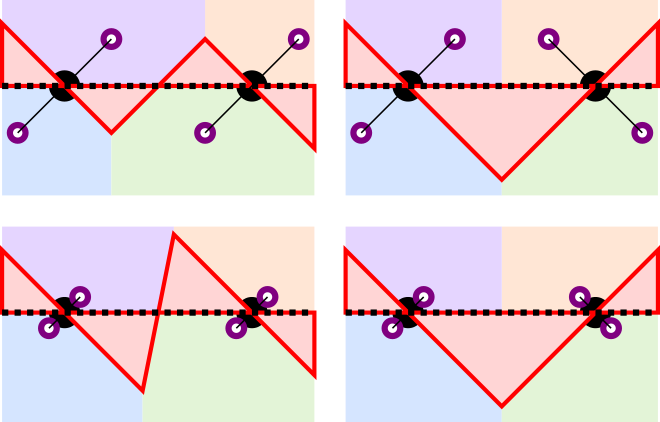}\\
        (a) No point belonging to $\pointcloud_{\same}$.\\
        ~\\
        \includegraphics[width=0.9\linewidth]{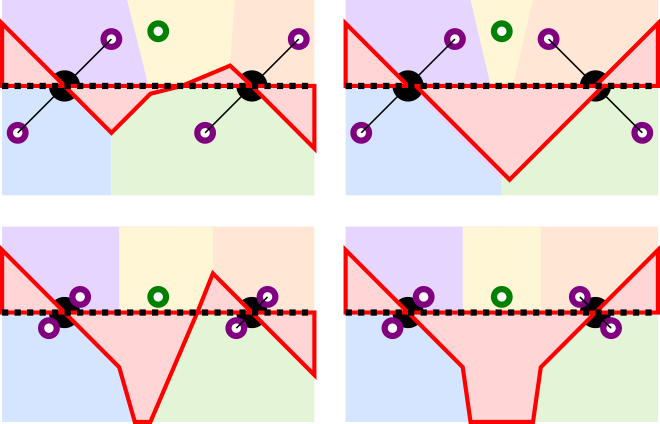}\\
        (b) A point belonging to $\pointcloud_{\same}$ is also present.\\
        \includegraphics[width=0.9\linewidth]{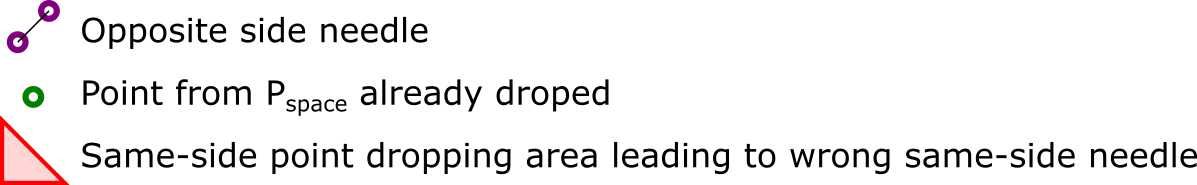}\\
    \end{tabular}
    \caption{Influence of $\sigma_{\vect{h}}$ on the probability of picking a wrong needle for $Q_\same$: (a) with only points from $\pointcloud_\opposite$, and (b) with points from $\pointcloud_\opposite$ and a point from $\pointcloud_\same$. ($\pointcloud_{\text{space}} \,{=}\, \pointcloud_\same \,{\cup}\, \pointcloud_\opposite$ is disjoint from~$\pointcloud_\shape$.)}
    \label{fig:misclassified_same}
\end{figure}

An illustration of this phenomenon on a 2D case is proposed in Figure~\ref{fig:misclassified_same}.
As explained in Section 3.2.2 of the paper, when constructing $Q_{\same}$ we first pick points in space and then build needles with the points of $\pointcloud_\same \cup \pointcloud_\opposite$.
We consider 8 cases, with two $\sigma_{\vect{h}}$ values (large value on the first row and small value on the second row), two configurations of needles (different orientations, first and second column) and two cases, with and without a point from $\pointcloud_\same$ to build a needle (Figure~\ref{fig:misclassified_same}(a) and (b)).
In the latter case, in the illustration, the point from $\pointcloud_\same$ is placed at distance $~\sigma_{\vect{h}}$ from the shape.

The colored areas (except those in red) represent the Vorono\"i cells around each point, inside which a sampled point of $\pointcloud_\same$ yields a good needle. On the contrary, if a new point of $\pointcloud_\same$ falls in the red area, then the constructed needle falls on the opposite side of the surface and produces a bad needle for $Q_{\same}$.
In all these configurations, reducing $\sigma_{\vect{h}}$ (i.e., going from first row to second row) leads to larger red areas, 
thus increasing the probability of wrong needles in $Q_{\same}$.
It corroborates the observation made on ShapeNet in Table~\ref{tab:sigma}.

\section{Point sampling at training time}
\label{supp:sec:sampl}

In the main paper, for comparison purposes, we followed the same training procedure as in the previous papers, i.e., sampling new points on the surface of each object \emph{at each epoch}.
Therefore, over a long training period of $N_\text{e}$ epochs, the overall information originating from each shape cannot be strictly considered to be limited to a set of $300$ points. Yet, it is not equivalent either to training with $300 N_\text{e}$ points because the information comes in small fractions of $300$ points per shape, that have to be processed at a time. It impacts for instance the distance to the nearest neighbor, which is very different when computed on $300$ points or on $300 N_\text{e}$ points.

Here, we perform another set of experiments with fixed points for each training shape (same points at each epoch) to investigate the consequences of a truly limited amount of information on each shape at training time.
In this setting, each shape is sampled once and for all with 300 points. 

Results on the three datasets used in the main paper are presented in Table~\ref{tab:exp_complete2}.
Conclusions are threefold.
First, we notice that the training is stable, no divergence is observed.
Second, fixed-points training reaches almost the same performances as training with point re-sampling.
Third, the size of the training dataset has an impact on the robustness to sampling procedure: on the car subset, the average Chamfer distance increase by $7 \times 10^{-5}$, while on all ShapeNet performances are nearly identical with both point sampling strategies.

\begin{table}
\centering
\setlength{\tabcolsep}{3pt}
\begin{tabular}{l|ccc}
\toprule
                & \multicolumn{2}{c}{Chamfer $\ell_2$ $\downarrow$} \\
Point sampling & Mean & Median\\
\midrule
Resampled  & 1.703 & 1.109\\
Fixed       & 2.461 & 1.897 \\
\bottomrule
\end{tabular}

\vspace*{1mm}
{\small (a) ShapeNet cars, closed meshes \cite{NEURIPS2019_39059724}, results $\times 10^{-4}$, cf.~\cite{chibane2020ndf}.}

\vspace*{3mm}
\setlength{\tabcolsep}{2pt}
\begin{tabular}{l|cc}
\toprule
Point sampling & IoU $\uparrow$ & Cha. $\ell_1$ $\downarrow$\\
\midrule
Resampled  & 0.666 & 0.112 \\
Fixed       & 0.669 & 0.106 \\
\bottomrule
\end{tabular}

\vspace*{1mm}
{\small (b) ShapeNet subset of\cite{choy20163d}, all classes, results $\times 10^{-1}$, cf.~\cite{mescheder2019occupancy}.}

\vspace*{3mm}
\setlength{\tabcolsep}{3pt}
\begin{tabular}{l|ccc}
\toprule
                &\multicolumn{3}{c}{Chamfer $\ell_2$ $\downarrow$}\\
Sampling  & 5\%   & 50\%& 95\%  \\
\midrule
Resampled  & 0.269 & 0.433 & 1.149 \\
Fixed       & 0.202 & 0.526 & 1.322 \\
\bottomrule
\end{tabular}

\vspace*{1mm}
{\small (c) DFaust, results $\times 10^{-3}$, cf.~\cite{atzmon2020sal}.}
\vspace*{1.5mm}
\caption{Performance of \method\ on complete point clouds with two different point sampling procedures at training time: 300 points newly sampled at each epoch on each shape (resampled); same 300 points sampled on each shape for all epochs (fixed).}
\label{tab:exp_complete2}
\vspace*{-1.5mm}
\end{table}

\section{Robustness to noise and density variations}
\label{supp:sec:noise}

In the main paper, as in previous works, we experimented without noise except for the multi-category ShapeNet experiment where we used a very low level of noise equal to the level used in~\cite{mescheder2019occupancy}.
In this section, we are interested in studying further how a model behaves when the number of input points and the level of noise differ from the values used in the training dataset.

On Figure~\ref{fig:noise}, we present a qualitative robustness study of the predictions of a network trained on the car subset of ShapeNet, with 300 input points and no noise.
With this single network, and for the same test shape, we predict the surface from various point cloud samplings: from no noise to a high level of noise, and from the ``native'' point cloud size used at training time (300 points) to a higher resolution (up to 10k points).

First, without noise, our model is globally insensitive to the number of points.
This is due to the PointNet structure in the encoder: the information being gathered using global max-pooling, no neighborhood or density notions are involved in the latent vector predictions.

Second, for the same small number of points,
the effect of noise is limited. We observe a progressive loss of details and an inflation of the 
shape, with hallucinated surfaces appearing inside the actual car shape.
This is a behavior somehow similar to what we can be observed when looking at the convex envelop of the point cloud with added noise.

Finally, the conjunction of noise and of an increased number of points worsens a lot the quality of the latent representations, and thus of the predicted shape.
The point-wise information aggregation of the encoder, which ensures robustness to the number of points when no noise is added, is conversely the source of non-robustness to noise: no local averaging effect is then possible.
In addition, as the network has been trained without noise, each point is considered meaningful, thus leading to a wrong 
latent vector and a resulting surface that is very complex, with a lot of folds.
This configuration is particularly challenging for the networks as it differs a lot from the training conditions.

A solution to palliate this degradation could be to add noise and use variable point cloud sizes at training time, or to pre-process the point cloud with a denoising and outlier removal algorithm~\cite{han2017review}.

\begin{figure}[!t]
\begin{center}
    \setlength{\tabcolsep}{2.5pt}
    \begin{tabular}{@{}ccc@{}}
        \includegraphics[width=0.32\linewidth, trim=30 0 0 0, clip]{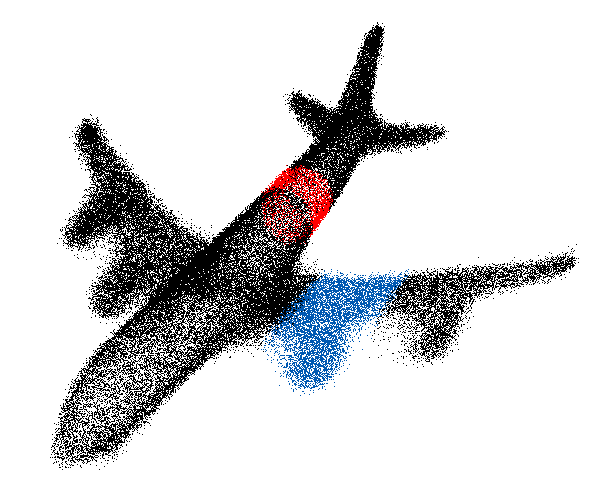}&
        \includegraphics[width=0.22\linewidth]{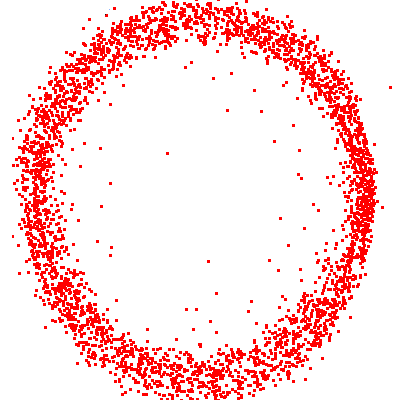}&
        \includegraphics[width=0.32\linewidth]{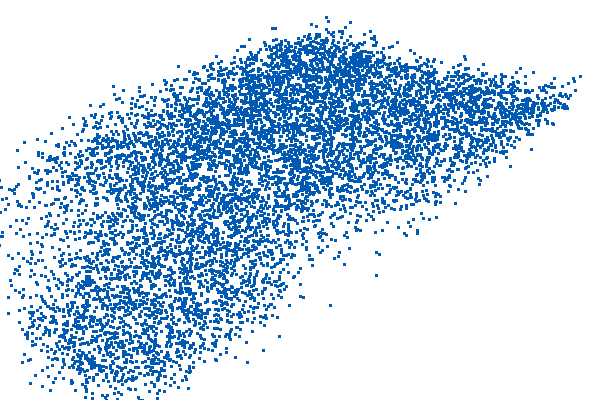} \\
        & {\small (a) ShapeGF}
        \\
        \includegraphics[width=0.36\linewidth]{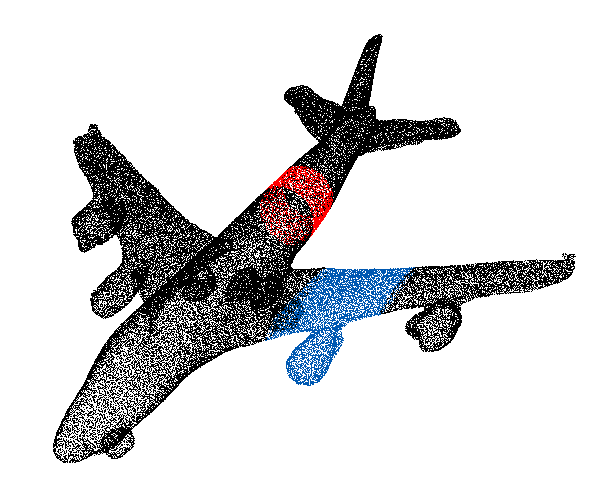}&
        \includegraphics[width=0.22\linewidth]{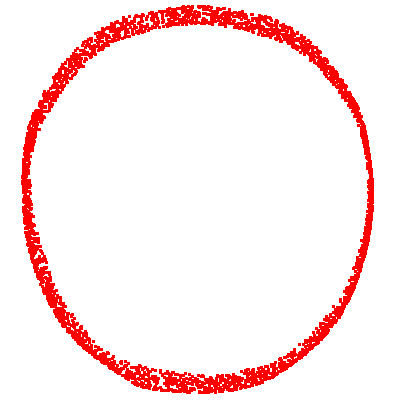}&
        \includegraphics[width=0.32\linewidth]{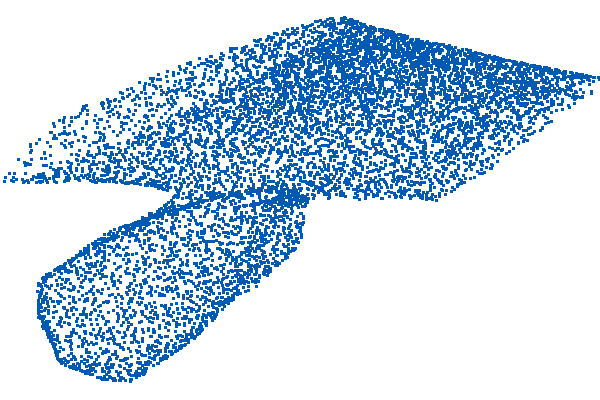}
        \\
        & {\small (b) NeeDrop}
    \end{tabular}
    \vspace*{2mm}
    \caption{Comparing the point cloud generated by ShapeGF to a point cloud (of the same size) sampled on our generated mesh.}
    \label{compareshapegf}
    \vspace{-3mm}
\end{center}
\end{figure}

\begin{figure*}[!p]
    \centering
    \setlength{\tabcolsep}{1.25pt}
    \begin{tabular}{ccccccc}
        & & \multicolumn{5}{c}{Noise level (standard deviation of the Gaussian noise)}\\
        & & 0 & 0.005 & 0.007 & 0.01 & 0.02 \\
    
        &&
        \includegraphics[width=0.155\linewidth]{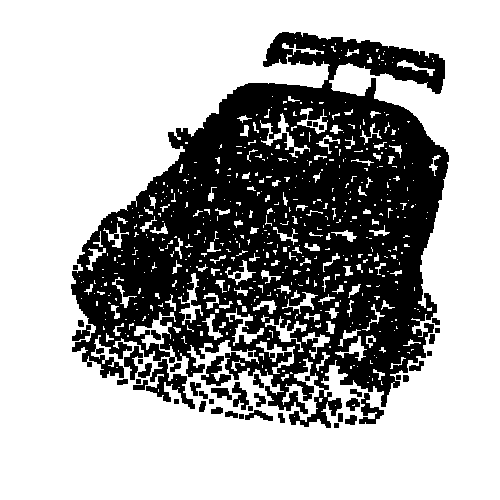}&
        \includegraphics[width=0.155\linewidth]{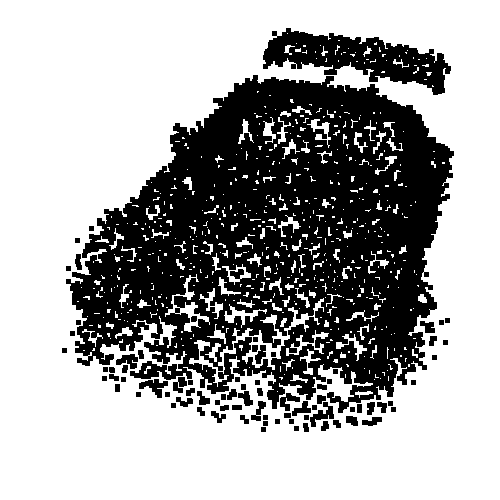}&
        \includegraphics[width=0.155\linewidth]{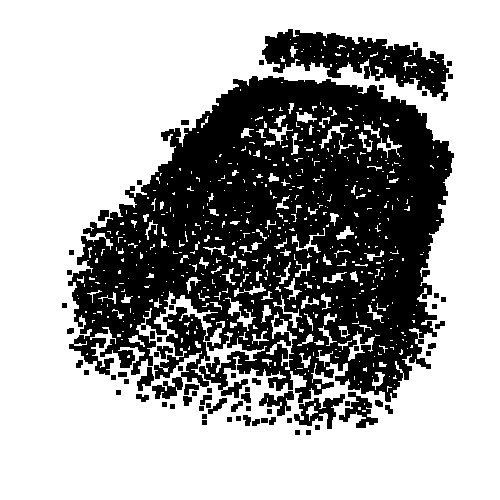}&
        \includegraphics[width=0.155\linewidth]{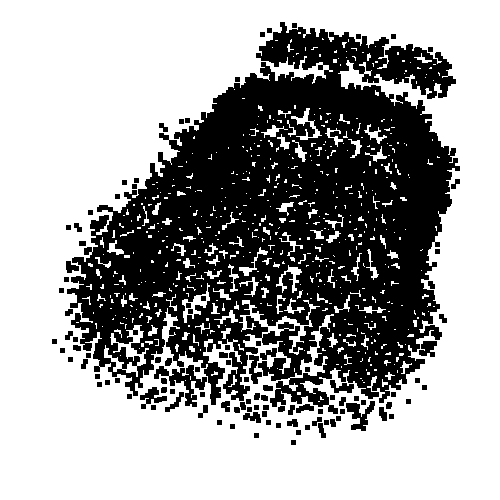}&
        \includegraphics[width=0.155\linewidth]{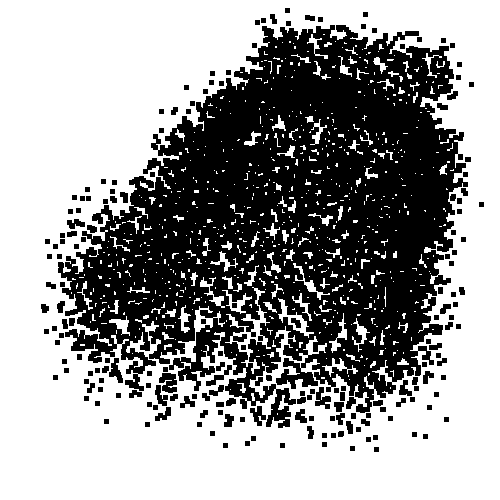}\\
        
        \rot{10k points}&
        \includegraphics[width=0.155\linewidth]{images/noise_npts/10000_000_p.png}&
        \includegraphics[width=0.155\linewidth]{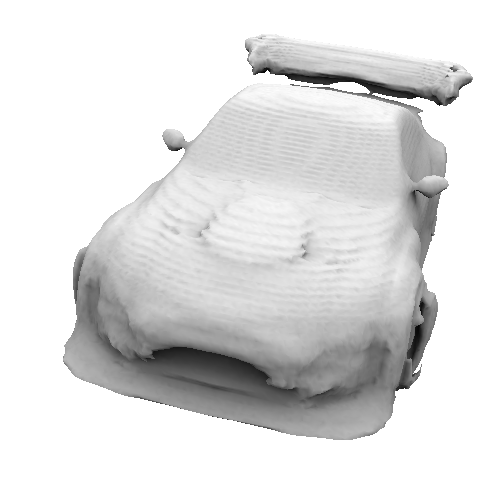}&
        \includegraphics[width=0.155\linewidth]{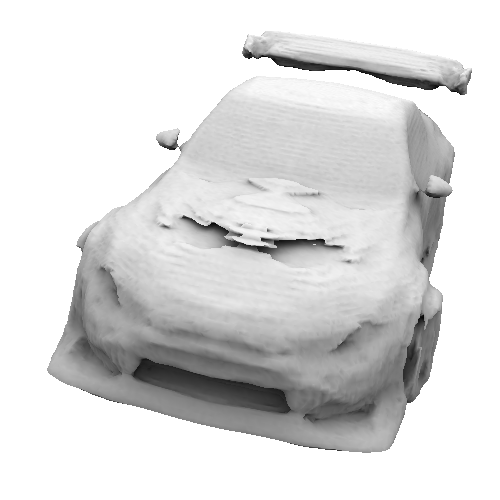}&
        \includegraphics[width=0.155\linewidth]{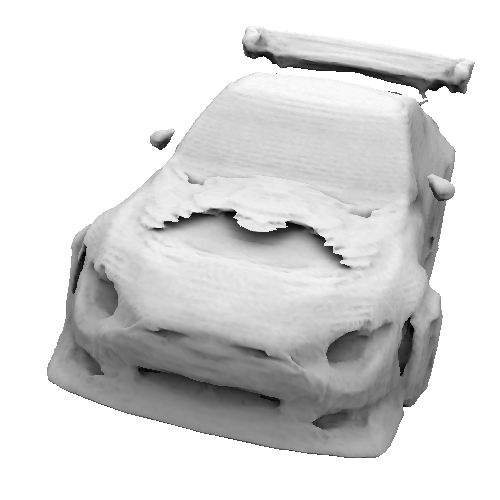}&
        \includegraphics[width=0.155\linewidth]{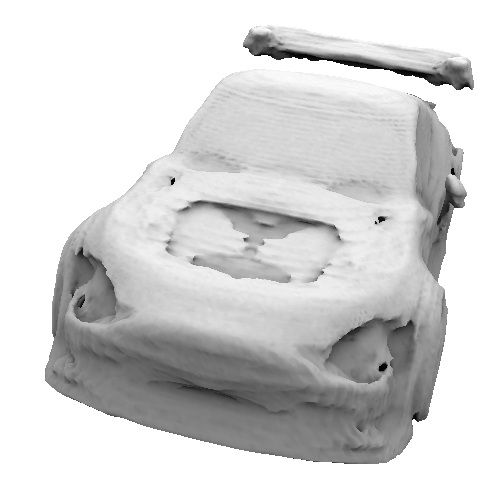}&
        \includegraphics[width=0.155\linewidth]{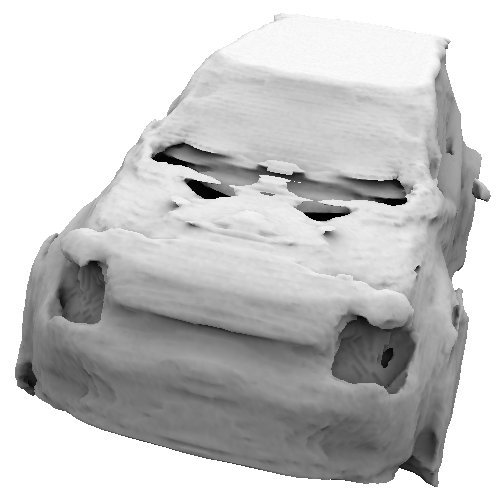}\\
        
        \rot{5000 points}&
        \includegraphics[width=0.155\linewidth]{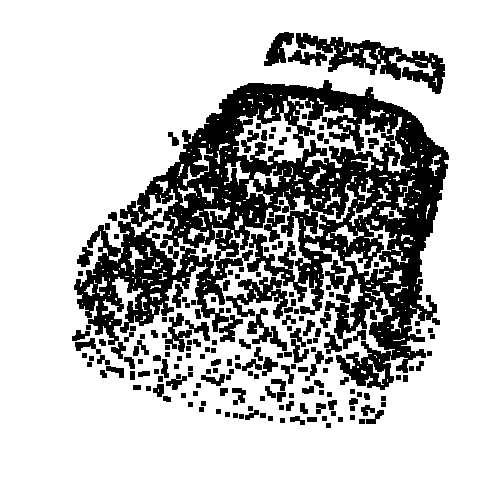}&
        \includegraphics[width=0.155\linewidth]{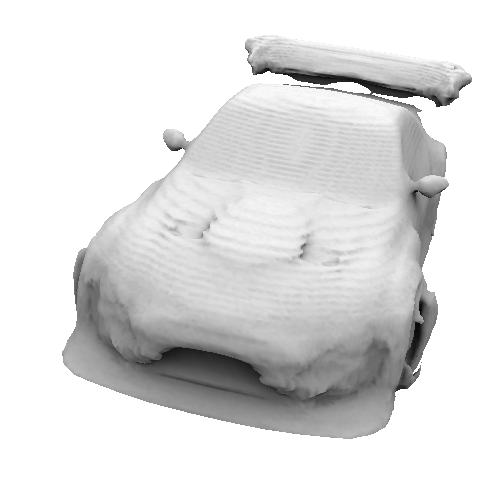}&
        \includegraphics[width=0.155\linewidth]{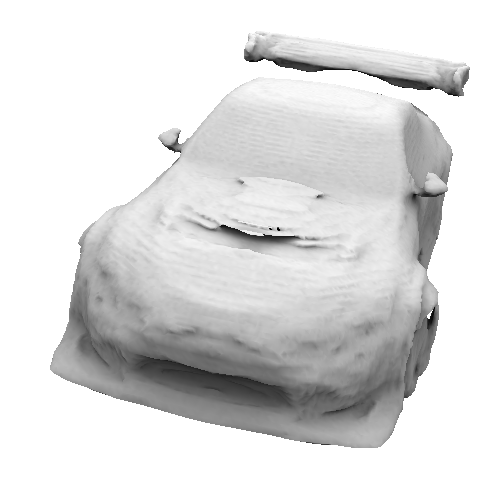}&
        \includegraphics[width=0.155\linewidth]{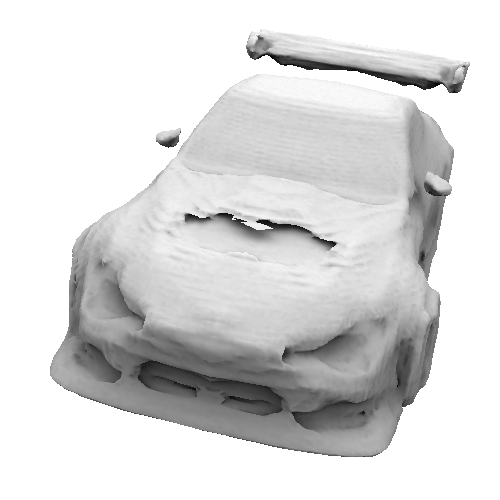}&
        \includegraphics[width=0.155\linewidth]{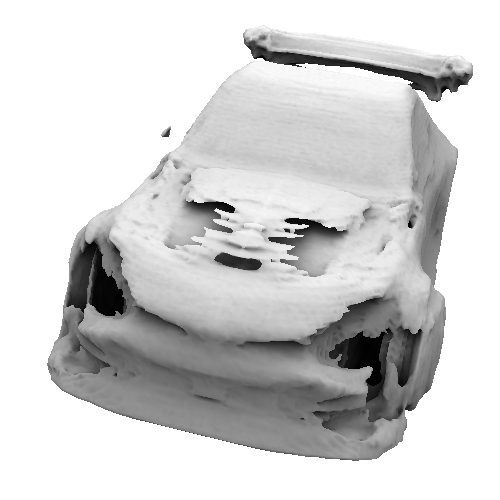}&
        \includegraphics[width=0.155\linewidth]{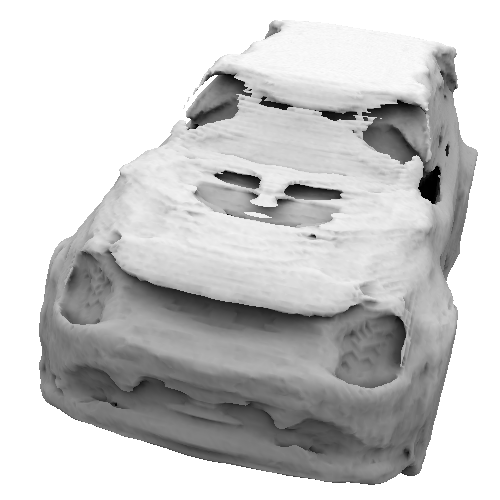}\\
        
        \rot{1000 points}&
        \includegraphics[width=0.155\linewidth]{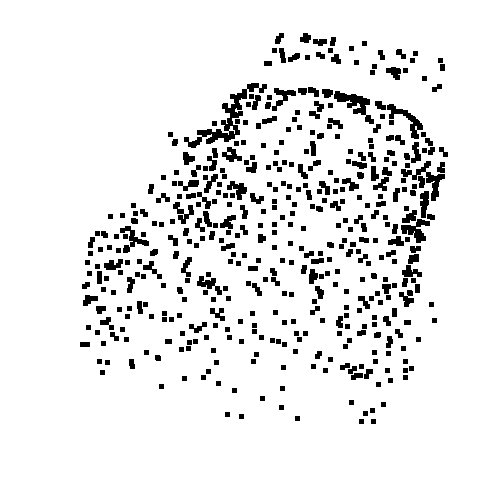}&
        \includegraphics[width=0.155\linewidth]{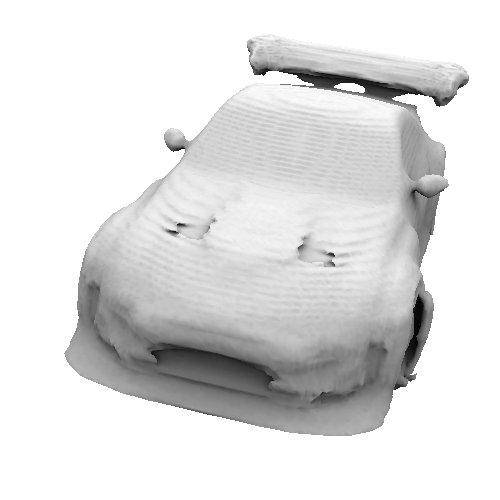}&
        \includegraphics[width=0.155\linewidth]{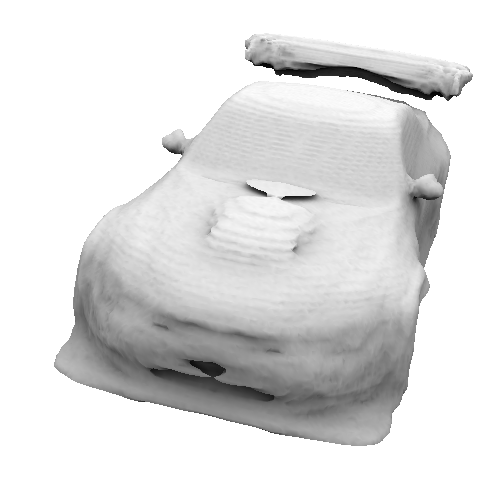}&
        \includegraphics[width=0.155\linewidth]{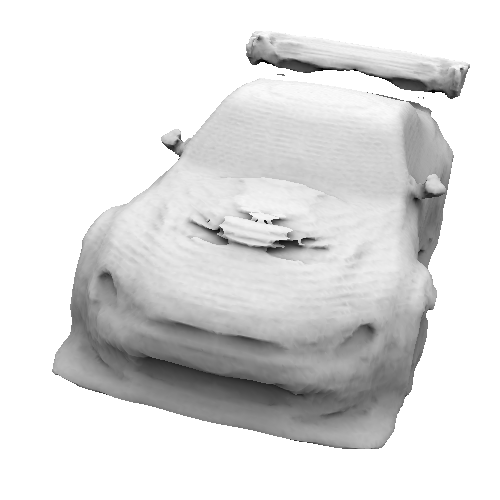}&
        \includegraphics[width=0.155\linewidth]{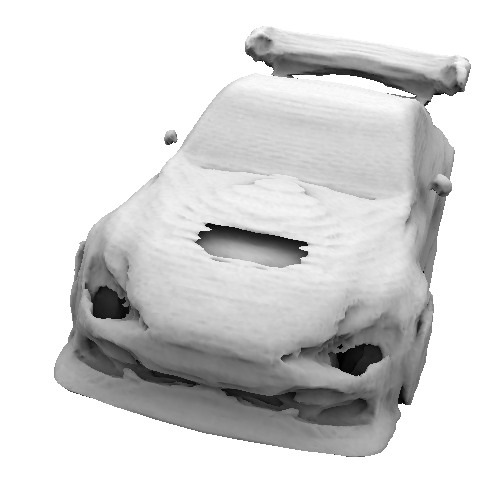}&
        \includegraphics[width=0.155\linewidth]{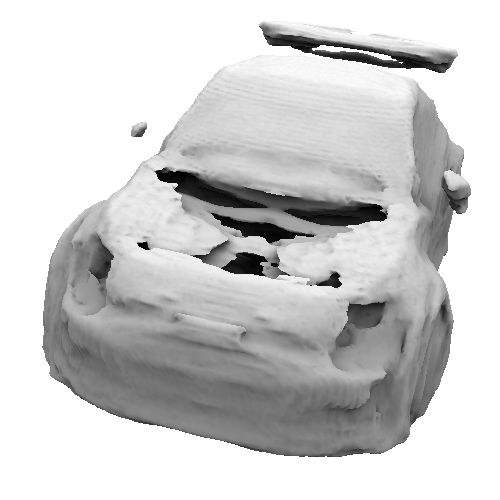}\\
        
        \rot{300 points}&
        \includegraphics[width=0.155\linewidth]{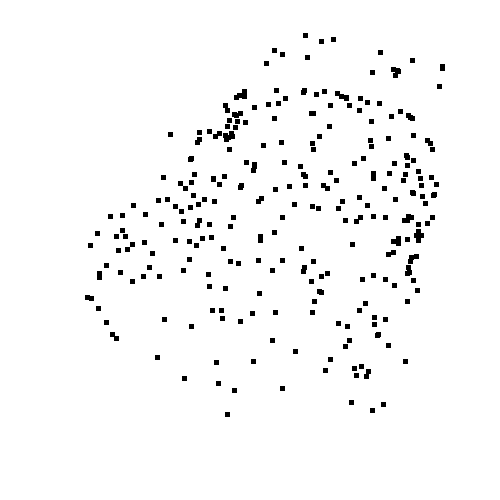}&
        \includegraphics[width=0.155\linewidth]{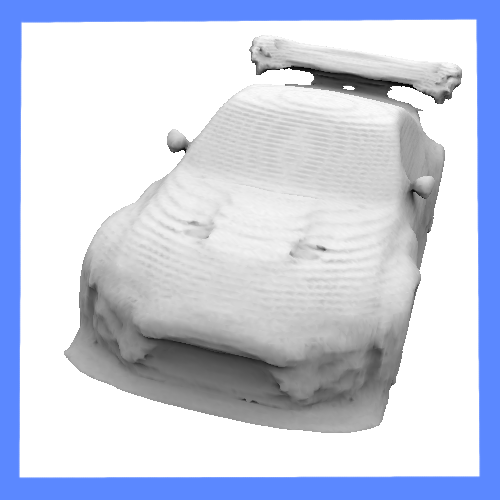}&
        \includegraphics[width=0.155\linewidth]{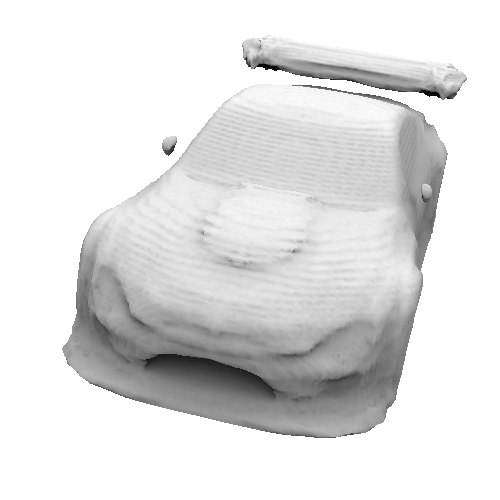}&
        \includegraphics[width=0.155\linewidth]{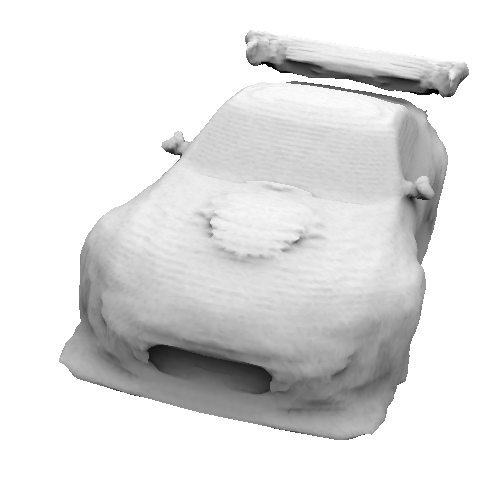}&
        \includegraphics[width=0.155\linewidth]{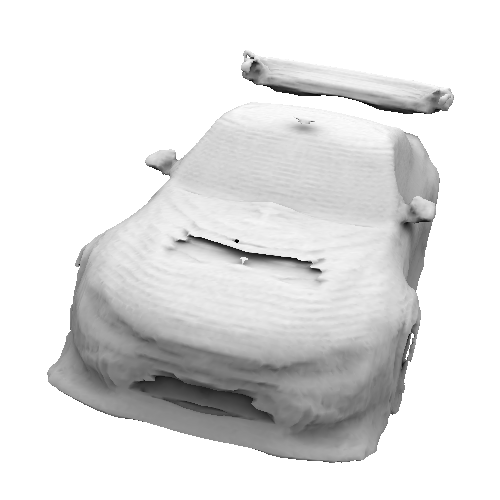}&
        \includegraphics[width=0.155\linewidth]{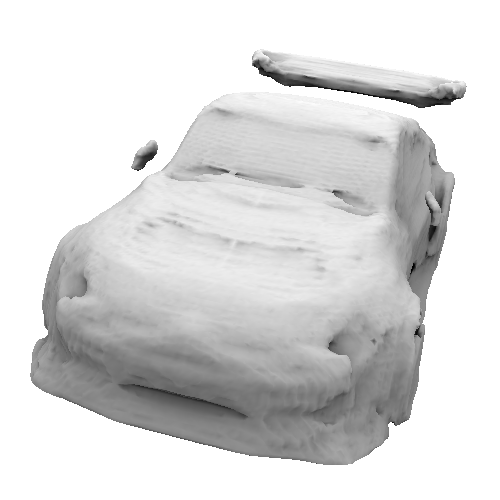}\\
        &\makecell[b]{
        $\uparrow$ \\ \emph{Increasing number} \\ \emph{of points}
        \\~\\  
        $\rightarrow$ \\ \emph{Increasing level} \\ \emph{of noise}
        }&
        \includegraphics[width=0.155\linewidth]{images/noise_npts/00300_000_p.png}&
        \includegraphics[width=0.155\linewidth]{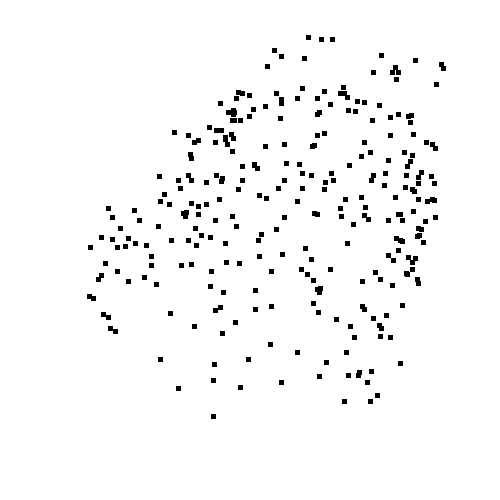}&
        \includegraphics[width=0.155\linewidth]{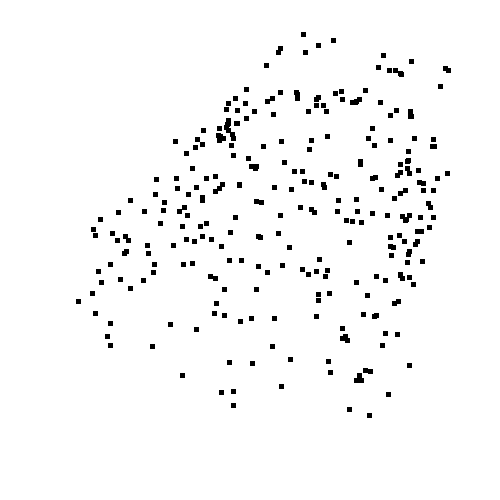}&
        \includegraphics[width=0.155\linewidth]{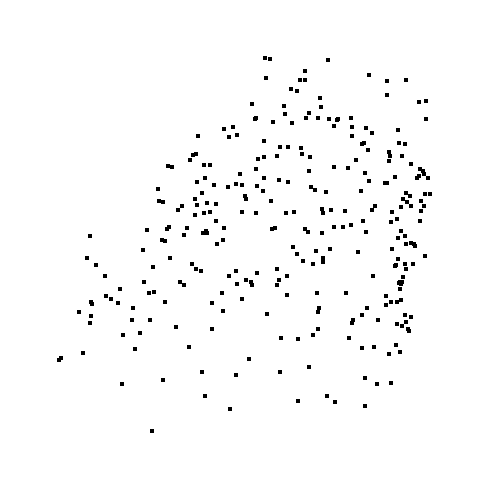}&
        \includegraphics[width=0.155\linewidth]{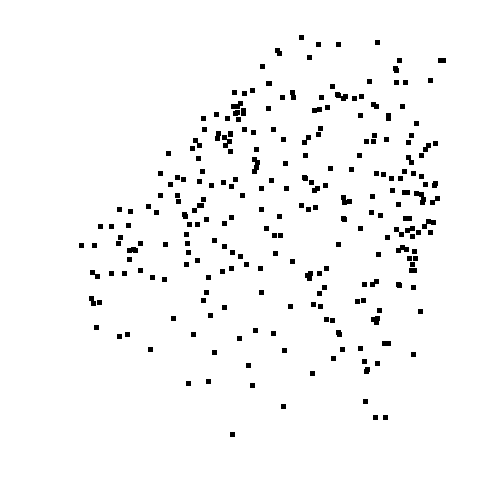}\\
        ~\\
    \end{tabular}
    \caption{Robustness of a single network to a varying number of points and a varying level of input noise. The framed reconstruction corresponds to the number of points and to the level of points used at training time; all predictions have been made with the same network.}
    \label{fig:noise}
\end{figure*}

\section{Comparing to ShapeGF}
\label{supp:sec:shapegf}

To illustrate why we wrote in Section~4.1 that ShapeGF \cite{cai2020learning} yields ``{noisy}'' points (cf.\ Fig.\,3(a)), we scanned as many points on our generated surface, and zoomed on a slice of the cabin and on an engine (see Fig.\,\ref{compareshapegf}). With ShapeGF, the aircraft engines are somehow blurry and not well localized, and the two surfaces of thin volumes, such as wings, cannot be told apart.

\section{Reconstructing KITTI shapes}
\label{supp:sec:transfer}

\begin{figure*}[!p]
\centering
\setlength{\tabcolsep}{1pt}
\renewcommand{\arraystretch}{0.9}
\begin{tabular}{cc|ccc|c|>{\columncolor{blue!5}}c>{\columncolor{blue!5}}c|}
&& Poisson & Ball pivot. &
{AtlasNet}&
\multicolumn{3}{c}{\method}\\
\cmidrule{3-8} 
\multicolumn{2}{r|}{Training set\mbox{~~}}
& - & - & KITTI & KITTI 
& \multicolumn{2}{c}{ShapeNet} \\
{\makecell{Reference\\Image}}&
{Input}&
&
&
{\makecell{raw point\\clouds}}&
{\makecell{raw point\\clouds}}
&
{\makecell{complete\\shapes}}&
{\makecell{partial\\shapes}}
\\
\includegraphics[width=0.1\linewidth]{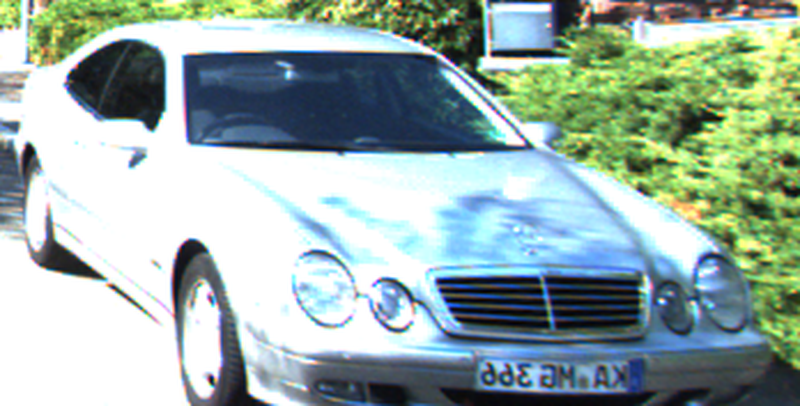}&
\includegraphics[width=0.1\linewidth]{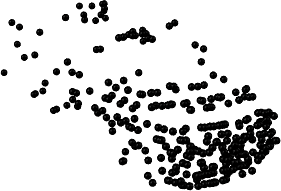}&
\includegraphics[width=0.1\linewidth]{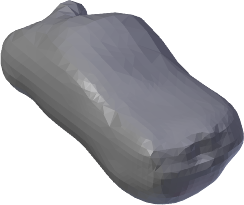}&
\includegraphics[width=0.1\linewidth]{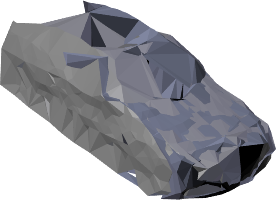}&
\includegraphics[width=0.1\linewidth]{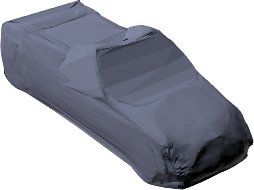}&
\includegraphics[width=0.1\linewidth]{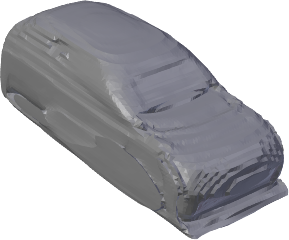}&
\includegraphics[width=0.1\linewidth]{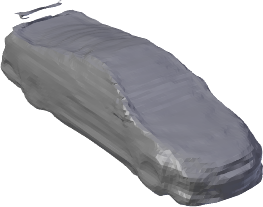}&
\includegraphics[width=0.1\linewidth]{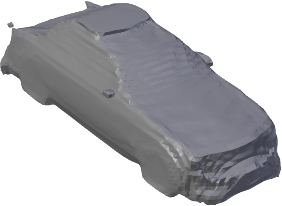}
\\
\includegraphics[width=0.1\linewidth]{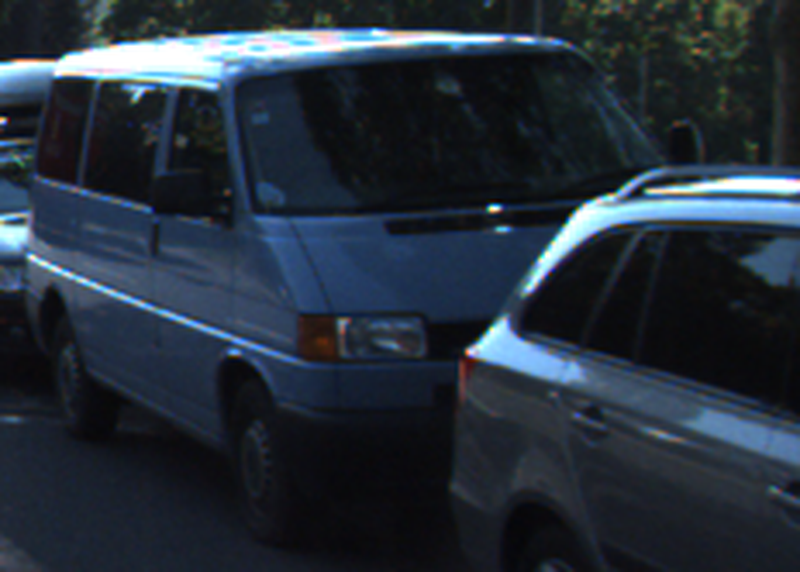}&
\includegraphics[width=0.1\linewidth]{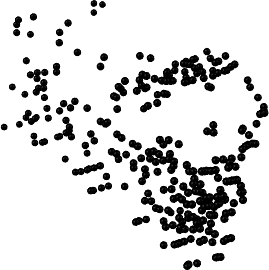}&
\includegraphics[width=0.1\linewidth]{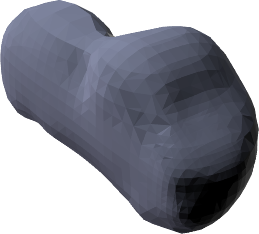}&
\includegraphics[width=0.1\linewidth]{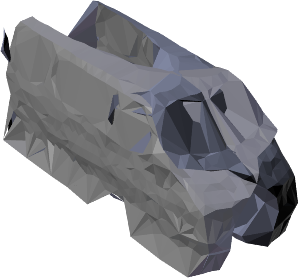}&
\includegraphics[width=0.1\linewidth]{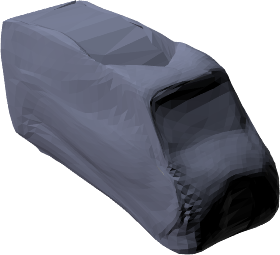}&
\includegraphics[width=0.1\linewidth]{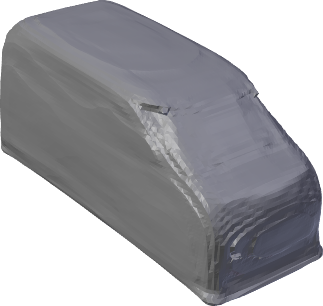}&
\includegraphics[width=0.1\linewidth]{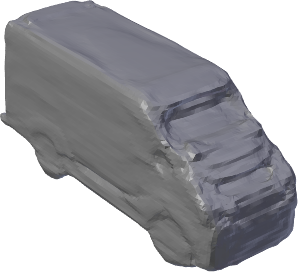}&
\includegraphics[width=0.1\linewidth]{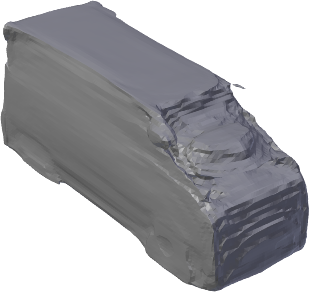}
\\
\includegraphics[width=0.1\linewidth]{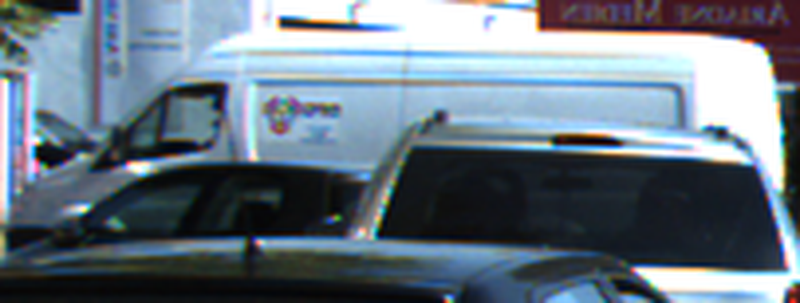}&
\includegraphics[width=0.1\linewidth]{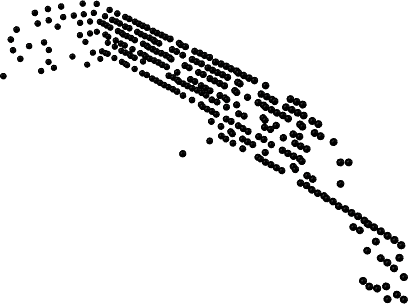}&
\includegraphics[width=0.1\linewidth]{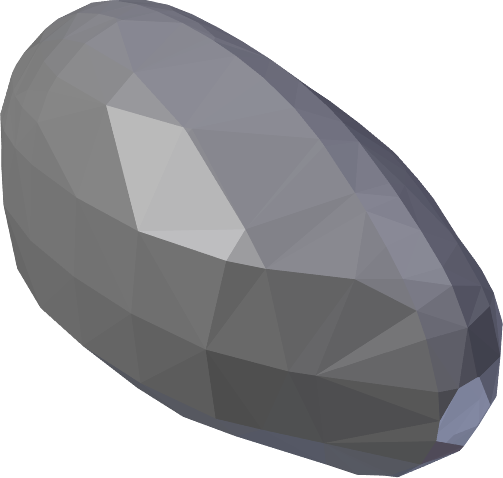}&
\includegraphics[width=0.1\linewidth]{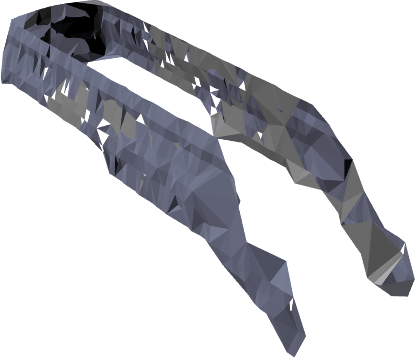}&
\includegraphics[width=0.1\linewidth]{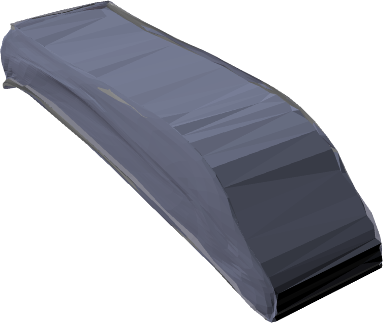}&
\includegraphics[width=0.1\linewidth]{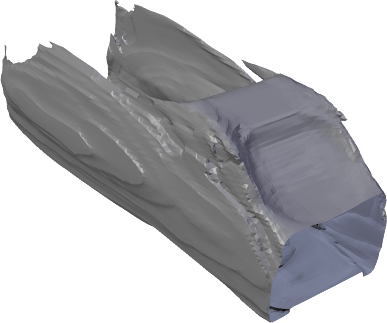}&
\includegraphics[width=0.1\linewidth]{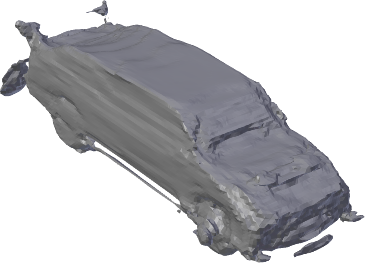}&
\includegraphics[width=0.1\linewidth]{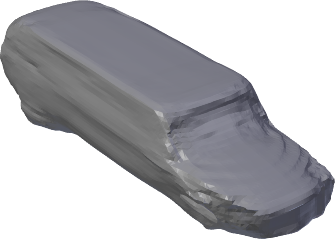}
\\
\includegraphics[width=0.1\linewidth]{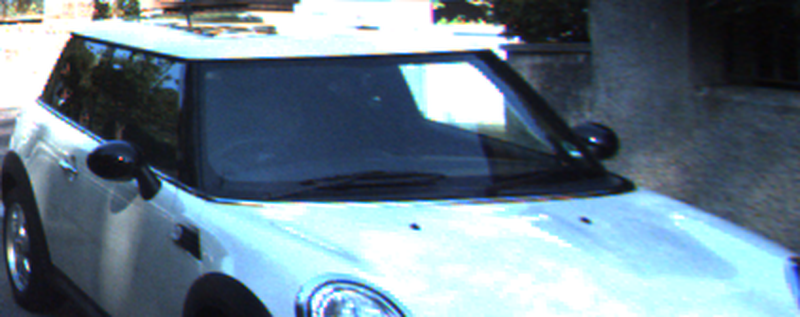}&
\includegraphics[width=0.1\linewidth]{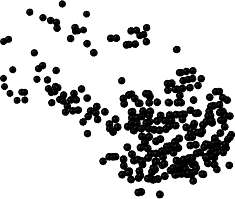}&
\includegraphics[width=0.1\linewidth]{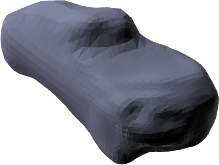}&
\includegraphics[width=0.1\linewidth]{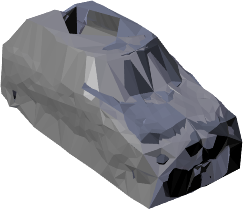}&
\includegraphics[width=0.1\linewidth]{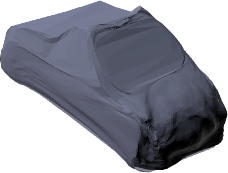}&
\includegraphics[width=0.1\linewidth]{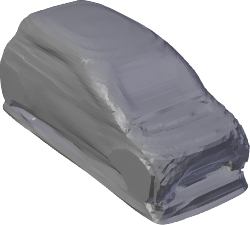}&
\includegraphics[width=0.1\linewidth]{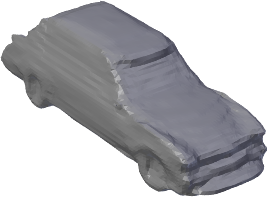}&
\includegraphics[width=0.1\linewidth]{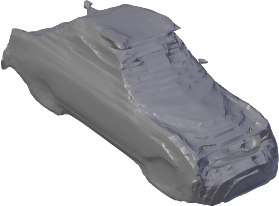}
\\
\includegraphics[width=0.1\linewidth]{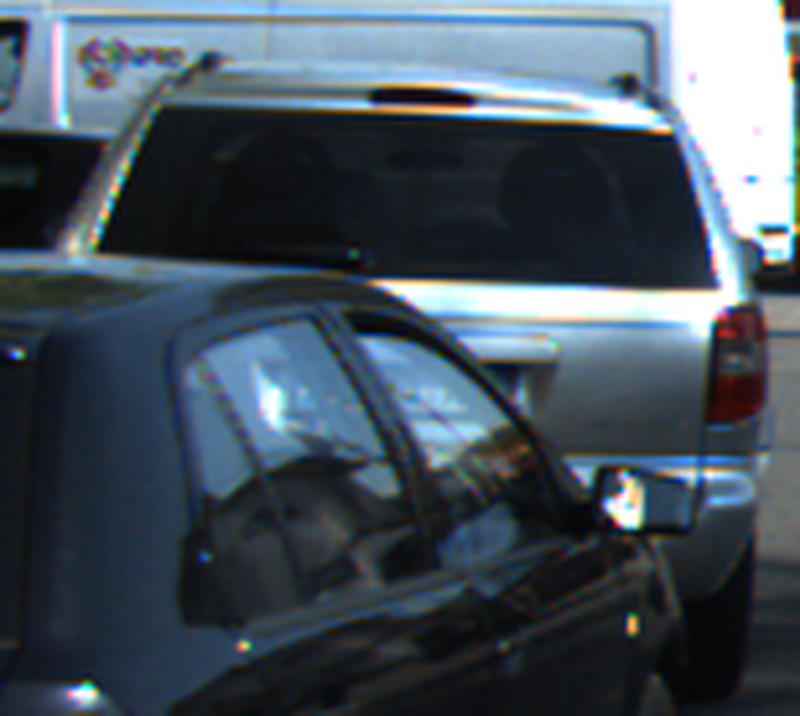}&
\includegraphics[width=0.1\linewidth]{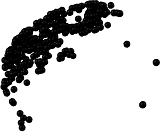}&
\includegraphics[width=0.1\linewidth]{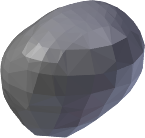}&
\includegraphics[width=0.1\linewidth]{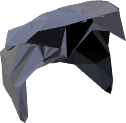}&
\includegraphics[width=0.1\linewidth]{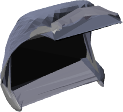}&
\includegraphics[width=0.1\linewidth]{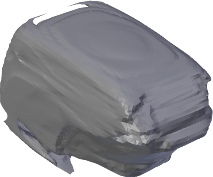}&
\includegraphics[width=0.1\linewidth]{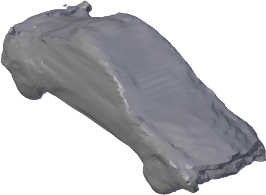}&
\includegraphics[width=0.1\linewidth]{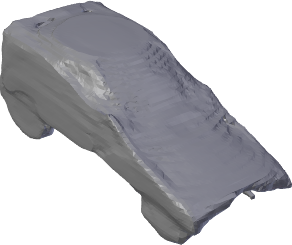}
\\
\includegraphics[width=0.1\linewidth]{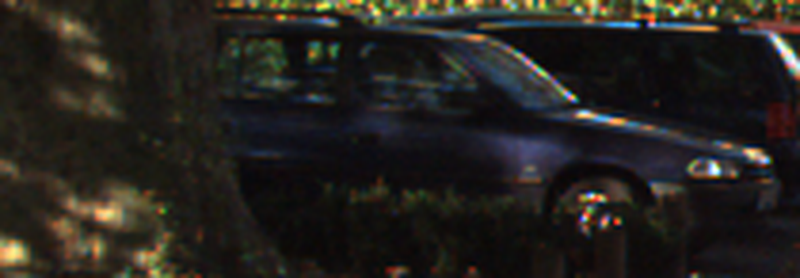}&
\includegraphics[width=0.1\linewidth]{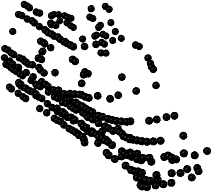}&
\includegraphics[width=0.1\linewidth]{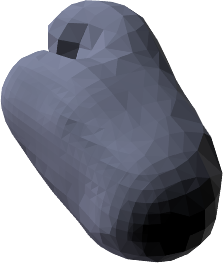}&
\includegraphics[width=0.1\linewidth]{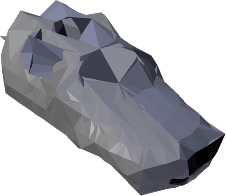}&
\includegraphics[width=0.1\linewidth]{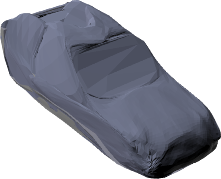}&
\includegraphics[width=0.1\linewidth]{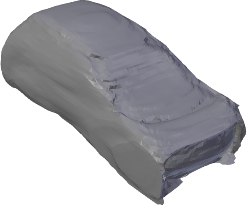}&
\includegraphics[width=0.1\linewidth]{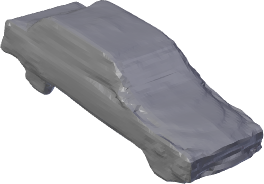}&
\includegraphics[width=0.1\linewidth]{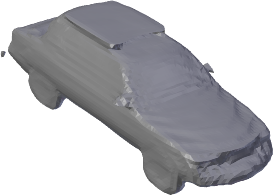}
\\
\includegraphics[width=0.1\linewidth]{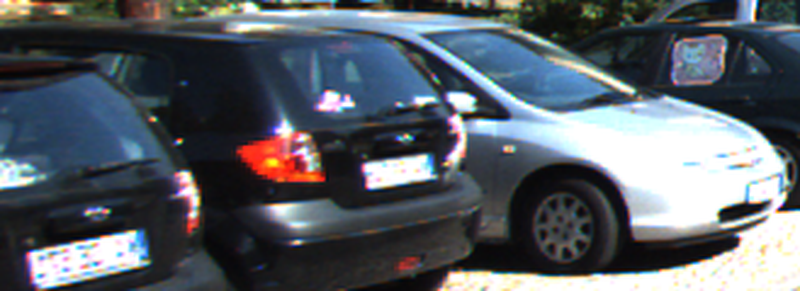}&
\includegraphics[width=0.1\linewidth]{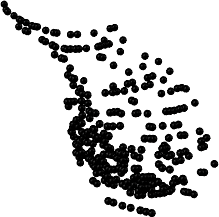}&
\includegraphics[width=0.1\linewidth]{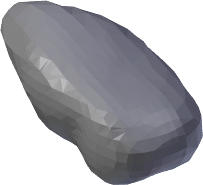}&
\includegraphics[width=0.1\linewidth]{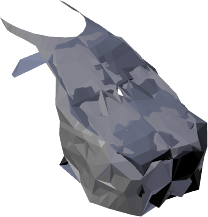}&
\includegraphics[width=0.1\linewidth]{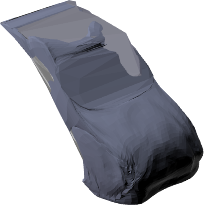}&
\includegraphics[width=0.1\linewidth]{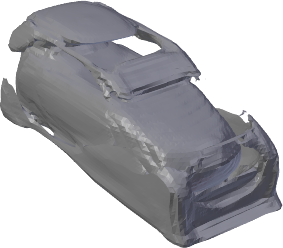}&
\includegraphics[width=0.1\linewidth]{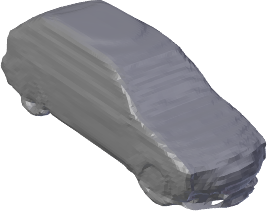}&
\includegraphics[width=0.1\linewidth]{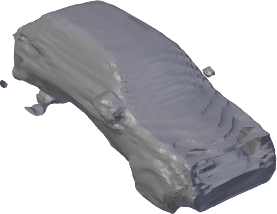}
\\
\includegraphics[width=0.1\linewidth]{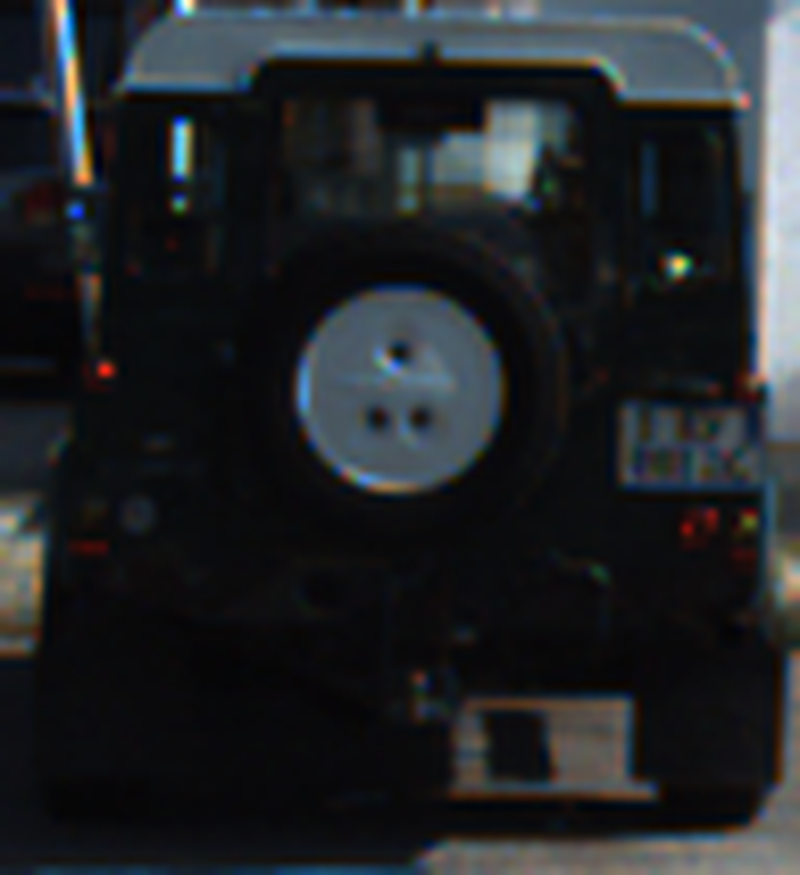}&
\includegraphics[width=0.1\linewidth]{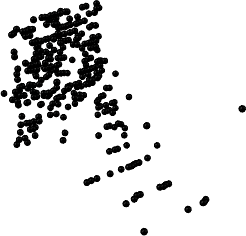}&
\includegraphics[width=0.1\linewidth]{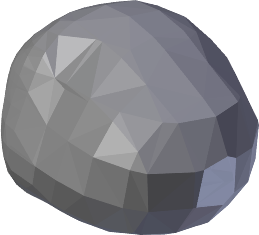}&
\includegraphics[width=0.1\linewidth]{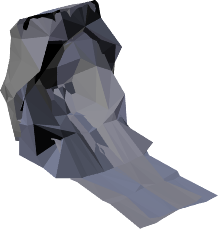}&
\includegraphics[width=0.1\linewidth]{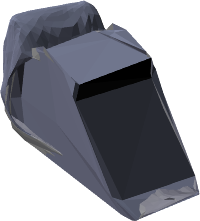}&
\includegraphics[width=0.1\linewidth]{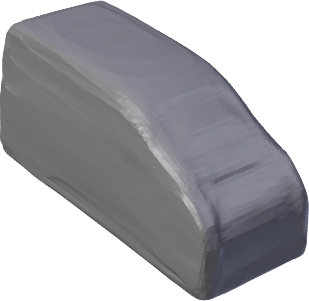}&
\includegraphics[width=0.1\linewidth]{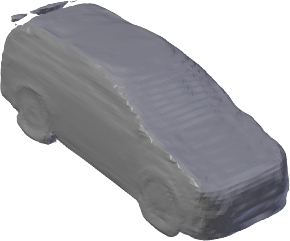}&
\includegraphics[width=0.1\linewidth]{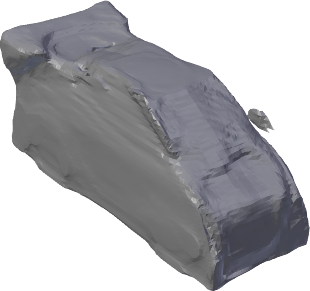}
\\
\includegraphics[width=0.1\linewidth]{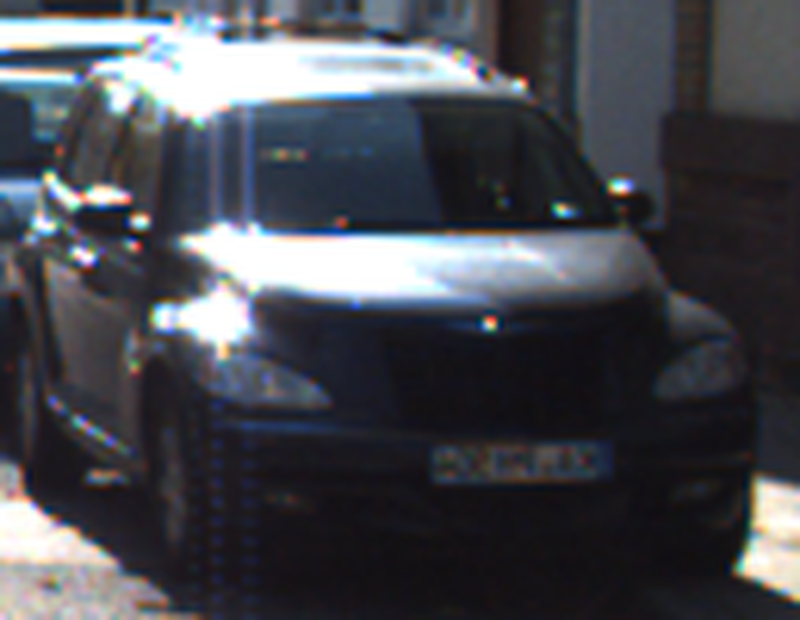}&
\includegraphics[width=0.1\linewidth]{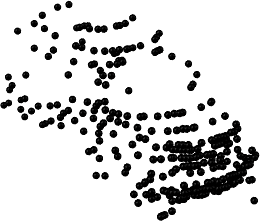}&
\includegraphics[width=0.1\linewidth]{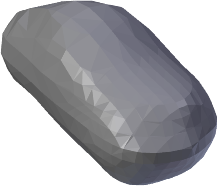}&
\includegraphics[width=0.1\linewidth]{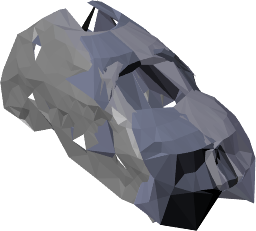}&
\includegraphics[width=0.1\linewidth]{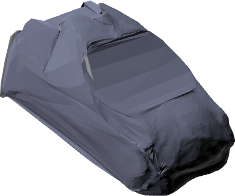}&
\includegraphics[width=0.1\linewidth]{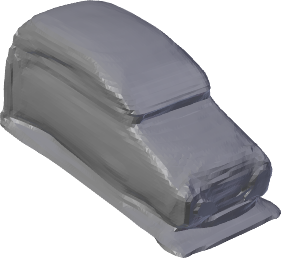}&
\includegraphics[width=0.1\linewidth]{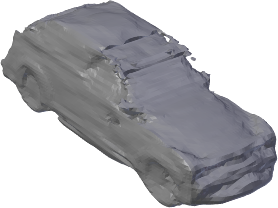}&
\includegraphics[width=0.1\linewidth]{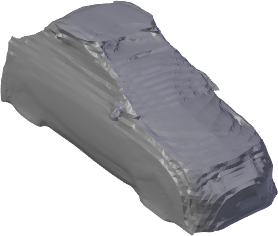}
\\
\includegraphics[width=0.1\linewidth]{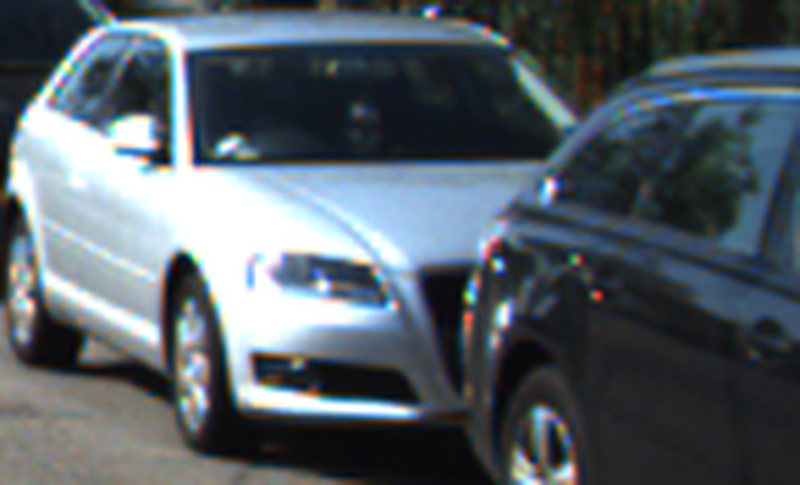}&
\includegraphics[width=0.1\linewidth]{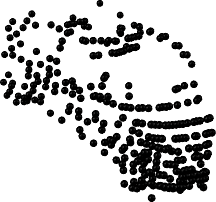}&
\includegraphics[width=0.1\linewidth]{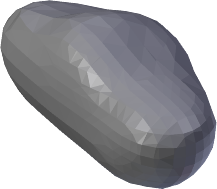}&
\includegraphics[width=0.1\linewidth]{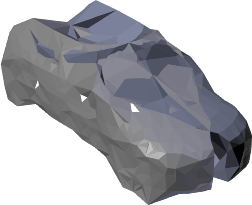}&
\includegraphics[width=0.1\linewidth]{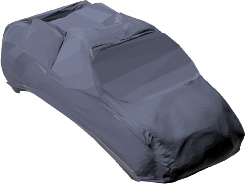}&
\includegraphics[width=0.1\linewidth]{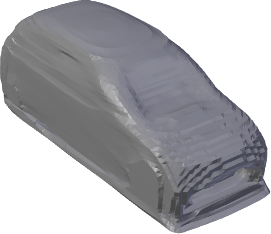}&
\includegraphics[width=0.1\linewidth]{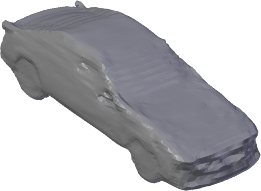}&
\includegraphics[width=0.1\linewidth]{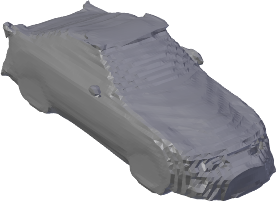}
\\
\includegraphics[width=0.1\linewidth]{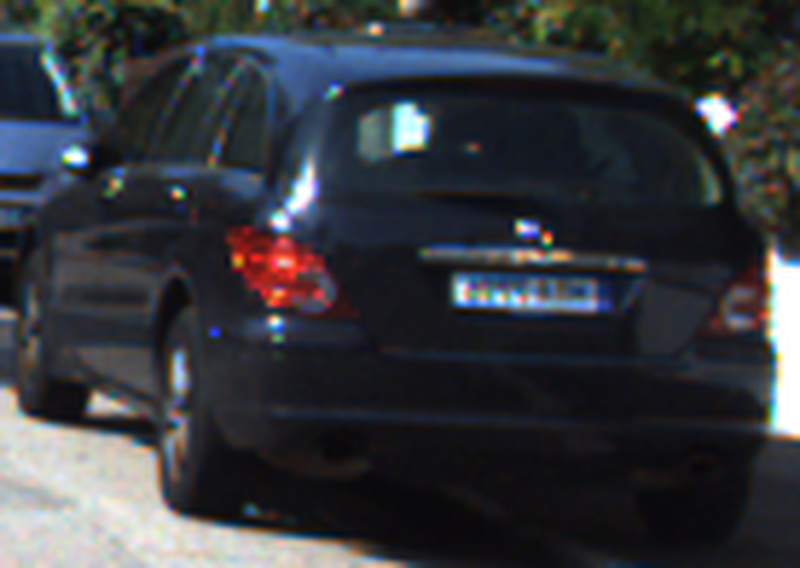}&
\includegraphics[width=0.1\linewidth]{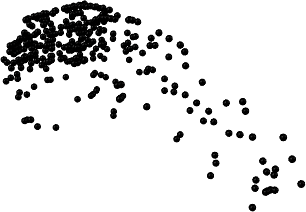}&
\includegraphics[width=0.1\linewidth]{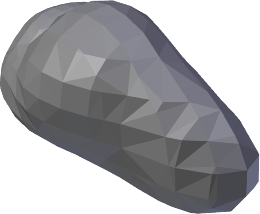}&
\includegraphics[width=0.1\linewidth]{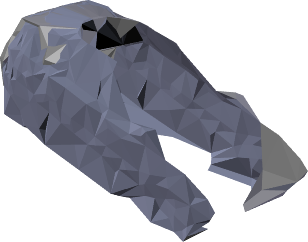}&
\includegraphics[width=0.1\linewidth]{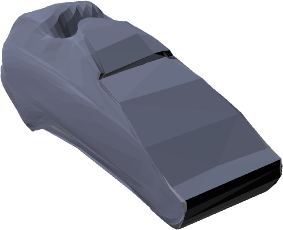}&
\includegraphics[width=0.1\linewidth]{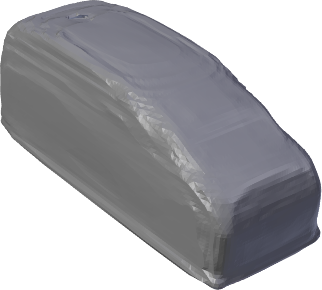}&
\includegraphics[width=0.1\linewidth]{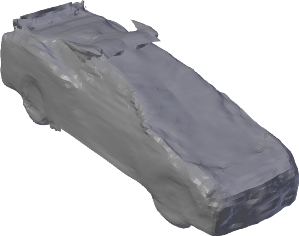}&
\includegraphics[width=0.1\linewidth]{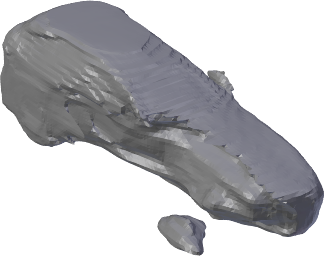}
\\
\end{tabular}
\vspace*{3mm}
\caption{Comparison of reconstructions on KITTI car point clouds.}
\label{fig:kitti}
\vspace*{-3mm}
\end{figure*}

Figure~\ref{fig:kitti} presents more results in the same setting as Figure~3(c) in the main paper. AtlasNet is trained on the KITTI point clouds.

The figure also shows the transfer capacity of our model between two datasets: training on ShapeNet and testing on KITTI.
We actually provide transfer results for two models trained on the ShapeNet dataset: a first model trained with complete shapes (model from the main paper), and a second model trained with partial point clouds.
For the latter model, at training time, we generate partial point clouds on the fly using the visibility operator of~\cite{katz2007direct} to simulate visibility from a single viewpoint.
As there exists a symmetry plane (vertical, from front to rear of the car) for most cars, we exploit this symmetry at training time, by concatenating the sampled point cloud and its mirror-image symmetrized version. Please note that although it helps densifying and completing the training point clouds, it does not guarantee a complete coverage of the whole shape. In particular, cars that are seen from the front are not completed in the back and conversely. Besides, the underside of cars remain unsampled as their are never seen by the lidar.

The domain gaps include: (1) going from a training on points uniformly sampled on perfect shapes without noise to a training on points captured by an actual lidar, and (2) going from a training on complete shapes to a training on partial shapes (i.e., point cloud not covering the whole shape).

We observe that all models transfer well to the KITTI dataset, producing plausible reconstructions.

However, contrary to our intuition, we also observe that the model trained on partial views does not perform better than the model trained on the complete point clouds.
To our understanding,
this is mainly due to the fact a complete point cloud carries much more information than a partial one.
The full-model latent space is therefore more guided and constrained by the whole geometry of a car (front and rear should exists, with four wheels for each model, etc.), compared to a space learned only on partial point clouds.

\paragraph{Details of the training for the KITTI experiment.}

The object-specific point clouds for cars and pedestrians are extracted from the lidar scans using the 3D bounding boxes provided for the KITTI 3D detection challenge.
Training point clouds are generated using the training 3D boxes.
The shapes presented in the paper and in the supplementary material are not generated from points seen at training time; they are generated using point clouds extracted from validation 3D boxes.

At training time, as well as test time, we sample 300 points inside a given 3D bounding box.
If the box contains less than 300 points, some points are re-sampled to reach the desired number of points.
In practice, duplicating points (in case there is less than 300 points) has no particular effect on the predicted latent vector, beyond the fact that there is less information to rely on. Indeed, the backbone (a residual PointNet) processes the points independently, and operates feature aggregation with max-pooling operations, which are invariant to redundancy.


\begin{lstlisting}[language=Python,caption={\raisebox{4mm}{}Needrop loss pseudo-code (PyTorch style)},label={lst:loss},float=*]
class NeeDropLoss(Function):
  # input0, input1: network output logit (without sigmoid) for first and second needle end point
  # target: needle label for the loss to be computed with 1 for same-side, 0 for opposite-side

  def forward(ctx, input0, input1, target):
  
    ctx.save_for_backward(input0, input1, target) # save for backward

    # clamp to avoid infinite value in logarithm
    input0, input1 = input0.clamp(-10,10), input1.clamp(-10,10)

    e0, e1 = exp(input0), exp(input1) 
    
    loss = log(1+e0) + log(1+e1)
    mask = (target==1) # term specific to label 1 (same-side)
    loss[mask] = loss[mask] - log(1+e0*e1)[mask]
    mask = (target==0) # term specific to label 0 (opposite-side)
    loss[mask] = loss[mask] - log(e0+e1)[mask]
    return loss

  def backward(ctx, grad_output):
  
    input0, input1, target = ctx.saved_tensors # get saved input and targets

    # gradient for input_0
    grad_input0 = sigmoid(input0)
    mask = (target==1) # term specific to label 1 (same-side)
    grad_input0[mask] = grad_input0[mask] - sigmoid(input0+input1)[mask]
    mask = (target==0) # term specific to label 0 (opposite-side)
    grad_input0[mask] = grad_input0[mask] - sigmoid(input0-input1)[mask]
    grad_input0 = grad_input0 * grad_output
    
     # gradient for input_1
    grad_input1 = ... # exact same operations as for grad_input0 with input1
    
    return grad_input0, grad_input1, None
\end{lstlisting}

\section{Pseudo-code of the loss}
\label{supp:sec:loss}

Listing~\ref{lst:loss} provides the pseudo-code of the {\method} loss for Python with a Pytorch function implementation style.
This function implements equations (13) and (14) from the paper.
The only extra step needed is the clamping of the inputs in the forward function to prevent the logarithm to produce infinite values.
Please note that this operation is only used in the forward function, and that it has no influence on the backward function.

\end{document}